%% file: main.tex
\documentclass{article}
\makeatletter
\def\input@path{{styles/}}
\makeatother
\usepackage[preprint]{colm2026_conference}
\usepackage{fontspec}

\renewcommand{\encodingdefault}{T1}
\normalfont
\usepackage[protrusion=false]{microtype}

\input{math_commands.tex}

\usepackage{graphicx}
\usepackage{booktabs}
\usepackage{colortbl}
\usepackage{multirow}
\usepackage{tabularx}
\usepackage{wrapfig}
\usepackage{needspace}
\let\sectionwithoutfloatbarrier\section
\usepackage[section]{placeins}
\usepackage{caption}
\usepackage{subcaption}
\usepackage{listings}
\usepackage[most]{tcolorbox}
\usepackage{hyperref}
\usepackage{url}
\usepackage{amsthm}

\definecolor{backbonegray}{gray}{0.93}
\definecolor{oursblue}{RGB}{235,243,252}
\definecolor{resultfirst}{HTML}{E2F0D9}
\definecolor{resultsecond}{HTML}{FFF2CC}
\definecolor{resultthird}{HTML}{FFF2CC}
\definecolor{codeheaderblue}{HTML}{3478A8}
\definecolor{codebackground}{HTML}{F5F6F7}
\definecolor{codekeyword}{HTML}{005CC5}
\definecolor{codecomment}{HTML}{5C6370}
\definecolor{codestring}{HTML}{22863A}
\definecolor{exposurecode}{HTML}{C62828}
\definecolor{tablegapred}{RGB}{249,179,182}
\definecolor{tablegapgreen}{RGB}{208,237,216}
\definecolor{notepositive}{HTML}{176B5B}
\definecolor{notepositiveback}{HTML}{F5FAF8}
\definecolor{notenegative}{HTML}{C75461}
\definecolor{notenegativeback}{HTML}{FFF7F7}
\newcolumntype{Y}{>{\centering\arraybackslash}X}
\newcolumntype{L}{>{\raggedright\arraybackslash}X}

\tcbset{codecasebase/.style={
  enhanced,
  breakable,
  listing only,
  listing engine=listings,
  colback=codebackground,
  colframe=codeheaderblue,
  colbacktitle=codeheaderblue,
  coltitle=white,
  fonttitle=\bfseries,
  arc=1.8mm,
  boxrule=0.6pt,
  left=1.5mm,
  right=1.5mm,
  top=1.0mm,
  bottom=1.0mm,
  before skip=7pt,
  after skip=5pt
}}

\newtcblisting{codecasebox}[1]{
  codecasebase,
  title={#1},
  listing options={
    language=Python,
    basicstyle=\fontencoding{T1}\fontfamily{FiraMono-TLF}\fontsize{5.3}{6.9}\selectfont,
    keywordstyle=\color{codekeyword}\bfseries,
    commentstyle=\color{codecomment},
    stringstyle=\color{codestring},
    showstringspaces=false,
    upquote=true,
    columns=fullflexible,
    keepspaces=true,
    breaklines=true,
    breakatwhitespace=false,
    tabsize=4,
    aboveskip=0pt,
    belowskip=0pt
  }
}

\newtcblisting{exposedcodebox}[1]{
  codecasebase,
  title={#1},
  listing options={
    language=Python,
    basicstyle=\fontencoding{T1}\fontfamily{FiraMono-TLF}\fontsize{5.3}{6.9}\selectfont,
    keywordstyle=\color{codekeyword}\bfseries,
    commentstyle=\color{codecomment},
    stringstyle=\color{codestring},
    emph={_SCHEDULES,_STEP_PWC_WIN,_STEP_PWC_THR,_CHIRP_PHASE_W,
      _CHIRP_ENV_W,_SHORT_SIGNAL_THRESHOLD,_classify,kind,spec,
      _hilbert_chirp,_sin_demod,_step_pwc,out,zc_trend,zc_final,
      jump_frac,std_ratio,peaks,ac},
    emphstyle=\color{exposurecode}\bfseries,
    showstringspaces=false,
    upquote=true,
    columns=fullflexible,
    keepspaces=true,
    breaklines=true,
    breakatwhitespace=false,
    tabsize=4,
    aboveskip=0pt,
    belowskip=0pt
  }
}

\tcbset{notecardbase/.style={
  enhanced,
  coltitle=white,
  fonttitle=\fontencoding{T1}\fontfamily{FiraMono-TLF}\bfseries\fontsize{6.9}{9.6}\selectfont,
  fontupper=\fontencoding{T1}\fontfamily{FiraMono-TLF}\fontsize{6.3}{8.4}\selectfont,
  arc=1.6mm,
  boxrule=0.65pt,
  left=2.2mm,
  right=2.2mm,
  top=1.4mm,
  bottom=1.4mm,
  before skip=4.0pt,
  after skip=4.0pt,
  before upper={\raggedright},
}}

\newtcolorbox{positivenotecard}[2]{
  notecardbase,
  top=1.15mm,
  bottom=1.15mm,
  before skip=7pt,
  after skip=7pt,
  colframe=notepositive,
  colback=notepositiveback,
  colbacktitle=notepositive,
  title={#1\enspace #2}
}

\newtcolorbox{negativenotecard}[2]{
  notecardbase,
  before skip=7pt,
  after skip=7pt,
  colframe=notenegative,
  colback=notenegativeback,
  colbacktitle=notenegative,
  title={#1\enspace #2}
}

\definecolor{noteguidance}{HTML}{7B5EA7}
\definecolor{noteguidanceback}{HTML}{F7F5FB}
\newtcolorbox{guidancecard}[2]{
  notecardbase,
  colframe=noteguidance,
  colback=noteguidanceback,
  colbacktitle=noteguidance,
  title={#1\enspace #2}
}

\definecolor{promptgrey}{HTML}{555C58}
\definecolor{promptgreyback}{HTML}{F4F5F4}
\newtcolorbox{promptcard}[2]{
  notecardbase,
  breakable,
  colframe=promptgrey,
  colback=promptgreyback,
  colbacktitle=promptgrey,
  before upper={\raggedright\setlength{\parskip}{4.5pt}},
  title={#1\enspace{\mdseries (#2)}}
}

\definecolor{hookbrown}{HTML}{8A5A2B}
\definecolor{hookbrownback}{HTML}{FBF7F2}
\newtcolorbox{hookcard}[2]{
  notecardbase,
  breakable,
  colframe=hookbrown,
  colback=hookbrownback,
  colbacktitle=hookbrown,
  before upper={\raggedright\setlength{\parskip}{4.5pt}},
  title={#1\enspace #2}
}

\graphicspath{{pic/}}

\title{Where Autonomy Belongs: A Separation of Powers for Agentic Program Evolution}

\newcommand{\reportauthorblock}{%
  {\fontsize{10.5}{14}\selectfont\bfseries
  \mbox{Zhengyang~Ai\textsuperscript{1}}\quad
  \mbox{Tong~Liu\textsuperscript{1}}\quad
  \mbox{Jingxian~Tang\textsuperscript{1}}\quad
  \mbox{Xiaobing~Du\textsuperscript{1}}\quad
  \mbox{Tao~Xiao\textsuperscript{1}}\quad
  \mbox{Fengshi~Zhang\textsuperscript{1}}\quad
  \mbox{Zishuo~Li\textsuperscript{1}}\quad
  \mbox{Xiaodong~Ai\textsuperscript{1}}\quad
  \mbox{Pinyan~Lu\textsuperscript{1,2,\textdagger}}\par}
  \vspace{3pt}
  {\fontsize{9.5}{12}\selectfont
  \mbox{\textsuperscript{1}Huawei Taylor Lab}\quad
  \mbox{\textsuperscript{2}Shanghai University of Finance and Economics}\par}
  \vspace{2pt}
  {\fontsize{9}{11}\selectfont
  \textsuperscript{\textdagger}Corresponding author.\par}
}

\author{Zhengyang Ai, Tong Liu, Jingxian Tang, Xiaobing Du, Tao Xiao, Fengshi Zhang, Zishuo Li, Xiaodong Ai, Pinyan Lu}
\usepackage{evotrias-adaptevo}

\hypersetup{pdfauthor={Zhengyang Ai, Tong Liu, Jingxian Tang, Xiaobing Du, Tao Xiao, Fengshi Zhang, Zishuo Li, Xiaodong Ai, Pinyan Lu}}

\begin{document}

\fancyhead{}
\renewcommand{\headrulewidth}{0pt}
\fancyfoot[C]{\thepage}
\raggedbottom
\color{abyss}
  \thispagestyle{empty}
  \vspace*{-0.44in}
  \begin{tcolorbox}[
    width=\linewidth,
    colback=covercream,colframe=covercream,boxrule=0pt,
    arc=14pt,outer arc=14pt,boxsep=0pt,
    left=20pt,right=20pt,top=13pt,bottom=11pt
  ]
    {\outfit\fontsize{21.5}{25.5}\selectfont\bfseries\centering
      \textcolor{coveraccent}{Where Autonomy Belongs}%
      \textcolor{black}{: A Separation of Powers for Agentic Program Evolution}\par}
    \vspace{0.7em}
    {\centering\reportauthorblock\par}
    \vspace{0.75em}
    \begingroup
      \normalfont
      \setlength{\parindent}{0pt}\setlength{\parskip}{0pt}
      \renewenvironment{abstract}{\ignorespaces}{\par}
      \input{sections/0_abstract}
      \par
    \endgroup
    \vspace{0.65em}
    \noindent
    \begin{minipage}[b]{0.74\linewidth}
      \outfit\fontsize{8.4}{10.2}\selectfont
      \faGithub\hspace{0.4em}\textbf{Project \& Code:} \url{https://evotrias.github.io/}
    \end{minipage}
  \end{tcolorbox}

\input{sections/1_introduction}
\input{sections/2_related_work}
\input{sections/3_preliminaries}
\input{sections/4_method}
\input{sections/5_experiments}
\input{sections/6_analysis}
\input{sections/7_conclusion}
\input{sections/9_statements}

\bibliography{iclr2027_conference}
\bibliographystyle{iclr2027_conference}

\input{sections/8_appendix}

\end{document}

%% file: math_commands.tex
\usepackage{amsmath,amsfonts,bm}

\def\eqref#1{equation~\ref{#1}}

\def\1{\bm{1}}

\DeclareMathAlphabet{\mathsfit}{\encodingdefault}{\sfdefault}{m}{sl}
\SetMathAlphabet{\mathsfit}{bold}{\encodingdefault}{\sfdefault}{bx}{n}

\DeclareMathOperator*{\argmax}{arg\,max}

%% file: sections/0_abstract.tex
\begin{abstract}
LLM-driven program evolution rests on three decisions: govern directs the search from accumulated evidence, propose constructs the next candidate, and judge settles validity and fitness. All three appear in every system, and existing work splits into two lines over who makes each. Controller-centric systems strengthen govern and neglect the autonomy of propose, which caps what evolution can reach. Agent-centric systems let a coding agent take over both govern and propose, at a cognitive and computational cost. Neither line protects the independent authority of judge, so benchmark reliability suffers. These failures share one root: autonomy is misallocated across the three decisions. We address this problem with EvoTrias, a separation-of-powers architecture in which each decision has one owner: Structured Governance for what spans evolution rounds, Open Proposal for construction inside a task, and Protected Judgment behind a private evaluator. These owners exchange typed contracts only, and runtime hooks reinforce the boundaries. This separation also lets EvoTrias run as a plugin that turns an existing coding agent into an evolution system. EvoTrias sets a new state of the art on mathematical, systems, and algorithm-design benchmarks. Three further results show what each owner buys: separating Governance from Proposal cuts token cost by 64\%, Open Proposal extends evolution to a 680K-line industrial repository, and Protected Judgment keeps benchmark scores faithful to transfer.
\end{abstract}

%% file: sections/1_introduction.tex
\section{Introduction}
\label{sec:introduction}

\begin{figure}[t]
  \centering
  \includegraphics[width=\linewidth]{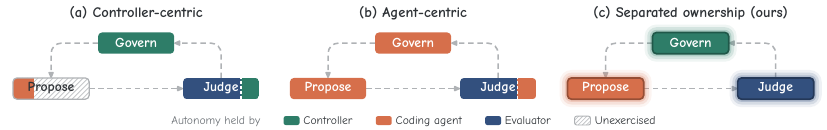}
  \vspace{-15pt}
  \caption{Who holds govern, propose, and judge in an LLM-driven evolutionary
  loop. \textbf{(a)} A controller holds govern,
   while propose is a single model call with little autonomy of its own. 
   \textbf{(b)} The coding agent holds both govern and propose. In
  \textbf{(a)} and \textbf{(b)} a dashed seam marks judge: the evaluator still
  returns the official score, but its sources are not private, so govern
  and propose erode its autonomy. \textbf{(c)} EvoTrias gives each decision one owner, and a runtime
  hook layer reinforces the split.}
  \label{fig:autonomy-allocation}
  \vspace{-1.4\baselineskip}
\end{figure}

LLM-driven program evolution repeatedly constructs, verifies, and retains
candidate programs under a finite evaluation budget. A public task contract
and an evaluator guide the search, which has produced results in mathematics,
algorithm design, and systems optimization
~\citep{romeraparedes2024funsearch,novikov2025alphaevolve,sharma2025openevolve}.
Its ceiling depends on three decisions. \emph{Govern} uses accumulated
evidence to direct the remaining search. \emph{Propose} constructs the next
candidate. \emph{Judge} settles whether a candidate is valid and how well it
scores. All three appear in every system. Existing work mainly studies how
to strengthen them, and by the decision it strengthens, it forms two lines:
controller-centric systems strengthen govern, and agent-centric systems
strengthen propose.

Controller-centric systems give govern to a controller, the evolutionary
algorithm that drives the loop. AdaEvolve, EvoX, and their successors
~\citep{cemri2026adaevolve,liu2026evox,du2026mlevolve,zhang2026aevo,xing2026compute} keep
asking how to build a better algorithm for assembling each iteration's
context. 
Propose stays a single LLM call inside that context and has little autonomy of its own. It sees only what the controller assembles, so harder problems, such as end-to-end optimization of an industrial code repository, stay out of reach.
The judge boundary weakens as well. Evaluator sources are not kept private, so govern and propose
can both learn how candidates are scored. A search that sees the evaluator's fixed test cases can fit them
instead of the task. The official score then overstates how well its programs transfer to new
instances.

Agent-centric systems move the other way. As coding agents grew more capable
of acting on their own, this line asked how much autonomy they could be given.
CORAL represents the direction~\mbox{\citep{qu2026coral}}. It replaces the single LLM
call with a coding agent that is both more autonomous and more capable, and it
hands govern to that agent as well. Subsequent work pushes the delegation
further
~\citep{mishra2026prism,virk2026swarmresearch}.
The agent that writes the candidates now
also reads the history, writes the shared memory, and plans the next move. Two
costs follow. The first is cognitive. One agent's lesson from a single attempt
goes straight into the memory every other agent reads. No independent owner
checks whether it holds, how far it should travel, or when it should expire,
so the search narrows before the evidence justifies it. The
second cost is computational. Long trajectories spend expensive model compute
on re-reading history, rebuilding search state, and planning the next move.
Judge fares no better. CORAL grants its agents unrestricted tool access and points them at the
evaluator source, which invites the same overfitting to the evaluator.

The two lines fail in mirror-image ways. Controller-centric systems strengthen
 govern but leave propose too little autonomy for harder tasks. Agent-centric
  systems give propose that autonomy, yet let it spill into govern, where no owner checks the lessons the agent writes. Neither 
  line keeps judge behind a boundary. These failures share one root: autonomy is misallocated across
    the three decisions. Each line raises the autonomy of one decision and neglects the other two, along with the boundaries between them. The 
    question is therefore not how much autonomy to grant, but where it belongs.
     We call this problem \emph{autonomy allocation}: \emph{who should own
     each of the three decisions, and what holds the boundary once ownership
     is assigned} (Figure~\ref{fig:autonomy-allocation}).

We answer with \textbf{EvoTrias}, a separation of powers that gives each
decision its own owner, called a \emph{plane}. \textbf{Structured Governance} handles everything that
spans evolution rounds. It distills settled outcomes into evidence that drives
 the selection policy, the durable experience, and the context each new task
 receives. \textbf{Open Proposal} owns open-ended implementation and artifact
construction inside an assigned task. A general-purpose coding agent, such as Claude Code or OpenCode, realizes
this plane. It pulls the task's context from Governance over MCP, then calls
tools and reasons on its own until an artifact exists. The artifact crosses a
narrow interface into \textbf{Protected Judgment}, which keeps official
evaluator evidence private and returns only a normalized outcome: validity,
fitness, and metrics. Under this division, each plane decides alone
inside its own boundary and reaches the others only through explicit contracts.
 A runtime hook layer reinforces the split, so no plane can erode the rights
 of another.
These contracts and hooks sit outside the coding agent, so EvoTrias leaves
the agent unchanged and runs as a plugin. The user's existing session
orchestrates native subagents, and the coding agent itself becomes an
evolution system inside the development workflow.

Our contributions are threefold.
\textbf{(1)}~We pose autonomy allocation for program evolution with coding
agents and measure it per decision: the agent's autonomy pays off in
propose (\S\ref{sec:repo-scale-analysis}), whereas extending it to govern
lowers the score and raises the cost
(\S\ref{sec:efficiency-analysis}, \S\ref{sec:ownership-ablation}), and
extending it to judge inflates the score (\S\ref{sec:leakage-analysis}).
\textbf{(2)}~We present EvoTrias, which keeps the agent's full autonomy over
construction inside a task and holds govern and judge outside the agent,
behind typed contracts and runtime hooks; it therefore runs as a plugin on
unmodified coding agents.
\textbf{(3)}~EvoTrias sets a new state of the art on Math, ADRS, and
Frontier-CS, improves the CP-SAT local search of the 680K-line OR-Tools by
up to 16.9\%, and runs at about 64\% lower token cost than four-agent CORAL.

%% file: sections/2_related_work.tex
\section{Related Work}
\label{sec:related-work}

\paragraph{LLM-guided program evolution.}
LLM-guided program evolution replaces hand-designed mutation operators with
semantic transformations selected by executable feedback. ELM introduced
learned code variation, and FunSearch, EoH, and ReEvo established the loop
around an LLM proposer and an
evaluator~\citep{lehman2022elm,romeraparedes2024funsearch,liu2024eoh,ye2024reevo}.
AlphaEvolve scales it to multi-component programs, and OpenEvolve provides an
open implementation~\citep{novikov2025alphaevolve,sharma2025openevolve}.  Later work splits into two lines that advance govern and propose.

\paragraph{Controller-centric evolution.}
One line refines the controller that allocates search effort. AdaEvolve
regulates exploration, cross-island resources, and tactic intervention from
improvement signals~\citep{cemri2026adaevolve}. EvoX evolves the executable
search strategy during a run, and ShinkaEvolve improves sample efficiency
through archive-aware sampling and novelty
control~\citep{liu2026evox,lange2025shinkaevolve}. Later systems widen the
adaptive surface to context, memory, strategy populations, cost, and the
evolution procedure
itself~\citep{yan2026pacevolve,du2026mlevolve,luo2026seaevo,jiang2026smcevolve,zhang2026costada,agrawal2025gepa,zhang2025ace,zhang2026aevo,xing2026compute}.
The effort concentrates on govern, the evolutionary algorithm itself, while the
capability of propose goes unaddressed and evaluator sources stay reachable
from the controller side.

\paragraph{Agent-centric evolution.}
Coding agents opened a second line. SWE-agent and OpenHands made
repository-level navigation, execution, and repair
routine~\citep{yang2024sweagent,wang2025openhands}. CORAL applies long-running
coding agents to parallel evolutionary lineages and coordinates them through
scored attempts, Notes, Skills, and heartbeat reflection~\citep{qu2026coral}.
Later systems push the delegation further through evolutionary memory and
global steering of branch-local
trajectories~\citep{mishra2026prism,virk2026swarmresearch,li2026reasearch}.
One agent owns govern and propose together here, and its open tool access
erodes the independent authority of judge.

\paragraph{EvoTrias.}
EvoTrias
 assigns one owner per decision, and this clean division of responsibilities
  lets it deploy as a plugin on mainstream coding agents (Appendix~\ref{app:host-integration}).

%% file: sections/3_preliminaries.tex
\section{Problem Setup}
\label{sec:problem-setup}

We study agentic program evolution under a finite proposal budget.
A task $\mathcal Q=(R_{\mathcal Q},C_{\mathcal Q},J_{\mathcal Q},B_{\mathcal Q})$
contains an initial repository, a public contract, an official evaluator, and
a hard budget $B_{\mathcal Q}$ on proposal slots. Each slot
returns at most one valid artifact $x_i$; evaluating that artifact yields
\begin{equation}
\label{eq:search-objective}
y_i=J_{\mathcal Q}(x_i)=(v_i,f_i,\mathbf m_i),\qquad
x^\star=\argmax_{\substack{x_i:\,v_i=1\\
                            i\in\mathcal J_{B_{\mathcal Q}}}} f_i,
\end{equation}
where $v_i$, $f_i$, and $\mathbf m_i$ denote validity, scalar fitness
(larger is better), and aggregate metrics, and $\mathcal J_{B_{\mathcal Q}}$
indexes the terminal artifacts returned by the admitted proposal slots.

%% file: sections/4_method.tex
\section{Method}
\label{sec:method}

\subsection{Three Decision Planes}
\label{sec:decision-planes}

\begin{figure}[t]
  \centering
  \includegraphics[width=\linewidth]{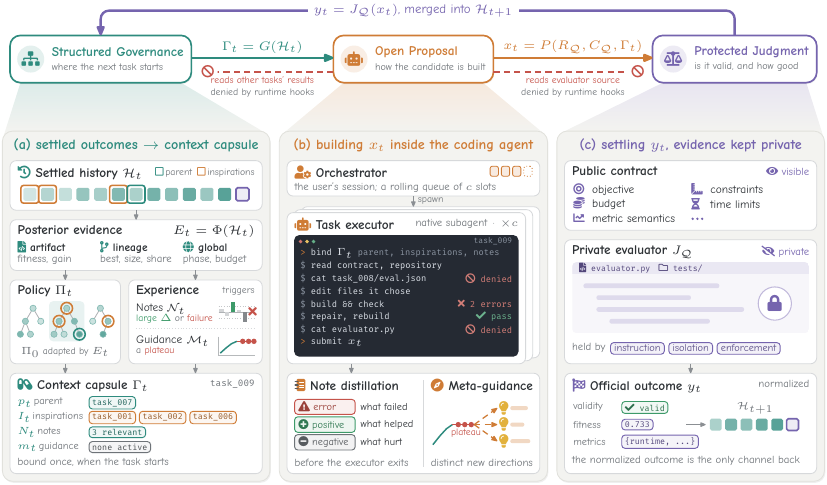}
  \caption{Overview of EvoTrias: Structured Governance
  (\S\ref{sec:structured-governance}) binds each task's context capsule,
  Open Proposal (\S\ref{sec:open-proposal}) constructs candidate artifacts,
  and Protected Judgment (\S\ref{sec:protected-judgment}) settles official
  outcomes that shape later bindings.}
  \label{fig:architecture}
    \vspace{-1.0\baselineskip}
\end{figure}

EvoTrias settles where autonomy belongs by giving each of the three decisions
its own plane (Figure~\ref{fig:architecture}). \textbf{Structured
Governance}, a Skill and
MCP control plane, owns govern: the persistent cross-trajectory state and the
context bound to every new task. \textbf{Open Proposal}, realized on a
general-purpose coding agent, owns propose, the construction of a candidate
inside that context. \textbf{Protected Judgment}, an isolated benchmark
evaluator, owns judge, the official validity and fitness. Each plane decides
alone inside its own boundary and holds no standing outside it. The planes
communicate only through typed contracts,
\begin{equation}
\label{eq:plane-contract}
\Gamma_t=G(\mathcal H_t),\qquad
x_t=P(R_{\mathcal Q},C_{\mathcal Q},\Gamma_t),\qquad
y_t=J_{\mathcal Q}(x_t),
\end{equation}
where Governance $G$ sends a context capsule $\Gamma_t$, Proposal $P$
submits a candidate artifact $x_t$, and Judgment $J_{\mathcal Q}$ returns a
normalized outcome $y_t$. Tasks run asynchronously. Here $t$ marks the logical
instant at which a new task binds its context, and $\mathcal H_t$ collects the
outcomes settled by then, so each settled outcome shapes only later bindings.
The hard budget terminates the run. These contracts fix the semantics of $G$
and $J_{\mathcal Q}$ independently of how $P$ is implemented, so any host
that meets them can supply candidate construction under the same allocation
of decision rights. Appendix~\ref{app:discussion} discusses what this
division costs and what it buys.

\subsection{Structured Governance}
\label{sec:structured-governance}

Structured Governance owns govern. For every new
task it selects a construction starting point: one parent candidate and a
set of inspiration candidates. When the search experience is worth
persisting, it also triggers two kinds of durable
experience, Notes and Meta-guidance. These decisions share one source, the
posterior evidence distilled from settled outcomes. The evidence feeds
three stages: policy construction decides how to select, experience
construction decides what to persist, and context construction decides
what the new task actually receives.

\paragraph{Posterior evidence distillation.}
Each official outcome $y_{t-1}$ merges into the settled history
$\mathcal H_t$, and the evidence operator $\Phi$ distills this history
into the current evidence
\begin{equation}
\label{eq:evidence}
E_t=\Phi(\mathcal H_t).
\end{equation}
$E_t$ organizes evidence at three scopes. Artifact-level fields describe a
single candidate: verified fitness, improvement over its parent, selection
count, and descendant potential. Lineage-level fields aggregate each
\emph{lineage}, the
descent subtree of a seeded root: best fitness, size, realized share of
the successful population, and lineage potential
(Appendix~\ref{app:potential}). Global-level fields
track the whole run: search phase, budget progress, and the recent
movement of the best fitness.

\paragraph{Policy construction.}
Governance maintains a default policy $\Pi_0$, adapted from
population-based program
evolution~\citep{sharma2025openevolve,cemri2026adaevolve}: select a
lineage, select a parent within it, then sample inspirations around it.
These form the three selection rules
$(\pi^L,\pi^P,\pi^I)$. Before each new task starts, evidence adjusts the
default into the current policy:
\begin{equation}
\label{eq:adapt}
\Pi_t=\operatorname{Adapt}(\Pi_0;E_t).
\end{equation}
Adapt makes four kinds of adjustment. \emph{Phase switching}: the root phase
assigns each task a distinct initial direction $d_i$ from
$\mathcal D_0=\{d_i\}_{i=1}^{n_{\mathrm{dir}}}$, with the parent and
inspiration branches closed. Once all $n_{\mathrm{dir}}$ roots settle, the run
enters the evolution phase and all three distributions take effect.
\emph{Progress annealing}: as the budget
depletes, selection temperatures and exploration coefficients tighten,
moving the search from broad to deep. \emph{Share rebalancing}: when the
entropy of lineage shares falls, the source
distribution of cross-lineage inspirations tilts toward the higher-quality
lineages within a fixed bound.
 \emph{Guidance anchoring}: a task assigned active
Meta-guidance switches its parent branch to the current canonical best, so
a new direction starts from the strongest point.

\paragraph{Experience construction.}
Search experience settles into two kinds of durable content, stored in the
notes pool $\mathcal N_t$ and the guidance pool $\mathcal M_t$. Notes
address single attempts. A note request fires when an attempt's relative
fitness delta clears an adaptive threshold, turns a failure into a
success, or ends in terminal failure. Governance sends the request to Open
Proposal, then validates the returned typed note and commits it to the
pool, and later tasks load it by relevance. Meta-guidance addresses the
whole search. When the best fitness stalls across a fixed window and
patience, Governance gathers the canonical best, the public contract,
lineage summaries, admitted notes, and the outcomes of earlier guidance.
It then asks Open Proposal for semantically distinct
directions. Each committed direction carries a bounded use count, and each
use is measured against the canonical best at the moment that use is
assigned.
Appendices~\ref{app:notes} and~\ref{app:guidance} give the full trigger
rules and replay both channels on representative
runs, and Appendix~\ref{app:channel-prevalence} audits them run by run.

\paragraph{Context construction.}
With the policy in place, Governance executes it and constructs the actual
context for the new task:
\begin{equation}
\label{eq:construct}
\Gamma_t=\operatorname{Construct}(\Pi_t;E_t,\mathcal N_t,\mathcal M_t)
=(p_t,I_t,N_t,m_t).
\end{equation}
Construction proceeds in order. The parent takes the highest-scoring
candidate,
\begin{equation}
\label{eq:parent}
p_t=\operatorname{argmax}_{i\in\ell_t}\bigl(q_i+U_{i,t}+\lambda_P\,g_i\bigr),
\end{equation}
where $\ell_t$ is the lineage drawn from $\pi^L$, which favors lineages
whose quality-deserved share exceeds their realized share; $q_i$ is
rank-normalized quality; $U_{i,t}$ is a UCB bonus that shrinks with
selection count; $g_i$ is descendant potential, and $\lambda_P$ bounds its
influence. A task assigned
active guidance skips this rule: $p_t$ anchors to the canonical best and
$m_t$ takes that guidance; otherwise $m_t$ is empty. Each inspiration
entry is drawn independently from a two-level mixture,
\begin{equation}
\label{eq:inspiration}
i\sim(1-\mu)\,\pi_{\mathrm{local}}(\cdot\mid\ell_t,T_t)
+\mu\sum_{k\ne\ell_t}\rho_t(k\mid E_t)\,\pi_{\mathrm{elite}}(\cdot\mid k).
\end{equation}
With probability $1-\mu$ it stays in the task's own lineage $\ell_t$,
where $\pi_{\mathrm{local}}$ is Boltzmann over quality at temperature
$T_t$. With probability $\mu$ it crosses: $\rho_t$ weights each donor
lineage $k$ by the elites it can supply and, as lineage shares grow
uneven, by its best fitness; $\pi_{\mathrm{elite}}$ then draws uniformly
from the top-$n_e$ candidates of $k$. This tilt in $\rho_t$ is the share
rebalancing above. Finally, $N_t$ loads the notes from
$\mathcal N_t$ most relevant to $p_t$ and $I_t$.
Appendix~\ref{app:routing} gives the full details, and Appendix
Table~\ref{tab:hyperparams} lists the default value of every scheduled
constant. The finished capsule crosses to Open Proposal
(\S\ref{sec:open-proposal}).

\subsection{Open Proposal}
\label{sec:open-proposal}

Open Proposal owns propose. Its host is a general-purpose coding agent that
supplies file editing, command execution, MCP calls, native subagents, and
hook points around tool calls and subagent lifecycle. A workflow skill
activates EvoTrias in the user's existing session and makes it the
orchestrator, which never touches task content and only spawns native
subagents under a fixed scheduling policy. These subagents produce the content Governance requests: candidates, Notes, and Meta-guidance. Every exchange runs over MCP into Structured Governance: the orchestrator asks only whether to continue and how many tasks to mint, and each task executor binds its own capsule.

\paragraph{Candidate proposal.}
At initialization, a dedicated subagent generates $\mathcal D_0$, with one
direction guiding the root-artifact construction of each lineage
(Appendix~\ref{app:init-directions}). The orchestrator executes the resulting
$n_{\mathrm{dir}}$ root proposals through a rolling queue with $c$ concurrent
slots. Each
slot spawns a task-executor subagent that binds its capsule
$\Gamma_t$ and studies the contract and repository. It then conducts the
construction trajectory: implement, build, run permitted development checks,
read failures, and repair within bounded attempts. Trajectory observations
decide tool order, edit locations, and the local plan; no mutation template
constrains the candidate. The subagent submits its artifact for official
evaluation and terminates. The freed slot is refilled immediately, so
settled outcomes from finished tasks reach Governance while sibling tasks
still run.

\paragraph{Note and guidance proposal.}
A note request comes back on the same control-plane call that records the
attempt, and the task executor distills that attempt into a typed note
before it terminates. An \emph{error} note names the
failure and its trigger. A \emph{positive} or \emph{negative} note records
which tactic helped or hurt, and under what conditions. A fixed analysis
prompt enforces condition-scoped summaries over universal rules, keeping one
attempt's experience from hardening into doctrine. A guidance request is
answered elsewhere, by a stateless meta actor whose job is to lift the search
out of the plateau that triggered the request. It reads the search's own
trace and returns a few directions that are semantically distinct from one
another and from what the search has already tried.

\paragraph{Runtime hooks.}
The hook layer spans all three planes, but its rules bear almost entirely on
Open Proposal. A coding agent's openness is what the design wants, and the
same openness puts two things at risk. The agent may drift from the
evolution workflow until the run's state no longer adds up. It may also
reach past the information the design granted it. Scheduler hooks address
the first risk and permission hooks the second.

Scheduler hooks maintain the run's task pool. The hook is its sole writer,
and every agent is a reader. It records each state change from a host event
only the runtime can emit: a spawn returning a handle, a subagent
starting or stopping. The orchestrator polls the pool to decide what has
finished and how many tasks to mint. Scheduling therefore follows what
happened, not what an agent reports.

Permission hooks run before every tool call. One rule family denies file
reads of evaluator sources and limits the shell to the evaluator command,
which keeps judge with Protected Judgment alone. A second denies reads of
another task's results. A
task therefore works from exactly the capsule Governance bound to it, which
keeps govern with Structured Governance alone. A thin adapter maps each
host's native events onto this layer, which Appendix~\ref{app:hooks} details.

\subsection{Protected Judgment}
\label{sec:protected-judgment}

None of the compared systems protects the independent
authority of judge. Each keeps a separate evaluator with non-private
sources, so govern and propose can erode judge; AdaEvolve, EvoX, and CORAL
direct their components to read them (Appendix~\ref{app:judge-baselines}).
Protected Judgment closes this gap through a contract-visible,
evidence-private split. The
public contract exposes the objective, constraints, metric semantics, time
limits, and budget. The evaluator stays private. The artifact $x_t$ enters
through a narrow interface, and the response returns only the normalized
outcome $y_t$.

Three measures keep evaluator sources out of reach. The first is
instruction: every proposal prompt states the prohibition
(Appendix~\ref{app:prompts}). The second is isolation: where the
benchmark allows, an isolated service holds the evaluator behind a
request interface. The third is
enforcement: the permission hooks of \S\ref{sec:open-proposal} deny
evaluator reads at the tool level. The
settled outcome then merges into $\mathcal H_{t+1}$ and shapes only contexts
bound afterwards, closing the search loop.

%% file: sections/5_experiments.tex
\section{Experiments}
\label{sec:experiments}

\subsection{Experimental Setup}
\label{sec:exp-setup}

We evaluate on three benchmarks:
Math~\citep{novikov2025alphaevolve},
ADRS~\citep{cheng2025adrs}, and
Frontier-CS~\citep{mang2025frontiercs}. We compare
against OpenEvolve~\citep{sharma2025openevolve}, EvoX~\citep{liu2026evox},
AdaEvolve~\citep{cemri2026adaevolve}, and CORAL~\citep{qu2026coral} in one-
and four-agent configurations. On Math and ADRS we run two backbones,
GLM-5.2 and Opus-5, with three runs per problem, and on Frontier-CS
GLM-5.2 once on the final 30 problems (compute constraints preclude
repeats). All iteration-based
methods receive 50 proposal slots and the same number of evaluator
calls (Table~\ref{tab:evaluator-calls}); CORAL is instead matched by wall
clock, with CORAL-4 reporting the best of four complete top-level agents
against EvoTrias's one (Appendix~\ref{app:coral-budget}). Main EvoTrias experiments use
$n_{\mathrm{dir}}=4$ initial directions and $c=4$ concurrent task-executor
slots; \S\ref{sec:concurrency-analysis} analyzes $c$. EvoTrias and CORAL run on Claude Code with web search disabled.
Appendix~\ref{app:experiment-details} gives details.

\subsection{Main Results}
\label{sec:main-results}

Table~\ref{tab:combined-mean-results} reports the Math and ADRS results.
Under each of the two backbones, EvoTrias ranks first on nine of the twelve
tasks. The largest margins arise where the search still has headroom: under
GLM-5.2, EvoTrias leads the strongest baseline by 6.3\% on Heilbronn Convex
and by 3.3\% on Signal Processing. The five ties under Opus-5 all fall on
saturated tasks, where the leading methods return the same score to the last
reported digit. Appendix~\ref{app:mean-best-results} adds best-run and
standard-deviation results and a paired task-level analysis
(Table~\ref{tab:math-adrs-paired-stats}): EvoTrias improves on OpenEvolve,
EvoX, AdaEvolve, and CORAL-1 on all 12 tasks, with every confidence interval
positive and Holm-corrected $p=0.0024$, and leads CORAL-4 on 9 of 12.

\begin{figure}[t]
  \centering
  \begin{minipage}[t]{0.49\textwidth}
    \centering
    \vspace{0pt}
    \includegraphics[width=\linewidth]{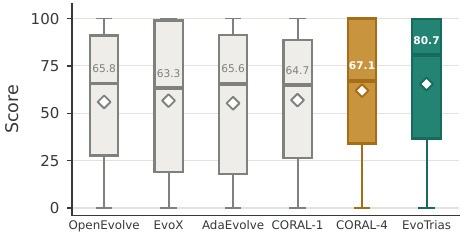}
    \vspace{1.7pt}
    \vspace{-0.75em}
    \captionof{figure}{Frontier-CS score distributions.}
    \label{fig:frontier-dist}
  \end{minipage}\hfill
  \begin{minipage}[t]{0.49\textwidth}
    \centering
    \vspace{0pt}
    \includegraphics[width=\linewidth]{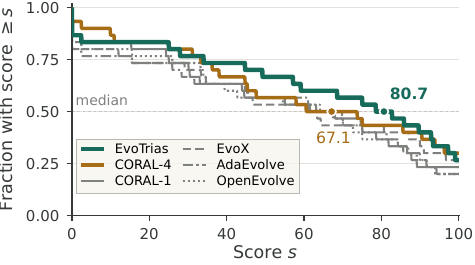}
    \vspace{-9.7pt}
    \vspace{-0.75em}
    \captionof{figure}{Frontier-CS survival curves.}
    \label{fig:frontier-ecdf}
  \end{minipage}
  \vspace{-1.5\baselineskip}
\end{figure}

Figure~\ref{fig:frontier-dist} reports the Frontier-CS score distributions.
EvoTrias obtains the best mean and, by a wide margin, the best median: 80.67
against 67.14 for CORAL-4. Figure~\ref{fig:frontier-ecdf} shows the same
outcomes as survival curves, where each curve reports the fraction of problems
with score at least $s$; from the middle of the score range to the mid-90s,
EvoTrias's curve lies above every baseline.
Appendix~\ref{app:frontier-analysis} lists per-problem scores and a paired
problem-level analysis that confirms
this lead: EvoTrias's mean gain is positive against every baseline, with
positive confidence intervals against four of the five.

\begin{table}[t]
  \centering
  \caption{Mean results on Math and ADRS. Best results are
  highlighted as \colorbox{resultfirst}{first} and
  \colorbox{resultsecond}{second}.}
  \label{tab:combined-mean-results}
  \vspace{-5pt}
  \scriptsize
  \setlength{\tabcolsep}{0.70pt}
  \renewcommand{\arraystretch}{1.12}
  \begin{tabular*}{\textwidth}{@{\extracolsep{\fill}}l r r r r r r
    @{\hspace{3.4pt}\vrule width 0.35pt\hspace{1.0pt}}r r r r r r@{}}
    \toprule
    & \multicolumn{6}{c}{GLM-5.2}
    & \multicolumn{6}{c}{Opus-5} \\
    \cmidrule(lr){2-7}\cmidrule(lr){8-13}
    Task
    & OpenE. & EvoX & AdaE. & CORAL-1 & CORAL-4 & EvoTrias
    & OpenE. & EvoX & AdaE. & CORAL-1 & CORAL-4 & EvoTrias \\
    \midrule
    \rowcolor{backbonegray}
    \multicolumn{13}{@{}l}{\textit{Math}} \\
    Circle-Pack. (Sq.)
    & 2.54317 & 2.60956 & 2.63152
    & 2.62337 & \cellcolor{resultsecond}2.63418
    & \cellcolor{resultfirst}2.63484
    & \cellcolor{resultsecond}2.63598 & \cellcolor{resultsecond}2.63598
    & \cellcolor{resultsecond}2.63598 & \cellcolor{resultsecond}2.63598
    & \cellcolor{resultsecond}2.63598
    & \cellcolor{resultfirst}2.63599 \\
    Circle-Pack. (Rect.)
    & 2.35930 & 2.34876 & \cellcolor{resultsecond}2.36487
    & 2.32097 & 2.36459
    & \cellcolor{resultfirst}2.36493
    & \cellcolor{resultsecond}2.36583 & \cellcolor{resultsecond}2.36583
    & \cellcolor{resultfirst}2.36584
    & \cellcolor{resultsecond}2.36583 & \cellcolor{resultsecond}2.36583
    & \cellcolor{resultfirst}2.36584 \\
    Heilbronn-Tri.
    & 0.0253 & 0.0302 & 0.0318
    & 0.0320 & \cellcolor{resultsecond}0.0332
    & \cellcolor{resultfirst}0.0335
    & 0.0357 & 0.0361 & \cellcolor{resultsecond}0.0364
    & \cellcolor{resultsecond}0.0364
    & \cellcolor{resultfirst}0.0365
    & \cellcolor{resultfirst}0.0365 \\
    Heilbronn-Conv.
    & 0.0250 & 0.0269 & 0.0275
    & \cellcolor{resultsecond}0.0284 & 0.0283
    & \cellcolor{resultfirst}0.0302
    & 0.0290
    & \cellcolor{resultsecond}0.0307
    & 0.0304
    & \cellcolor{resultfirst}0.0309
    & \cellcolor{resultfirst}0.0309
    & \cellcolor{resultfirst}0.0309 \\
    MinMax-Dist.
    & 0.23937 & 0.23970 & 0.23987
    & 0.23975 & \cellcolor{resultsecond}0.24003
    & \cellcolor{resultfirst}0.24005
    & \cellcolor{resultfirst}0.24005
    & \cellcolor{resultfirst}0.24005
    & \cellcolor{resultfirst}0.24005
    & \cellcolor{resultfirst}0.24005
    & \cellcolor{resultfirst}0.24005
    & \cellcolor{resultfirst}0.24005 \\
    Signal-Proc.
    & 0.6373 & \cellcolor{resultsecond}0.7098 & 0.6961
    & 0.6638 & 0.7083
    & \cellcolor{resultfirst}0.7329
    & 0.6930 & 0.7861 & 0.7976
    & 0.8251
    & \cellcolor{resultsecond}0.8280
    & \cellcolor{resultfirst}0.8313 \\
    Third-Autocorr. $\downarrow$
    & 1.4683 & \cellcolor{resultfirst}1.4601
    & 1.4614 & 1.4814 & \cellcolor{resultsecond}1.4608
    & 1.4609
    & 1.4832 & 1.4587 & 1.4568
    & 1.4572 & \cellcolor{resultfirst}1.4559
    & \cellcolor{resultsecond}1.4563 \\
    \midrule
    \rowcolor{backbonegray}
    \multicolumn{13}{@{}l}{\textit{ADRS}} \\
    Cloudcast $\downarrow$
    & 664.114 & 646.304 & 631.130 & 622.570
    & \cellcolor{resultsecond}618.0869
    & \cellcolor{resultfirst}618.0671
    & 618.069 & \cellcolor{resultfirst}618.0603
    & 618.071 & 618.066 & 618.068
    & \cellcolor{resultsecond}618.0648 \\
    EPLB
    & 0.1278 & 0.1328 & 0.143327 & 0.1423
    & \cellcolor{resultfirst}0.1437
    & \cellcolor{resultsecond}0.143332
    & 0.1436 & 0.1437 & 0.1439 & 0.1445
    & \cellcolor{resultfirst}0.1450
    & \cellcolor{resultsecond}0.1448 \\
    LLM-SQL
    & 0.6942 & 0.6794 & \cellcolor{resultsecond}0.7148
    & 0.6742 & 0.6890 & \cellcolor{resultfirst}0.7206
    & 0.7162 & 0.7229 & 0.7232 & 0.7240
    & \cellcolor{resultsecond}0.7318
    & \cellcolor{resultfirst}0.7398 \\
    Prism
    & 26.213 & 26.246 & 25.359
    & \cellcolor{resultsecond}26.252 & 26.248
    & \cellcolor{resultfirst}26.256
    & \cellcolor{resultfirst}26.256
    & \cellcolor{resultfirst}26.256
    & \cellcolor{resultfirst}26.256
    & \cellcolor{resultfirst}26.256
    & \cellcolor{resultfirst}26.256
    & \cellcolor{resultfirst}26.256 \\
    TXN
    & 3{,}700 & 3{,}843 & 3{,}928 & 4{,}078
    & \cellcolor{resultfirst}4{,}280
    & \cellcolor{resultsecond}4{,}203
    & 3{,}920 & 4{,}281 & 4{,}292 & 4{,}498
    & \cellcolor{resultsecond}4{,}532
    & \cellcolor{resultfirst}4{,}573 \\
    \bottomrule
  \end{tabular*}
  \vspace{-0.8\baselineskip}
\end{table}

\Needspace{0.30\textheight}
\vspace{-4pt}
\subsection{Ablation Study}
\label{sec:ablation}

\begin{wraptable}{r}{0.40\textwidth}
  \vspace{-1.0\baselineskip}
  \centering
  \captionsetup{hypcap=false,justification=raggedright,singlelinecheck=false}
  \captionof{table}{Ablation study results.}
  \label{tab:ablation}
  \vspace{-2pt}
  \scriptsize
  \setlength{\tabcolsep}{2.0pt}
  \renewcommand{\arraystretch}{0.94}
  \newcommand{\abstd}[1]{{\fontsize{4.6}{5.0}\selectfont\color{black!70}$\pm$#1}}
  \scalebox{0.95}{%
\begin{tabularx}{1.05263\linewidth}{@{}
    >{\hsize=1.45\hsize\linewidth=\hsize\fontsize{6.7}{7.0}\selectfont\raggedright\arraybackslash}X
    >{\hsize=0.775\hsize\linewidth=\hsize\fontsize{6.2}{6.5}\selectfont\centering\arraybackslash}X
    >{\hsize=0.775\hsize\linewidth=\hsize\fontsize{6.2}{6.5}\selectfont\centering\arraybackslash}X
    @{\hspace{\tabcolsep}}}
    \toprule
    Configuration & \shortstack{Circle\\Packing} & \shortstack{Signal\\Processing} \\
    \midrule
    \mbox{\textbf{Full EvoTrias}}
      & \textbf{2.6348}\abstd{0.0019} & \textbf{0.7329}\abstd{0.0058} \\
    \midrule
    \rowcolor{backbonegray}
    \multicolumn{3}{l}{\textbf{\textit{Level 1: Governance and initialization}}} \\
    \mbox{w/o Notes}
      & \cellcolor[RGB]{253,232,233}2.6122\abstd{0.0264}
      & \cellcolor[RGB]{251,207,209}0.7193\abstd{0.0281} \\
    \mbox{w/o Meta-guidance}
      & \cellcolor[RGB]{253,232,233}2.6087\abstd{0.0102}
      & \cellcolor[RGB]{248,174,177}0.7041\abstd{0.0094} \\
    \mbox{w/o adaptation ($\Pi_0$)}
      & \cellcolor[RGB]{253,232,233}2.6267\abstd{0.0079}
      & \cellcolor[RGB]{244,125,129}0.6926\abstd{0.0132} \\
    \mbox{Initial directions $n_{\mathrm{dir}}=1$}
      & \cellcolor[RGB]{251,207,209}2.5969\abstd{0.0448}
      & \cellcolor[RGB]{244,125,129}0.6763\abstd{0.0554} \\
    \mbox{Initial directions $n_{\mathrm{dir}}=2$}
      & \cellcolor[RGB]{253,232,233}2.6310\abstd{0.0067}
      & \cellcolor[RGB]{251,207,209}0.7167\abstd{0.0190} \\
    \midrule
    \rowcolor{backbonegray}
    \multicolumn{3}{l}{\textbf{\textit{Level 2: Host and model portability}}} \\
    \mbox{ClaudeCode + DS-V4-Flash}
      & \cellcolor[RGB]{253,232,233}2.6241\abstd{0.0126}
      & \cellcolor[RGB]{248,174,177}0.7104\abstd{0.0117} \\
    \mbox{OpenCode + GLM-5.2}
      & 2.6325\abstd{0.0016}
      & \cellcolor[RGB]{199,235,208}0.7383\abstd{0.0240} \\
    \mbox{OpenCode + DS-V4-Flash}
      & \cellcolor[RGB]{253,232,233}2.6240\abstd{0.0126}
      & \cellcolor[RGB]{248,174,177}0.7094\abstd{0.0208} \\
    \mbox{OpenCode + Opus-5}
      & 2.6360\abstd{0.0000}
      & \cellcolor[RGB]{126,211,147}0.8145\abstd{0.0153} \\
    \bottomrule
  \end{tabularx}%
}%
  \vspace{-1.2\baselineskip}
\end{wraptable}

Table~\ref{tab:ablation} lists Structured Governance and host/model portability
variants; $\Pi_0$ is the default policy from
Section~\ref{sec:structured-governance}, and each cell is mean fitness over three runs.

The Governance ablations stay close to the baseline on Circle Packing but
separate more clearly on Signal Processing, where a single initial direction
costs 7.7\%, the default policy $\Pi_0$ 5.5\%, and removing Meta-guidance
3.9\%. This matches their task scope:
Signal Processing optimizes a filtering algorithm over a family of signals,
whereas Circle Packing targets one fixed 26-circle arrangement, so the
broader task gains more from Meta-guidance, adaptation, and initial
search breadth. Elevated standard deviations after removing Notes or setting
$n_{\mathrm{dir}}=1$ show that Notes and directional diversity stabilize
evolution; a single direction anchors search to the lineage root, amplifying
root-dependent variation. Appendix~\ref{app:agentic-baselines} adds an
agentic proposer to the controller-centric baselines; none of the resulting
configurations reaches EvoTrias.

The host and model ablations separate two sources of variation. With
GLM-5.2 fixed, Claude Code and OpenCode yield mean scores within 0.8\% of
each other on both tasks; changing the LLM under OpenCode produces a wider
spread, and Opus-5 lifts Signal Processing by 10\%. EvoTrias is thus less
sensitive to the host than to
the model: the Proposal host can be replaced without a pronounced loss
(Appendix~\ref{app:host-integration}), and construction quality still
benefits from stronger LLMs.

%% file: sections/6_analysis.tex
\section{Analysis}
\label{sec:analysis}

\subsection{Repository-Scale Proposal}
\label{sec:repo-scale-analysis}

\begin{wrapfigure}{r}{0.42\textwidth}
  \vspace{-0.75\baselineskip}
  \centering
  \captionsetup{hypcap=false,skip=2pt}
  \includegraphics[width=\linewidth]{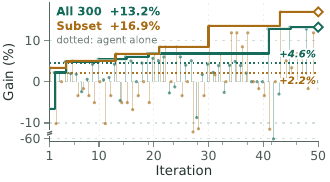}
  \captionof{figure}{Gain over the unpatched solver; dotted lines:
  the agent iterating alone.}
  \label{fig:ortools-gain}
  \vspace{-1.0\baselineskip}
\end{wrapfigure}

Beyond the scores of \S\ref{sec:main-results}, Open Proposal holds a second
advantage over controller-centric systems: the kind of task it can take on.
The benchmarks of \S\ref{sec:exp-setup} hand the proposer one file behind a
fixed interface, so the only open question is what to write inside it. In
contrast, this section evolves the CP-SAT local search of
OR-Tools 9.15~\citep{perron2024ortools}, a C++ repository of about 680K
lines, scored by end-to-end solver quality on 300 MiniZinc instances. The
repository marks no region to evolve, and solver quality comes from code
spread over several functions and files. The proposer must first read the
repository, trace how the local search works, and choose the functions to
edit. The single LLM call that controller-centric systems use for proposal
cannot do this, because a 680K-line repository far exceeds its context
window.

EvoTrias gives Open Proposal only the objective and a broad edit surface. Open Proposal does the rest: it retrieves and reads the relevant
code, selects its change sites, keeps cross-function invariants intact, and
exports a Git patch that applies to the untouched source tree. The 300
instances come from the MiniZinc Challenge 2023--2025, and each candidate is
scored against the unpatched solver by the Challenge's Borda-count rule;
Appendix~\ref{app:ortools} details the settings. Figure~\ref{fig:ortools-gain}
reports the gain over 50 iterations on all 300 instances and on a 60-instance
subset. Both rise steadily, ending at $+13.2\%$ and $+16.9\%$ over the
unpatched solver. The same agent refining its own patch for 50 rounds
without EvoTrias, under the same configuration and evaluator, reaches
$+4.6\%$ and $+2.2\%$ (dotted lines).

\subsection{Runtime and Token Efficiency}
\label{sec:efficiency-analysis}

\begin{figure}[!htbp]
  \vspace{-2pt}
  \centering
  \captionsetup{skip=2pt}
  \begin{subfigure}[h]{0.47\textwidth}
    \centering
    \includegraphics[width=\linewidth]{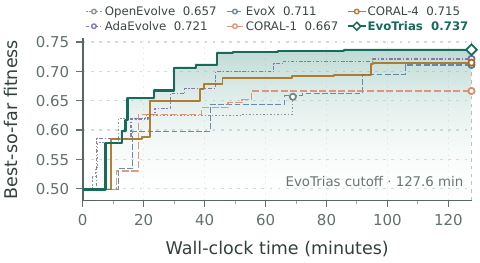}
    \caption{Best-so-far fitness over wall-clock time.}
    \label{fig:sp-wallclock}
  \end{subfigure}\hfill
  \begin{subfigure}[h]{0.51\textwidth}
    \centering
    \includegraphics[width=\linewidth]{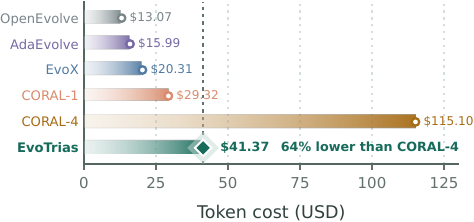}
    \caption{Mean token cost of a full run.}
    \label{fig:sp-tokens}
  \end{subfigure}
  \caption{Runtime and token efficiency with GLM-5.2. (a) Signal Processing, best of three runs, cut off at EvoTrias's 50-iteration runtime. (b) Average over the seven Math tasks, budgets as in \S\ref{sec:exp-setup}.}
  \label{fig:sp-efficiency}
\end{figure}

Figure~\ref{fig:sp-efficiency} measures the computational cost of agent-centric evolution (\S\ref{sec:introduction}). On Signal Processing, in
the best run of each method,
EvoTrias holds the best fitness from about 15 minutes onward and ends
at 0.737, while CORAL-4 consumes the full matched budget to reach 0.715
(Figure~\ref{fig:sp-wallclock}); only OpenEvolve completes its 50
iterations before the cutoff. The token account tells the same story across
all seven Math tasks (Figure~\ref{fig:sp-tokens}): a full 50-iteration
EvoTrias run costs \$41.37 on average, 64\% below the
\$115.10 of CORAL-4.

On time, the advantage of EvoTrias and OpenEvolve comes mainly from
parallel task execution; \S\ref{sec:concurrency-analysis} isolates this
trade-off. On tokens, the three controller-centric systems spend
the least, and both agent-centric systems pay for coding-agent
trajectories, but
EvoTrias keeps the bill bounded. Widening CORAL from one agent to four multiplies its bill by 3.9$\times$, whereas EvoTrias's cost varies by less than 5\% from $c=1$ to $c=8$. Leading six of seven Math tasks
(Table~\ref{tab:combined-mean-results}) at about a third of CORAL-4's token
cost
supports the allocation claim: Structured
Governance moves history interpretation and search routing out of the
expensive agent trajectory, so agent compute concentrates on candidate
construction.

\paragraph{Matched token budget.}
To further compare the methods under the same token budget, we give
every baseline the mean token budget of a 50-iteration EvoTrias run on the
same task; Table~\ref{tab:token-matched} shows the results. This buys the
controller-centric baselines 102--158 iterations, 2.0--3.2 times the
proposals and evaluator calls of EvoTrias, and stops CORAL-4 after
0.68--0.79 hours. EvoTrias keeps the highest mean on both tasks, and both
CORAL configurations stay below it with the same tokens: among systems built
on coding agents, token efficiency is a direct advantage of EvoTrias.

\begin{table}[t]
  \centering
  \caption{Token-matched comparison (GLM-5.2, three runs). Each baseline
  receives the mean token budget of a 50-iteration EvoTrias run on the same
  task. Budget: the iterations this buys, or wall-clock hours for CORAL.
  $\Delta$: change from the same method's mean in
  Table~\ref{tab:combined-mean-results}.}
  \label{tab:token-matched}
  \vspace{-5pt}
  \scriptsize
  \setlength{\tabcolsep}{2.0pt}
  \renewcommand{\arraystretch}{1.12}
  \newcommand{\tmstd}[1]{{\fontsize{4.8}{5.2}\selectfont\color{black!70}$\pm$#1}}
  \begin{tabular*}{\textwidth}{@{\extracolsep{\fill}}l r r r
    @{\hspace{6pt}\vrule width 0.35pt\hspace{6pt}}r r r@{}}
    \toprule
    & \multicolumn{3}{c}{Circle Packing}
    & \multicolumn{3}{c}{Signal Processing} \\
    \cmidrule(lr){2-4}\cmidrule(lr){5-7}
    Method & Budget & Score & $\Delta$ & Budget & Score & $\Delta$ \\
    \midrule
    OpenEvolve & 158 iter. & 2.5940\tmstd{0.0265} & $+0.0508$
      & 153 iter. & 0.6638\tmstd{0.0223} & $+0.0265$ \\
    EvoX       & 102 iter. & \cellcolor{resultsecond}2.6297\tmstd{0.0064} & $+0.0201$
      & 110 iter. & 0.7057\tmstd{0.0120} & $-0.0041$ \\
    AdaEvolve  & 129 iter. & 2.6277\tmstd{0.0039} & $-0.0038$
      & 131 iter. & \cellcolor{resultsecond}0.7273\tmstd{0.0042} & $+0.0312$ \\
    CORAL-1    & 2.66 h & 2.6266\tmstd{0.0064} & $+0.0033$
      & 3.04 h & 0.6781\tmstd{0.0593} & $+0.0143$ \\
    CORAL-4    & 0.68 h & 2.6216\tmstd{0.0204} & $-0.0126$
      & 0.79 h & 0.6894\tmstd{0.0134} & $-0.0189$ \\
    \midrule
    EvoTrias   & 50 iter. & \cellcolor{resultfirst}2.6348\tmstd{0.0019} & --
      & 50 iter. & \cellcolor{resultfirst}0.7329\tmstd{0.0058} & -- \\
    \bottomrule
  \end{tabular*}
  \vspace{-0.6\baselineskip}
\end{table}

\begin{figure}[t]
  \centering
  \includegraphics[width=\linewidth]{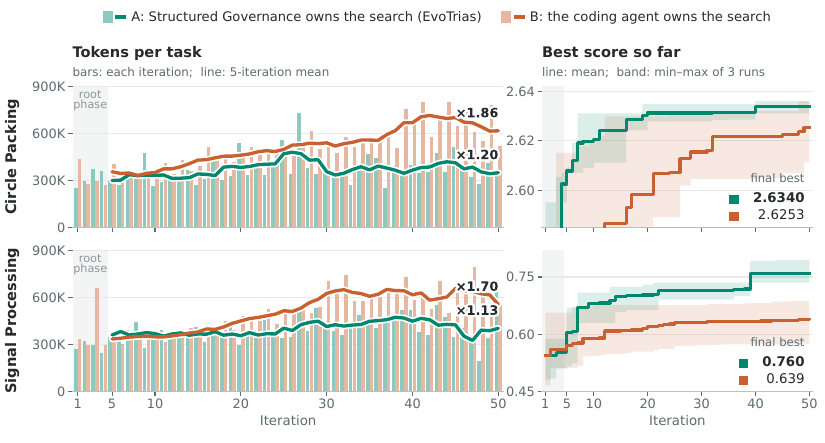}
  \vspace{-14pt}
  \caption{Controlled ownership ablation (three runs).
  \textbf{Left:} tokens per task (bars) and 5-iteration means (lines);
  ratios: iterations 40--50 / 5--15.
  \textbf{Right:} mean best-so-far score, with min--max bands.}
  \label{fig:ownership}
  \vspace{-0.9\baselineskip}
\end{figure}

\subsection{Who Owns the Search: A Controlled Ablation}
\label{sec:ownership-ablation}

To isolate the separation of govern from propose, we compare two
configurations that differ only in who makes the search decisions: the
parent and inspirations of each task, the lessons that enter the shared
notes, and when to change direction. In configuration~A (EvoTrias),
Structured Governance makes them. In configuration~B, the coding agent that
constructs each candidate makes them itself, with the full run history
available. The agent, model, construction instructions, protected evaluator,
and budget are identical, and each configuration runs three times on Circle
Packing and Signal Processing (Appendix~\ref{app:ownership-ablation}).

As Figure~\ref{fig:ownership} shows, A scores higher on both tasks:
2.6340 against 2.6253 on Circle Packing, which is close to saturation, and
0.760 against 0.639 on Signal Processing. B also costs more: its agent
rereads a history that lengthens every iteration, so its tokens per task grow
by a factor of 1.70--1.86 over a run, against 1.13--1.20 for A, and a B run
spends 19--28\% more tokens. With everything else held constant, separating
govern from propose raises the score and lowers the cost.

\FloatBarrier
\Needspace{0.48\textheight}
\subsection{Evaluator Exposure and Overfitting}
\label{sec:leakage-analysis}

\begin{wrapfigure}{r}{0.42\textwidth}
  \vspace{-0.75\baselineskip}
  \centering
  \captionsetup{hypcap=false}
  \captionof{table}{Evaluator exposure on Signal Processing
  (Gap: OOD $-$ Original).}
  \label{tab:evaluator-exposure}
  \vspace{1.5pt}
  \scriptsize
  \setlength{\tabcolsep}{1.5pt}
  \renewcommand{\arraystretch}{0.95}
  \begin{tabular*}{\linewidth}{@{\extracolsep{\fill}}llcccr@{}}
    \toprule
    \multirow{2}{*}{Method} & \multirow{2}{*}{Evaluator}
      & \multicolumn{3}{c}{Score $\uparrow$}
      & \multicolumn{1}{r}{\multirow{2}{*}{Gap $\uparrow$}} \\
    \cmidrule(lr){3-5}
      & & Original & IID & OOD & \\
    \midrule
    \multirow{2}{*}{CORAL-1}
      & Exposed   & 0.793 & 0.750 & 0.705 & \cellcolor[RGB]{244,125,129}$-$0.089 \\
      & Protected & 0.664 & 0.646 & 0.639 & \cellcolor[RGB]{253,222,223}$-$0.025 \\
    \midrule
    \multirow{2}{*}{CORAL-4}
      & Exposed   & 0.789 & 0.756 & 0.711 & \cellcolor[RGB]{247,148,151}$-$0.078 \\
      & Protected & 0.708 & 0.716 & 0.731 & \cellcolor[RGB]{157,219,172}0.023 \\
    \midrule
    \multirow{2}{*}{EvoTrias}
      & Exposed   & 0.803 & 0.745 & 0.721 & \cellcolor[RGB]{246,138,142}$-$0.083 \\
      & Protected & 0.733 & 0.732 & 0.741 & \cellcolor[RGB]{224,243,229}0.007 \\
    \bottomrule
  \end{tabular*}

  \vspace{0.85em}
  \includegraphics[width=\linewidth]{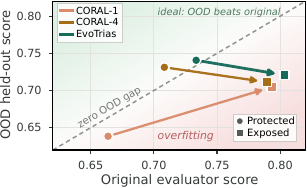}
  \vspace{-1.5em}
  \captionsetup{hypcap=false}
  \captionof{figure}{Generalization shift from protected to exposed
  evaluators.}
  \label{fig:evaluator-shift-map}
  \vspace{-1.5\baselineskip}
\end{wrapfigure}

This section measures what the judge boundary of
\S\ref{sec:protected-judgment} buys. For task-family benchmarks such as
Signal Processing, evaluator source reveals
both problem-level guidance (the task definition and scoring rule) and
instance-level information (fixed test cases). The former can improve
algorithms; we test whether the latter encourages overfitting and distorts
performance on unseen instances. We compare exposed and protected settings for
CORAL-1, CORAL-4, and EvoTrias, with three runs per configuration. The exposed
setting removes Protected Judgment, so this is an ablation of the third plane.
Programs are
evaluated on Original, IID held-out (new instances of the same families), and
OOD held-out (new signal and noise distributions). Appendix~\ref{app:evaluator-exposure}
details the protocol.

Table~\ref{tab:evaluator-exposure} shows that exposure raises the Original
score by 0.093 on average, but this advantage contracts to 0.009 OOD. CORAL-1
retains a 0.066 OOD gain, whereas CORAL-4 and EvoTrias each incur a 0.020 loss;
their protected programs also score higher OOD than on Original.
Figure~\ref{fig:evaluator-shift-map} summarizes how each method's Original and
OOD performance changes as evaluator access shifts from protected to exposed.
CORAL-4 and EvoTrias cross the zero-gap
diagonal. Thus, exposure can aid task and score
understanding, but visible cases can inflate benchmark gains and reduce OOD
performance. Protecting evaluator code preserves the Original score as a more
reliable estimate of transfer.  On Circle Packing, whose evaluator scores the one specified arrangement, exposure shifts the score by less than 1\%. Appendix~\ref{app:evaluator-code-cases} provides
code-level examples.

\subsection{Concurrency Trade-off}
\label{sec:concurrency-analysis}

\begin{wrapfigure}{r}{0.35\textwidth}
  \vspace{-0.75\baselineskip}
  \centering
  \captionsetup{hypcap=false}
  \includegraphics[width=\linewidth]{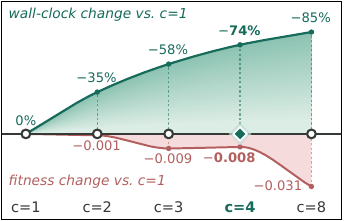}
  \vspace{-1.6em}
  \captionof{figure}{Concurrency trade-off on Signal Processing.}
  \label{fig:concurrency-tradeoff}
  \vspace{-1.5\baselineskip}
\end{wrapfigure}

Our main experiments use $c=4$ concurrent task-executor slots
(\S\ref{sec:exp-setup}). Here we examine how $c$ affects wall-clock time and
final fitness. Parallel execution reduces wall-clock time, but tasks bind
their contexts before sibling tasks finish
(\S\ref{sec:open-proposal}), so recent outcomes remain unavailable as
$c$ grows.

To characterize this trade-off, we vary $c\in\{1,2,3,4,8\}$ on Signal
Processing. Each setting is run three times, and
Figure~\ref{fig:concurrency-tradeoff} reports the mean wall-clock and fitness
changes from $c=1$. Wall-clock savings show diminishing returns:
$c=4$ already achieves a 74\% reduction, while doubling to $c=8$ adds only
11 percentage points. Fitness loss remains below 0.009 through $c=4$ but
rises to 0.031 at $c=8$. Together, these results place the knee at $c=4$.
Appendix~\ref{app:concurrency-staleness} models the evidence lag behind this loss.

%% file: sections/7_conclusion.tex
\section{Conclusion}
\label{sec:conclusion}

We introduced EvoTrias, a separation-of-powers architecture for LLM-driven
program evolution. It answers autonomy allocation by giving govern, propose,
and judge one owner each. Typed
contracts connect the three planes, and a runtime hook layer reinforces the
split. Across the Math, ADRS, and Frontier-CS benchmarks, EvoTrias sets a
new state of the art. It also extends program evolution to a 680K-line
industrial repository, cuts token cost by 64\% against four-agent CORAL,
and keeps the evaluator private so that benchmark gains carry over to unseen
instances.
The same decision boundaries support deployment as a coding-agent plugin,
equipping existing hosts with multi-round optimization under dedicated
governance and evaluation.

%% file: sections/9_statements.tex
\par\bigskip

\section*{AI use statement}

We used AI tools in limited, supporting roles for writing assistance
(including grammar correction and language polishing), retrieval and
discovery of related work, and implementation of parts of the code. For these
uses, the authors retained full responsibility for the research decisions and
carefully reviewed and verified all resulting text, code, analyses, and
results. The coding agents in EvoTrias and in every baseline use LLMs as
their behavioral backbone as part of the experimental systems studied here.
The authors take responsibility for the final content and conclusions.

\section*{Ethics statement}

This work does not involve human subjects or personal data. All
experiments run on public benchmarks (Math, ADRS, and Frontier-CS) and on
the open-source OR-Tools repository, each used under its license, and every
LLM backbone is accessed under its provider's terms of use for research.

%% file: sections/8_appendix.tex
\clearpage
\phantomsection
\section*{Appendix Contents}
\label{app:contents}

\begingroup
\normalsize
\hypersetup{hidelinks}
\setcounter{tocdepth}{1}
\makeatletter
\newcommand{\appcontentssection}[2]{%
  \vspace{0.2em}%
  \@dottedtocline{0}{0em}{2.5em}%
    {\hyperref[#1]{\textbf{\ref*{#1}\enspace #2}}}%
    {\hyperref[#1]{\textbf{\pageref*{#1}}}}}
\newcommand{\appcontentssubsection}[2]{%
  \@dottedtocline{1}{1.5em}{3.2em}%
    {\hyperref[#1]{\ref*{#1}\enspace #2}}%
    {\hyperref[#1]{\pageref*{#1}}}}

\appcontentssection{app:method-details}{Method Details}
\appcontentssubsection{app:host-integration}{Integration with Coding-Agent Hosts}
\appcontentssubsection{app:routing}{Design Details of Structured Governance}
\appcontentssubsection{app:init-directions}{Initial Directions}
\appcontentssubsection{app:notes}{Notes}
\appcontentssubsection{app:guidance}{Meta-guidance}
\appcontentssubsection{app:concurrency-staleness}{Context Staleness under Concurrency}
\appcontentssubsection{app:hooks}{Runtime Hooks}
\appcontentssubsection{app:judge-baselines}{Judge in Existing Systems}
\appcontentssubsection{app:discussion}{Discussion: What the Boundaries Cost and Buy}
\appcontentssubsection{app:limitations}{Limitations}

\appcontentssection{app:experiment-details}{Experimental Setup Details}
\appcontentssubsection{app:eval-protocol}{Evaluation Protocol}
\appcontentssubsection{app:coral-budget}{Budget Alignment Across Baselines}
\appcontentssubsection{app:hyperparameters}{Hyperparameter Configuration}
\appcontentssubsection{app:baseline-configs}{Baseline Configurations}

\appcontentssection{app:additional-results}{Additional Results and Analysis}
\appcontentssubsection{app:mean-best-results}{Math and ADRS: Statistical Results}
\appcontentssubsection{app:backbone-gap}{Where Backbone Strength Shows Up}
\appcontentssubsection{app:frontier-analysis}{Frontier-CS: Statistical Results}
\appcontentssubsection{app:iteration-trajectories}{Best-so-Far Optimization Trajectories}
\appcontentssubsection{app:agentic-baselines}{Controller-Centric Baselines with an Agentic Proposer}
\appcontentssubsection{app:channel-prevalence}{Cross-Run Prevalence of the Two Experience Channels}
\appcontentssubsection{app:note-content-mechanism}{Mechanistic Content of Evolution Notes}
\appcontentssubsection{app:ortools}{OR-Tools Case: Experimental Settings}
\appcontentssubsection{app:lineage-case-studies}{Evolutionary Lineage Case Studies}
\appcontentssubsection{app:evaluator-exposure}{Evaluator Exposure: Protocol, Held-out Suites, and Code Cases}
\appcontentssubsection{app:ownership-ablation}{Ownership Ablation: Protocol and Full Results}

\appcontentssection{app:prompts}{Prompts}
\appcontentssubsection{app:prompt-orchestrator}{Orchestrator}
\appcontentssubsection{app:prompt-init}{Initial Directions}
\appcontentssubsection{app:prompt-executor}{Task Executor}
\appcontentssubsection{app:prompt-note}{Notes}
\appcontentssubsection{app:prompt-meta}{Meta-guidance}
\makeatother
\endgroup

\clearpage
\let\section\sectionwithoutfloatbarrier
\raggedbottom
\input{sections/appendix/a_method_details}
\input{sections/appendix/b_experimental_setup}
\input{sections/appendix/c_additional_results}
\input{sections/appendix/d_prompts}

%% file: sections/appendix/a_method_details.tex
\appendix


\section{Method Details}
\label{app:method-details}

\subsection{Integration with Coding-Agent Hosts}
\label{app:host-integration}

EvoTrias's allocation of decision rights defines how an existing coding agent
can participate in evolutionary search. Governance owns cross-round
selection, experience admission, and budget accounting; Judgment owns
official evaluation. Proposal consumes the assigned context and returns
artifacts through the contracts of Equation~\ref{eq:plane-contract}.
Separating these responsibilities allows the implementation of Proposal to
change across hosts while retaining the governance rules and evaluation
semantics. EvoTrias realizes this separation as a plugin that adds an
evolution workflow to the host's native execution capabilities; every EvoTrias
result in this paper was produced through this plugin, on Claude Code or OpenCode.

\paragraph{Workflow in the host session.}
For a prepared task and evaluator, a skill activates the workflow in the
user's current coding-agent session. That session becomes the orchestrator:
it initializes the run, follows the scheduling decisions returned over MCP,
and dispatches native subagents for construction and requested experience
generation. Executors retrieve their assigned context from Governance and
submit artifacts to Judgment; settled outcomes inform subsequent tasks.
The plugin supplies the workflow skills, agent roles, MCP configuration, and
runtime hooks. The host supplies the conversational entry point, tools, and
native subagent execution; dedicated components maintain search state and
official evaluation. Appendix~\ref{app:prompt-orchestrator} gives the
orchestrator's bounded dispatch protocol.

\paragraph{What a host must supply.}
A compatible host must support the context, artifact, and outcome contracts
of Equation~\ref{eq:plane-contract}. At runtime it must provide native task
dispatch, stable run and task identities, and lifecycle events that can
maintain the shared ledger. Its integration must also enforce the
task-access and evaluator-access boundaries of \S\ref{sec:open-proposal}.
A thin adapter translates the host's native tool calls, identifiers, and
events into these shared interfaces: the four event roles of
Table~\ref{tab:hook-events} for scheduling, the pre-tool event for
permission interception, and the host's own role and MCP configuration.
Appendix~\ref{app:host-portability} shows this mapping for Claude Code and
OpenCode. Under these requirements, host replacement changes the
Proposal implementation while selection, Notes admission, budget accounting,
and evaluation retain their defined semantics.

\paragraph{Two paths into a task.}
Project knowledge and search evidence reach a task by different paths.
Project files, build instructions, and skills enter through the public task
contract and the task's access rules, so knowledge the project already holds
is available to construction. Search evidence enters only through
Governance, which decides which parent, inspirations, Notes, and guidance
each task receives. Subagents run in separate conversation contexts, so this
second path is the only route by which one attempt's experience reaches
another. Artifacts leave a task the way project changes normally do: in the
OR-Tools case (Appendix~\ref{app:ortools}), construction exports a Git patch
against the frozen tree, and the project's review workflow applies it
directly.

\begin{figure}[htbp]
  \centering
  \includegraphics[width=\linewidth]{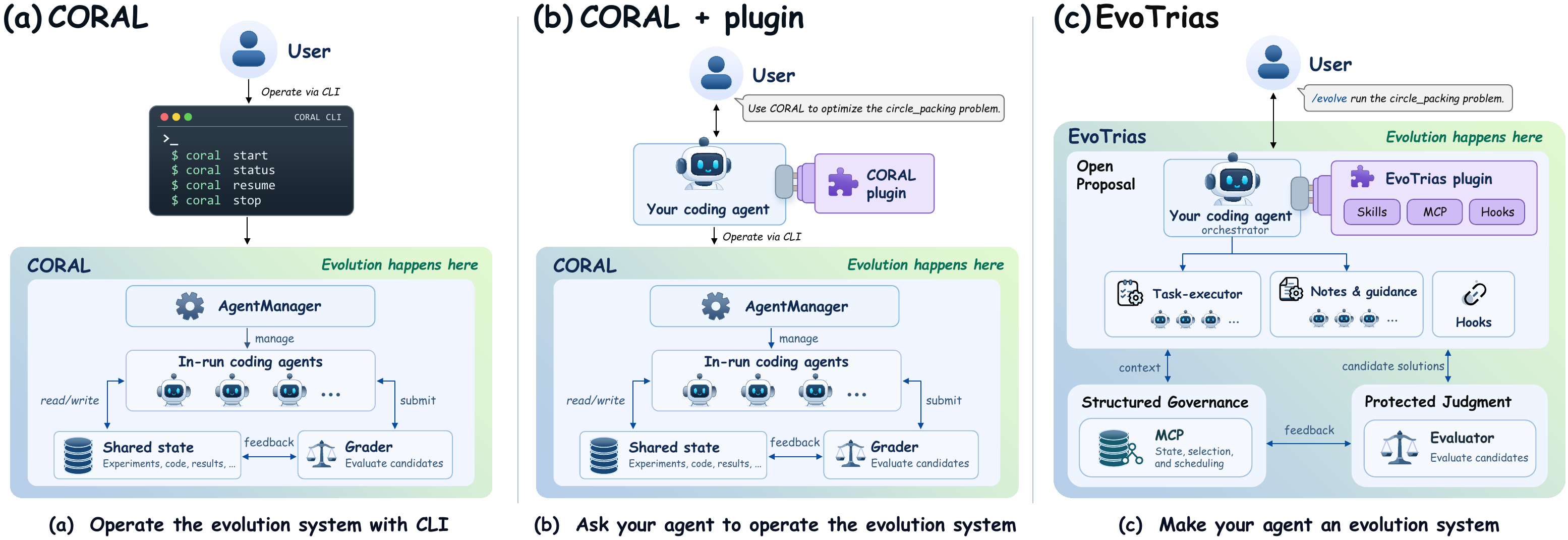}
  \caption{From operating an evolution system to becoming one. (a)~The user
  operates CORAL through its CLI. (b)~CORAL's plugin lets the user's coding
  agent issue the same operations; the search still runs in a separately
  managed runtime. (c)~EvoTrias's plugin makes the user's coding agent the
  evolution system: the existing session orchestrates native subagents, while
  dedicated components govern the search and judge its artifacts. All three
  assume a prepared task and evaluator.}
  \label{fig:plugin-integration}
\end{figure}

\paragraph{Operating an evolution system versus becoming one.}
Figure~\ref{fig:plugin-integration} places the two systems side by side.
CORAL~\citep{qu2026coral} deploys as a standalone runtime: a Python
AgentManager launches and supervises the coding agents that carry out the
search, and the user operates that runtime through a CLI that launches,
inspects, resumes, and stops runs, with a dashboard for observation.
EvoTrias launches no agents of its own: it deploys as a plugin, and the
search runs inside the host coding agent. CORAL also ships a plugin, but as
a supplementary entry point: it lets the user's coding agent issue the same
CLI operations, so the user's agent operates the evolution system from
outside it.\footnote{CORAL public revision \texttt{0123dfb} (2026-09-06):
the operator skill under \texttt{plugin/skills/}, the CLI launch in
\texttt{coral/cli/start.py}, and the AgentManager in
\texttt{coral/agent/manager.py}.}
EvoTrias's plugin makes the user's coding agent the evolution system: the
session the user already works in orchestrates the search, its native
subagents construct candidates, and the plugin adds the governance and
judgment components those subagents talk to. The same session that develops
a project can therefore evolve it, with cross-round decisions and official
evaluation held outside the agent's discretion.
\FloatBarrier

\subsection{Design Details of Structured Governance}
\label{app:routing}

Appendices~\ref{app:routing} to~\ref{app:guidance} complete the design of
\S\ref{sec:structured-governance}. This section covers the parts owned by
Governance alone: the descendant potential and the full selection rules
behind $(\pi^L,\pi^P,\pi^I)$. The initial directions and the two
experience channels, whose content is produced on the Open Proposal side,
follow in Appendices~\ref{app:init-directions}, \ref{app:notes},
and~\ref{app:guidance}. Default values for all constants appear in
Table~\ref{tab:hyperparams}.

\subsubsection{Descendant Potential}
\label{app:potential}

For a parent or lineage $u$ whose settled descendants number $n_u$, with
$s_u$ of them improving on their parent,
\begin{equation}
\label{eq:potential}
g_u=\frac{\alpha+s_u}{\alpha+\beta+n_u}-\frac{\alpha}{\alpha+\beta}
\end{equation}
shrinks this improvement rate toward a neutral prior: credit accrues only
as improving descendants accumulate, and $g_u=0$ without descendant
evidence. The same form applies at both granularities. Parent-level
$g_i$ enters the parent score of Equation~\ref{eq:parent} with weight
$\lambda_P$; lineage-level $g_\ell$ enters the routing score below with
weight $\lambda_L$. Both weights are fixed and small, so potential breaks
ties among comparable candidates without overriding verified quality.

\subsubsection{Selection Rules}
\label{app:selection}

\paragraph{Annealing.}
Every scheduled constant follows one exponential form. For endpoints
$(x_0,x_1)$ and budget progress $\beta_t\in[0,1]$,
\begin{equation}
\label{eq:anneal}
x_t=x_0\,(x_1/x_0)^{\beta_t}.
\end{equation}
The routing temperature $\tau_t$, the exploration coefficient
$c_{\mathrm{local},t}$, and the inspiration temperature $T_t$ all anneal
this way, which moves the search from broad to deep as the budget
depletes.

\paragraph{Lineage routing.}
For lineage $\ell$, $q_\ell$ is the min-max normalized best verified
fitness and $a_\ell$ its share of the successful population:
\begin{equation}
\label{eq:lineage-routing}
\begin{aligned}
r_\ell
&=q_\ell+\lambda_Lg_\ell,
&d_\ell
&=\frac{\exp(r_\ell/\tau_t)}{\sum_k\exp(r_k/\tau_t)},\\
D_\ell
&=\frac{d_\ell}{a_\ell+1/N},
&\ell_t^\star
&=\argmax_\ell D_\ell(1+\eta r_\ell),\\
\pi_t^L(\ell)
&=p_{\mathrm{exploit}}\mathbf 1[\ell=\ell_t^\star]
 +(1-p_{\mathrm{exploit}})\frac{D_\ell}{\sum_kD_k}.
\end{aligned}
\end{equation}
Quality and potential set the deserved share $d_\ell$; the deficit $D_\ell$
compares it with the realized share $a_\ell$, smoothed by $1/N$ for a
population of $N$ candidates, favoring strong but
under-covered lineages. With probability $p_{\mathrm{exploit}}$ the
routing takes the deficit leader with a quality bonus of strength $\eta$;
the remaining mass is deficit-weighted roulette.

\paragraph{Parent rule.}
In Equation~\ref{eq:parent}, quality and exploration take
\begin{equation}
\label{eq:parent-internals}
q_i=\frac{N-\operatorname{rank}(i)}{\max(N-1,1)},\qquad
U_{i,t}=c_{\mathrm{local},t}
  \sqrt{\frac{\log(S_{\ell_t}+1)}{e_i^{\mathrm{eff}}+1}},
\end{equation}
where $N$ is the population size, $\operatorname{rank}(i)$ orders all
candidates by fitness, and $S_{\ell_t}$ counts parent selections from the
lineage. The effective exposure count applies an elite shield,
\begin{equation}
\label{eq:elite-shield}
e_i^{\mathrm{eff}}=
\begin{cases}
(1-s)\,e_i,&\operatorname{rank}(i)\le\lceil\kappa N\rceil,\\
e_i,&\text{otherwise},
\end{cases}
\end{equation}
where $e_i$ is the raw selection count, $\kappa$ the elite rank fraction,
and $s$ the shield strength. Repeated selection therefore suppresses
ordinary candidates faster than top-ranked ones, which keeps the best
candidates reachable late in the run. Each selection increments $e_i$
immediately, so concurrent tasks see the updated count.

\paragraph{Inspiration rule.}
In Equation~\ref{eq:inspiration}, local sampling is Boltzmann over
min-max normalized fitness within the parent lineage,
$\pi_{\mathrm{local}}(i\mid\ell_t,T_t)\propto\exp(\widetilde f_i/T_t)$.
The cross-lineage pool collects the top $n_e$ records of every other
lineage, and the donor distribution $\rho_t$ weights these pools. Drawing
uniformly over the pooled records is the baseline while the population stays
balanced: it gives every lineage donor mass in proportion to the elites it
can actually supply, which is equal once every lineage holds $n_e$ of them.
Let $H_t$ be the normalized Shannon
entropy of the lineage shares. As it falls, the donor mass tilts toward
quality,
\begin{equation}
\label{eq:donor}
\rho_t(k\mid E_t)\propto\min(|k|,n_e)\,\bigl(1+\gamma_t\,\widetilde q_k\bigr),
\qquad
\gamma_t=\gamma_{\max}\operatorname{clip}\!
  \Bigl(\frac{H^\star-H_t}{H^\star-H_{\min}},0,1\Bigr),
\end{equation}
where $\widetilde q_k$ is the min-max normalized best fitness of lineage
$k$ and $|k|$ is the number of records lineage $k$ holds. The ramp spans $[H_{\min},H^\star]$ rather than $[0,H^\star]$ because
normalized entropy is insensitive to mild skew: four lineages holding
$(20,5,5,5)$ records sit at $H_t=0.83$, and $(14,6,6,6)$ at $0.94$. The
tilt is bounded by construction: the quality factor never exceeds
$1+\gamma_{\max}$, so quality can move a lineage's donor mass by at most that
factor, and at $H_t\ge H^\star$ the weights fall back to the pool-size
baseline. The remaining variation is the pool-size factor
$\min(|k|,n_e)$, which is bounded by $n_e$ and equal across lineages once
each holds $n_e$ records. The migration rate stays
fixed, so the volume of cross-lineage material is unchanged and only its
composition moves. Each inspiration entry draws independently with
migration probability $\mu$, skips records already selected, and falls back
to the whole population when the local pool is exhausted. The prompt
presents the parent first, then local inspirations, then migrated ones,
each group sorted by fitness.

\paragraph{Boundary cases.}
The rules above run on the population of settled candidates that the
evaluator marked valid with a numeric fitness; invalid attempts never
enter $N$, the rank, or any lineage share (Appendix~\ref{app:ledger}).
$\operatorname{rank}(i)$ is global over this population and strict: two
candidates with equal fitness are ordered by settlement time, the earlier
one ranking higher, so $q_i$ never assigns the same quality to two
candidates. Every min-max normalization guards a zero range. When all
lineages share the same best fitness, $q_\ell$ and $\widetilde q_k$ are set
to zero, so routing falls back to potential and deficit alone and the donor
weights fall back to the pool-size baseline; when all candidates in a
lineage share one fitness, $\pi_{\mathrm{local}}$ is uniform. A lineage
without a settled success has $a_\ell=0$ and so the largest deficit for its
deserved share; this is the intended behavior, since such a lineage has
received the least coverage. With a single lineage, routing is skipped and
the parent rule runs directly; before any lineage exists, the parent is the
global best.

\subsection{Initial Directions}
\label{app:init-directions}

Root establishment opens every run. The orchestrator dispatches a
dedicated init-explorer agent that reads the public contract and the
initial reference program, writes a short structural analysis of the
problem (its type, the available feedback, and the decision axes where
solvers can diverge), and generates the $n_{\mathrm{dir}}$ directions
$\mathcal D_0=\{d_i\}_{i=1}^{n_{\mathrm{dir}}}$, one per seeded root. Each
$d_i$ is a single line of 8--25 words naming a concrete approach. A pairwise
cross-check then replaces any two directions that would converge on the
same core strategy, so the roots start from distinct approach families
instead of variations of the reference program. Each $d_i$ binds only its
root task, and the descent subtrees of the roots become the lineages
of \S\ref{sec:structured-governance}. The generation prompt appears in
Appendix~\ref{app:prompt-init}.

Figure~\ref{fig:init-directions} shows the four directions generated for
the MinMax-Dist-3 run of Figure~\ref{fig:meta-guidance-case}. The four
lineages settled 7, 24, 8, and 7 candidates. The polyhedron-construction
root grew into the dominant lineage: it carried the surrogate optimum
that round-1 guidance later diagnosed, and all three guided new bests
descend from it. The other three lineages topped out between 0.9948 and
0.9975.

\begin{center}
\begin{minipage}{\textwidth}
\begin{guidancecard}{Initial directions}{MinMax-Dist-3 run of
Figure~\ref{fig:meta-guidance-case}; one line per seeded root, quoted
verbatim}
\textbf{Root 1 - Spherical Tammes code.} "Spherical Tammes code: place
all 14 points on the unit sphere using the known optimal 14-point
spherical configuration to fix max distance at diameter 2."
\textbf{Fate:} 7 settled descendants; lineage best 0.9975.

\textbf{Root 2 - Polyhedron vertex construction.} "Catalan/Archimedean
polyhedron vertex construction: use vertices of a 14-vertex solid like
the rhombic dodecahedron or triakis octahedron as a closed-form symmetric
arrangement." \textbf{Fate:} 24 settled descendants; carries the
canonical best throughout and all three guided new bests of
Figure~\ref{fig:meta-guidance-case}; lineage best 1.00002.

\textbf{Root 3 - Hexagonal antiprism plus poles.} "Hexagonal antiprism
plus two poles: arrange 12 vertices on two staggered hexagonal rings plus
north and south poles along the symmetry axis." \textbf{Fate:} 8 settled
descendants; lineage best 0.99627.

\textbf{Root 4 - Numerical optimization.} "Numerical optimization
approach: use scipy.optimize with multiple deterministic seeds to
maximize the minimum pairwise distance, then normalize max distance to
1." \textbf{Fate:} 7 settled descendants; lineage best 0.99482.
\end{guidancecard}
\captionsetup{hypcap=false}
\captionof{figure}{The four initial directions of the MinMax-Dist-3 run,
quoted from the stored direction files, with the settled size and best
fitness of the lineage each root seeded.}
\label{fig:init-directions}
\end{minipage}
\end{center}

\subsection{Notes}
\label{app:notes}

\paragraph{Triggers.}
After each settled attempt with fitness $f$ and parent fitness $f_p$, the
relative delta is $\delta=(f-f_p)/\max(|f_p|,\varepsilon)$, with
$\varepsilon$ a small numerical floor. A note
request fires in three cases: $\delta$ clears the positive threshold; the
magnitude of a negative $\delta$ clears the negative threshold; or the
attempt history shows a repair (failure followed by success within the
task) or a terminal failure. Each threshold is the $q$-quantile of the
past positive or negative delta magnitudes, with $q$ annealed from $0.80$
to $0.60$ over iterations 11--50 and the result clamped to the
$[0.02,0.05]$ band; an empty history takes the band ceiling. Early noise
therefore cannot flood the pool, and late thresholds stay meaningful. A positive request must also
set a new best or satisfy $f/b^{-}\ge r_t$, where $b^{-}$ is the best
settled fitness before the attempt and $r_t$ anneals from $0.80$ to $1.00$
over iterations 11--50. Negative requests use only the corresponding
delta threshold. No requests fire during the first ten iterations.
Near-duplicate positive requests at the same fitness level are skipped,
and every request carries a deterministic id, so retries commit at most
once.

\paragraph{Loading.}
Note loading runs after the parent and inspirations are fixed. Together they
form the task's \emph{anchor set}, and a stored note is \emph{anchor-matched}
when its own provenance, the task it was distilled from together with that
task's parent and inspirations, intersects this set. The most recent error
notes load first, up to ten; they load by recency and are never anchored.
Positive and negative notes then fill a target of five: anchor-matched notes
take priority, and a seeded random draw fills the remainder, which keeps
repeated loads stable for the same context. The share of loaded positive and
negative notes that are anchor-matched is the \emph{anchor hit rate}, which
separates retrieval tied to the consuming context from a draw over the whole
pool; Appendix~\ref{app:channel-prevalence} measures it across the audited
runs. The analysis prompt sent with each note request appears in
Appendix~\ref{app:prompt-note}.

\begin{figure}[t]
  \centering
  \includegraphics[width=0.97\textwidth]{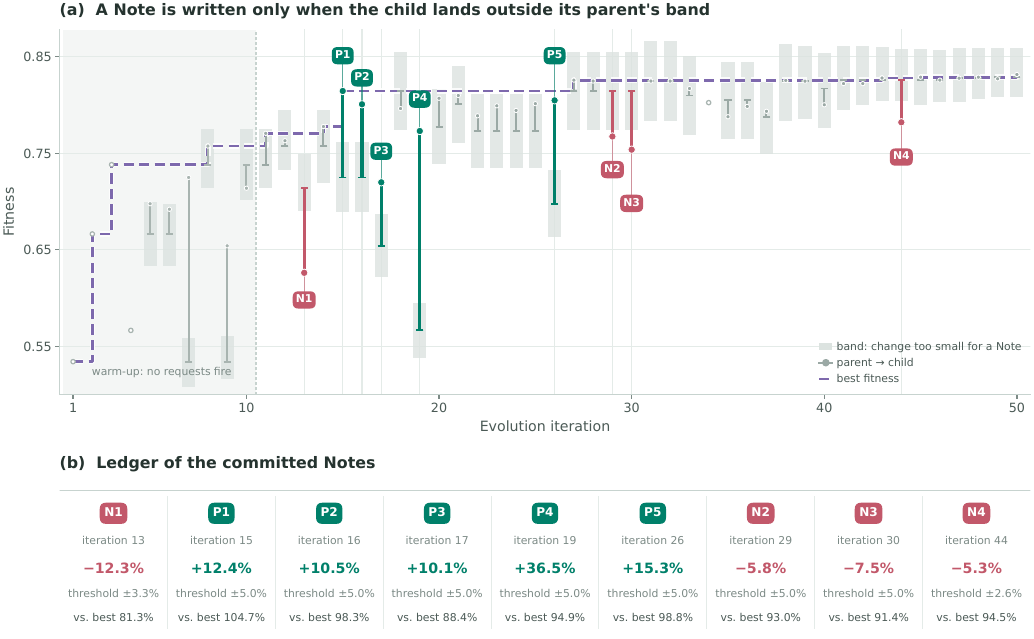}
  \caption{Note generation in a 50-iteration Signal Processing run
  with Opus-5. \textbf{(a)} Each settled attempt is a segment from its
  parent's fitness to its own; the grey band is the range around the parent
  within which the change is too small to trigger a request. Segments that
  leave the band are colored by the pool they fed. The dashed line is the
  best fitness settled so far, and the shaded opening interval is the
  ten-iteration warm-up. \textbf{(b)} The nine positive and negative Notes
  committed in this run, with the
  iteration, the child--parent delta, the threshold in force, and the ratio
  to the best fitness settled before the attempt; the run's single error Note
  is omitted, as in panel (a). P1--P5 and N1--N4 match
  Figures~\ref{fig:positive-note-cases} and~\ref{fig:negative-note-cases}.}
  \label{fig:note-trigger-replay}
\end{figure}

Figure~\ref{fig:note-trigger-replay} replays these rules on one 50-iteration
Signal Processing run with Opus-5. Both delta thresholds are relative to the
parent, so they can be drawn as a band
$[\,p\,(1-\tau^{-}_t),\; p\,(1+\tau^{+}_t)\,]$ around each parent's fitness
$p$. This reduces the magnitude test to one visual comparison: a request
fires exactly when the child lands outside that band. The band is asymmetric
because the positive and negative thresholds adapt independently, and the
first four attempts have no parent, so they carry no band. The panel omits
the gate; the ledger instead lists $f/b^{-}$, which for the five positive
Notes can be read against the $r_t$ schedule above. Governance committed
five positive and four negative Notes; its failed-to-success event produced a
separate error Note and is not counted in the two plotted pools.
\FloatBarrier

\enlargethispage{3\baselineskip}
\input{sections/generated/positive_notes_case}
\vspace{8pt}
{\captionsetup{hypcap=false}
\captionof{figure}{The five positive Notes committed during the Signal
Processing run in Figure~\ref{fig:note-trigger-replay}; each card
lightly condenses the stored fields, and P1--P5 match the trigger labels in
the replay.}
\label{fig:positive-note-cases}}

\input{sections/generated/negative_notes_case}
\vspace{8pt}
{\captionsetup{hypcap=false}
\captionof{figure}{The four negative Notes committed during the Signal
Processing run in Figure~\ref{fig:note-trigger-replay}. N1--N4 match the
trigger labels in the replay.}
\label{fig:negative-note-cases}}
\medskip

Across the five positive Notes, the recurring pattern is to preserve a
reasonably wide smoother and control reversals through a separate
shape-domain stage, while selecting bandwidth and pruning strength with an
accuracy-aware surrogate. The four negative Notes mark the complementary
failure modes: permissive knot refits, changepoint branches admitted without
reversal checks, and replacement criteria that reward smoothness while losing
correlation or noise reduction. These observations are specific to the
recorded comparisons. Several Notes explicitly state that the winning branch
or individual change was not isolated, so they support conditional guidance
rather than a general claim about an estimator family.

\subsection{Meta-guidance}
\label{app:guidance}

\paragraph{Stall detection.}
Let $b_1,\dots,b_n$ be the running best fitness after each success. With
window $w$, the windowed improvement is
\begin{equation}
\label{eq:plateau}
I_n=\frac{b_n-b_{n-w}}{\max(|b_{n-w}|,\varepsilon)},
\end{equation}
and a window stalls when $I_n\le 0$. The detector requires at least
$\max(w,5)$ successes before judging. A guidance request fires when
$P_{\mathrm{stall}}$ consecutive windows stall and the newest outcome is not
a new best; a new best resets the count.

\paragraph{Request and commit.}
At most one request is pending at a time. The request packages the
canonical best with an artifact excerpt, stall diagnostics and budget
progress, per-lineage summaries, a summary of admitted notes, and the
full history of earlier guidance with their recorded outcomes. The
proposal-side meta actor returns a small set of semantically distinct
directions. Each direction is a typed record: an approach label with a
one-line idea, a description that argues from the lineage record, a
step-level implementation plan, execution cautions, a novelty claim
against the tried history, and an avoid-repeating list that names the
tactics the lineages have already spent. Governance validates this
schema and commits the directions. The request payload and the
generation instructions appear in Appendix~\ref{app:prompt-meta}.

\paragraph{Active use.}
Committed directions activate one at a time. When a task is assigned the
active direction, its parent anchors to the canonical best at assignment
time, and the direction's use count increments, up to two uses. After
each guided task settles, the fitness change against the anchor is
recorded on that direction. Exhausted directions retire into the history
that the next request will carry, so the meta actor sees which directions
helped and which failed before proposing new ones.

\begin{figure}[t]
  \centering
  \includegraphics[width=0.97\textwidth]{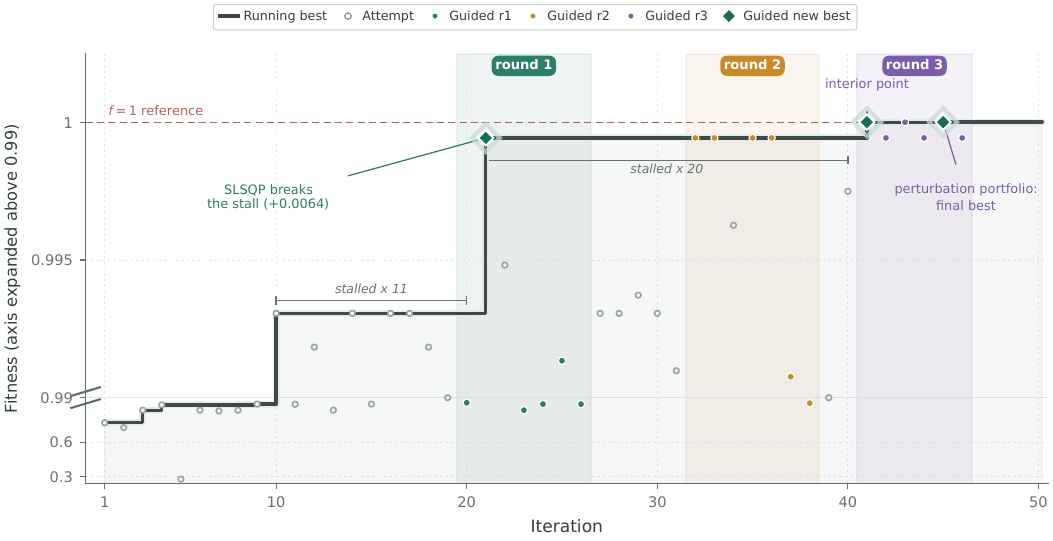}
  \caption{Meta-guidance lifecycle on a 50-iteration
  MinMax-Dist-3 run. The y-axis expands fitness above 0.99; the double
  slash marks the compressed range. Shaded bands cover the iterations
  bound to each guidance round, colored dots mark guided outcomes,
  diamonds mark guided outcomes that set a new best, and brackets measure
  the two stalls. Figure~\ref{fig:meta-guidance-rounds} lists the
  committed directions.}
  \label{fig:meta-guidance-case}
\end{figure}

\paragraph{A lifecycle case.}
Figure~\ref{fig:meta-guidance-case} replays this lifecycle on one
50-iteration MinMax-Dist-3 run, which asks for 14 points that maximize
the squared ratio of minimal to maximal pairwise distance. The fitness
$f$ in the figure is the evaluator's normalized score, so $f=1$ matches
the AlphaEvolve reference construction~\citep{novikov2025alphaevolve};
on the raw scale of Table~\ref{tab:combined-mean-results} this run's
final best $f=1.00002$ is $0.240051$, which rounds to the $0.24005$ that
table averages over three runs. Two stalls trigger three guidance
requests and nine committed directions
(Figure~\ref{fig:meta-guidance-rounds}).

The three rounds do different work. The first request diagnoses the
stall from the lineage record: eleven outcomes repeat one smooth
surrogate whose annealed gradients share a single basin, so all three
directions replace the machinery instead of tuning it. D1's hard
constrained program lands the new best, and its two uses also show the
protocol's tolerance, one failed implementation and one success. The
second round asks whether that optimum can be polished further and
spreads its directions over three method families. Two of them return
the anchor value to the last digit and the third lands below it; two
independent families arriving at the same point is consistent with an
exhausted basin. The third round answers that record. D7's closed form
returns the identical value a third time, while D8 keeps the program and
changes the solver family and D9 keeps the solution and perturbs it.
Both escapes cross the reference level, and D9 sets the final best.

The case also shows the lifecycle's bookkeeping at work. Failure history
is visible in content: the round-3 avoid-repeating lists name round-2
tactics, and D9 cites D6's seed portfolio verbatim. Anchoring is visible
in structure: D1's artifact is the anchor parent of rounds 2 and 3,
later directions load it by task id in their plans, and each
retirement records its effect against that anchor. The contrast is
visible in the unguided lane: between the rounds, unguided descendants
keep sampling the surrogate family and settle between 0.9894 and 0.9975,
all short of the anchor, and all three new bests after the ramp-up come
from guided tasks.

\begin{guidancecard}{Round 1}{requested after 11 stalled outcomes;
canonical best 0.9931 when the round opened}
\textbf{D1 - Constrained dual program.} The request names the
bottleneck: every lineage refines one smooth surrogate whose annealed
gradients share a single basin at 0.9931. This direction drops the
surrogate and attacks the hard program itself, maximizing $t$ subject to
$d_{ij}^2\ge t$ and $d_{ij}^2\le 1$ over all pairs with an active-set
SQP solver. \textbf{Plan:} encode the 43-variable program with explicit
pairwise constraints, run SLSQP from 8-12 normalized random starts,
keep the best feasible point, then polish on the raw ratio.
\textbf{Caution:} verify feasibility after SLSQP and re-normalize, since
the solver can stop at a slightly infeasible point. \textbf{Outcome:}
0.9452 and 0.99944; new best ($+0.0064$), and its artifact anchors both
later rounds.

\textbf{D2 - Exact algebraic construction.} Bypass numerical search:
build the 14-point configuration from coordinates reported for the
14-point Tammes problem, constructed algebraically so the ratio equals
the claimed optimum. \textbf{Plan:} hard-code the exact coordinates with
cited sources, normalize, and clean up floating-point error with one
surrogate polish. \textbf{Caution:} the request forbids fabricated
coordinates and demands a numerical fallback if the exact optimum is
uncertain. \textbf{Outcome:} 0.8803 and 0.9331; retired.

\textbf{D3 - Spherical-code portfolio.} Recast the task as a spherical
code: place 14 points on the unit sphere for maximal angular separation,
seeding from published code tables that the request judges more diverse
than the five ring-based seeds the lineages reuse. \textbf{Plan:}
hard-code three to four candidate codes, run each through the existing
polish pipeline, keep the best. \textbf{Caution:} skip seeds that are
rotations of the rhombic dodecahedron, which would collapse into the
known basin. \textbf{Outcome:} 0.9913 and 0.9332; retired.
\end{guidancecard}

\begin{guidancecard}{Round 2}{requested after 20 stalled outcomes;
canonical best 0.99944 when the round opened}
\textbf{D4 - Hybrid global-local polish.} Feed the round-1 SLSQP
artifact into the smooth-surrogate pipeline of the earlier lineages, on
the reasoning that SLSQP fixes the feasibility-boundary structure while
the surrogate refines interior spacing. \textbf{Plan:} load the anchored
best by task id, run the surrogate refine, a Nelder-Mead polish, and
two basin hops; return the best of the three stages. \textbf{Caution:} if
the polish regresses, return the anchor unchanged. \textbf{Outcome:} both
uses return exactly 0.99944.

\textbf{D5 - CMA-ES black box.} Treat the 42-dimensional configuration
as a black box and run derivative-free CMA-ES seeded at the anchored
best, since a population method can cross basins that gradient steps
cannot. \textbf{Plan:} 2{,}000 evaluations at $\sigma_0=0.05$, an optional
restart with doubled population, then a Nelder-Mead polish.
\textbf{Caution:} guard degenerate configurations and fix the random seed
for reproducibility. \textbf{Outcome:} both uses return exactly 0.99944.

\textbf{D6 - Structured multi-start.} Expand SLSQP from 11 seeds to
more than 40, drawn from named symmetry families (hexagonal antiprism
with poles, heptagonal bipyramid, snub cube, icosahedron with poles, and
others), so each start explores a different KKT neighborhood.
\textbf{Plan:} normalize every seed, run SLSQP per seed, keep the best
feasible result, polish. \textbf{Caution:} drop seeds that are rotations
of one another. \textbf{Outcome:} 0.9908 and 0.9412; retired.
\end{guidancecard}

\begin{guidancecard}{Round 3}{requested with the round-2 record in the
failure history; canonical best 0.99944 when the round opened}
\textbf{D7 - Dual-anchor ring construction.} Solve a closed-form
construction with two antipodal poles and two staggered rings of six,
where equal near-neighbor distances reduce the whole configuration to a
single co-latitude angle found by one-dimensional root finding.
\textbf{Caution:} if the construction lands below the anchor's raw ratio,
fall back to the anchor. \textbf{Outcome:} both uses return exactly
0.99944, rediscovering the same optimum from a third method family.

\textbf{D8 - Interior-point solver.} Keep the dual program of D1 and
swap the active-set solver for a primal-dual interior-point method
(trust-constr), which reaches KKT points through a barrier path that
active sets can miss. \textbf{Plan:} reuse the parent's analytic
constraint Jacobians, run from the 11 seeds plus the anchor, keep the
best feasible point, polish. \textbf{Caution:} bound iterations, and
return the anchor if the method fails to improve. \textbf{Outcome:}
1.000013 in both uses; crosses $f=1$.

\textbf{D9 - Perturbation portfolio.} Re-run SLSQP from twelve
structured perturbations of the anchored best: six small Gaussian kicks,
three rigid rotations against axis-alignment bias, and three point-pair
swaps against permutation symmetry. Its avoid-repeating list cites D6's
seed portfolio by name. \textbf{Caution:} every perturbation starts from
the anchor itself, never from another perturbation, to avoid drift.
\textbf{Outcome:} 1.0000159, the final best, and 0.99944.
\end{guidancecard}
\vspace{10pt}
{\captionsetup{hypcap=false}
\captionof{figure}{The nine directions committed during the
MinMax-Dist-3 run of Figure~\ref{fig:meta-guidance-case}, condensed from
the stored direction records and grouped by round. D1--D9 follow commit
order; plans and cautions are excerpts of the stored fields, and every
outcome is the official fitness of a guided task.}
\label{fig:meta-guidance-rounds}}

\FloatBarrier
\subsection{Context Staleness under Concurrency}
\label{app:concurrency-staleness}

This appendix models how concurrency delays the settled evidence
available when a task binds its context
(\S\ref{sec:concurrency-analysis}).

\paragraph{Setup.}
Index tasks by the logical instant $t=1,\dots,T$ at which each binds its
context (\S\ref{sec:decision-planes}), with a fixed task budget $T$. Under
concurrency $c$, up to $c$ tasks are in flight simultaneously. Assume equal
task durations and FIFO settlement. After the initial queue-filling phase,
$c-1$ previously bound tasks remain in flight at each new binding. With
$y_j$ denoting the settled record of task $j$, task $t$ therefore observes
\begin{equation}
\label{eq:stale-history}
\mathcal H^{(c)}_t=\{\,y_j : j \le t-c\,\},
\end{equation}
empty when $t\le c$. The capsule binds as $\Gamma_t=G(\mathcal H^{(c)}_t)$
with evidence state $E_t=\Phi(\mathcal H^{(c)}_t)$. Serial execution
recovers $\mathcal H^{(1)}_t=\{y_j : j\le t-1\}$, the largest history this
model permits. Each additional concurrent slot withholds one recent outcome
from every post-warm-up binding.

\paragraph{What concurrency withholds.}
Equation~\ref{eq:stale-history} gives the cost of concurrency a precise
form. For $c'>c$ and every binding instant $t>c'$,
$\mathcal H^{(c')}_t\subseteq\mathcal H^{(c)}_t$: raising concurrency from
$c$ to $c'$ removes the $c'-c$ most recent settled outcomes from every
post-warm-up binding, so a task under concurrency $c$ binds with an
evidence lag of $c-1$ outcomes. This subset relation follows from the
execution model alone. Whether it costs fitness depends on how much the
evidence state $E_t$ (Eq.~\ref{eq:evidence}) changes over a window of
$c-1$ outcomes, and that is an empirical question rather than a property
of the model.

Two features of the withheld entries matter for reading that question.
First, they are official outcomes settled by Protected Judgment and
distilled by Governance into ranks and potential estimates
(\S\ref{sec:structured-governance}). They are not the unchecked
single-attempt lessons that \S\ref{sec:introduction} identifies as the
source of premature narrowing in agent-centric memory. The concern there
is that a lesson enters shared memory with no owner to test it, so more of
it is not better. Here every withheld entry has already passed the check,
and Governance rather than the proposing agent decides how far it travels.
The loss under concurrency is therefore one of delayed evidence, not of
unvetted evidence. Second, the relation is local: it compares two
concurrencies on a common settled history. Full adaptive runs follow
concurrency-dependent trajectories, since later outcomes were themselves
produced under earlier stale contexts, so the end-to-end effect has to be
measured.

\paragraph{Measured effect.}
Figure~\ref{fig:concurrency-tradeoff} reports that measurement.
Degradation stays small through $c=4$ and grows at $c=8$, consistent with
the evidence lag $c-1$ crossing the horizon over which the evidence state
changes materially. Serial execution maximizes the evidence available at
each binding, while wall-clock time scales approximately as $T\bar\tau/c$
for mean task duration $\bar\tau$. Figure~\ref{fig:concurrency-tradeoff}
therefore identifies the practical operating point between these
objectives.

\subsection{Runtime Hooks}
\label{app:hooks}

This appendix details the enforcement layer of \S\ref{sec:open-proposal}.
Open Proposal grants broad authorship inside a task, and two things cannot
be left to that grant. First, the run's bookkeeping cannot rest on the
orchestrator's compliance, since the orchestrator is an autonomous agent
too. Second, the prohibition on reading evaluator sources guards the
interface to Protected Judgment, and a single file read would breach it. The
hook layer enforces both outside the agent, at the events the host runtime
emits before and after each tool call and at each subagent lifecycle
transition. Scheduler hooks maintain the task ledger. Permission hooks
decide individual tool calls.

\subsubsection{Interception and Dispatch}

\paragraph{The dispatch chain.}
Each hook event runs as a separate short-lived process. The host writes the
event payload to standard input as JSON and reads any response from standard
output. Inside that process a broker looks up the handlers registered for the
pair (host, event) and calls them in descending priority, chaining each
return value into the next call. A handler that raises is logged and skipped
while the rest still run, so one faulty rule cannot disable the layer.

\paragraph{Serialization.}
Handlers share run state across process boundaries, so in-process locks give
no protection. The broker takes a cross-process file lock around the whole
dispatch, which makes an event's read-modify-write of the ledger atomic
against concurrently spawning executors. On timeout the dispatch is abandoned
and logged; no handler runs and no partial state is written.

\subsubsection{The Task Ledger}
\label{app:ledger}

A run's tasks occupy one pool with the six states of
Table~\ref{tab:task-ledger}. Every transition is driven by a host event,
never by an agent's own report. A transition whose precondition fails is
rejected and logged, so duplicate or out-of-order events cannot corrupt the
pool.

\begin{table}[htbp]
  \centering
  \small
  \begin{tabular}{@{}lll@{}}
    \toprule
    State & Entered when & Event source \\
    \midrule
    \texttt{created}   & Governance mints identifiers at a convergence check   & control-plane call \\
    \texttt{pending}   & the hook allocates one to an observed executor spawn  & before the tool call \\
    \texttt{queued}    & the spawn returns an agent handle                     & after the tool call \\
    \texttt{running}   & the subagent starts                                   & subagent lifecycle \\
    \texttt{completed} & the subagent stops with a successful attempt on record & subagent lifecycle \\
    \texttt{failed}    & the subagent stops with no successful attempt on record & subagent lifecycle \\
    \bottomrule
  \end{tabular}
  \caption{States of the task ledger and the events that advance them. Only
  the first row originates in Governance; the host runtime reports the rest.}
  \label{tab:task-ledger}
\end{table}

\paragraph{Terminal states.}
The two terminal states distinguish a task that settled a valid candidate
from one that did not. When the executor stops, the hook consults the
attempt log rather than the executor: a task with a successful attempt on
record settles as \texttt{completed}. The \texttt{failed} state covers
every other stop, whether the attempts all failed verification, the
executor abandoned the trajectory, or it terminated early; a short failure
reason, taken from the executor's final message where the host supplies
it, is stored for reporting alone. Neither outcome drops a task
from the ledger, and both settle it for scheduling, so the loop replenishes
on either.

\paragraph{Scope of the ledger.}
The ledger counts tasks; it does not score them. Its fitness field copies
the evaluator's fitness from the successful attempt and serves reporting
alone. No selection, plateau, or budget decision reads it. The fitness that
drives evolution arrives on a separate path. The control-plane call that
records an attempt accepts no fitness argument: it reads the evaluator's
own result file and takes validity and fitness from there
(\S\ref{sec:protected-judgment}), and only a result the evaluator marked
successful with a numeric fitness enters the population that selection ranks.
A task none of whose attempts the evaluator marked successful settles as
\texttt{failed}, so it is counted but never selected from. No number an
agent reports therefore enters the ledger or the search.

\subsubsection{Identity and Budget}

\paragraph{Identity.}
An agent never states which task it is running. The hook resolves identity
in two steps. A host session identifier resolves to a run, recorded when the
orchestrator binds that run through the control plane. A host agent
identifier resolves to an entry in the run's agent index, written when the
subagent starts.

Identity is assigned in the opposite direction, from the hook to the
executor. On observing a spawn whose subagent type is the task executor, the
hook takes the next identifier from the pool and rewrites the spawn payload:
it sets the spawn description to that identifier and appends the run and
task identifiers to the prompt. The executor receives an identity it did not
choose, and its ledger entry predates it. The host echoes the rewritten
description in the post-call event, which lets the hook match the returned
agent handle to its task.

\paragraph{Budget.}
Governance mints identifiers at each convergence check, up to the requested
concurrency and within the run's total task budget. The count it returns
tells the orchestrator how many executors to spawn. This bound needs no
cooperation from the orchestrator. An orchestrator that spawns extra
executors finds an empty pool, so allocation yields nothing, those spawns
carry no task identity, and the budget stays intact. Minting also counts
tasks already created but not yet spawned, so unclaimed identifiers suppress
later minting instead of piling up.

\paragraph{The root barrier.}
The root phase uses the same minting rule as a barrier, with no separate
code path. Its mint ceiling equals the number of lineage roots. Once every
root has been minted the pool stays empty until roots settle, so the
orchestrator receives zero new tasks and keeps looping without needing to
know why. Settled roots lift the barrier, which puts a failed root on the
same footing as a successful one.

\subsubsection{Permission Control}

\paragraph{The decision pipeline.}
A permission hook runs before the tool executes. It normalizes the call into
a tool type with a small typed parameter set, resolves the caller through the
identity path above, and runs the pipeline registered for that caller's role.
Each rule declares a filter and a check, so inapplicable rules drop out
before any argument inspection.

The pipeline resolves by deny-overrides: a denial short-circuits and returns
its reason, and an endorsement stands only when no rule denies. A call that
no rule claims falls through to the host's own permission flow. That third
outcome keeps the layer additive: it only narrows what the host would
otherwise permit, never widens it, so one rule set works on any host
regardless of its native permission model. Denials are logged with the rule
that produced them.

\paragraph{The rule families.}
The pipeline carries two rule families. The first denies file reads, globs,
and greps whose target path resolves to the evaluator source. It also
confines the executor's shell to an allowlist: the evaluator command that
the benchmark contract names, run for the executor's own task, and the
creation of its own attempt directories. Every other shell command is
denied, so no shell path reaches the sources either. The second
governs another task's result directory, and it works from a whitelist: it
admits exactly the artifact paths Governance bound into the capsule and
denies every other path under another task's directory, including its
evaluation records and metrics. An executor can therefore read the parent and
inspirations it was granted, and nothing else of the history. The first rule
keeps judge with Protected Judgment; the second keeps govern with Structured
Governance.

\paragraph{Two denials in practice.}
Figures~\ref{fig:hook-denial} and~\ref{fig:hook-denial-runs} reproduce one
interception by each rule family, both from Circle Packing runs. In the
first, a task executor issued a \texttt{Read} whose path resolved to the
benchmark's \texttt{evaluator.py}. The pre-tool hook normalized the call,
resolved the caller to the task-executor role, and ran that role's
pipeline; the first rule family matched the path and returned \textsc{deny}
with its reason, and the hook answered the host with that decision. The
host dropped the call before it executed and handed the reason back to the
executor as the tool result, so the executor learns that the read is
prohibited and why. The task-executor prompt of
Appendix~\ref{app:prompt-executor} states the
same prohibition; this path holds when the prompt does not.

In the second, Governance had bound the executor of \texttt{task\_009} a
capsule with \texttt{task\_007} as parent and \texttt{task\_001},
\texttt{task\_002}, and \texttt{task\_006} as inspirations, so the
executor could read those four artifact directories and its own. It then
tried to read the evaluation record of \texttt{task\_008}, a sibling that
Governance had not selected for it. That path lies under another task's
directory and is not among the artifact paths bound into the capsule, so
the second rule family denied it. The reason names the whitelist in force:
the task's own directory and the \texttt{artifact\_ref} directories
returned by \texttt{select\_parent} and \texttt{select\_inspirations}.
What reaches a task about the rest of the history therefore remains what
Governance chose to bind, not what the executor decides to look up.

The two denials guard the two boundaries that Open Proposal could otherwise
erode. The first keeps judge with Protected Judgment: evaluator evidence
stays private, so the official score is never shaped by a proposer that has
read how it is computed. The second keeps govern with Structured
Governance: which parts of the history a task builds on is a selection
decision, and the executor cannot widen it on its own. In both cases the
executor retains full autonomy over construction inside its capsule and
none over the decisions the other two planes own.

\Needspace{16\baselineskip}
\begin{hookcard}{Permission hook}{one denial from a Circle Packing run; paths abbreviated}
\textbf{PreToolUse payload (host $\to$ hook)}\\
tool\_name: Read\\
tool\_input.file\_path: .../benchmarks/skills/circle\_packing/scripts/evaluator.py\\
caller: resolved from session\_id and agent\_id to role task-executor

\textbf{Pipeline log}\\
16:03:16 DEBUG pipeline.py[evaluate\_permission] permission: no\_evaluator\_access -> DENY
"Access to evaluator script '.../circle\_packing/scripts/evaluator.py' denied:
task-executors must NOT read the evaluator script."\\
16:03:16 INFO Final decision: DENY (handler: no\_evaluator\_access)

\textbf{Hook response (hook $\to$ host)}\\
\{"hookSpecificOutput": \{"hookEventName": "PreToolUse",
"permissionDecision": "deny",
"permissionDecisionReason": "Access to evaluator script ... denied:
task-executors must NOT read the evaluator script."\}\}
\end{hookcard}
\vspace{6pt}
{\captionsetup{hypcap=false}
\captionof{figure}{One interception by the first rule family. The host
drops the call before it executes and returns the stated reason to the
executor as the tool result.}
\label{fig:hook-denial}}
\medskip

\Needspace{16\baselineskip}
\begin{hookcard}{Permission hook}{one denial from a Circle Packing run; paths abbreviated}
\textbf{PreToolUse payload (host $\to$ hook)}\\
tool\_name: Read\\
tool\_input.file\_path: .../runs/circle\_packing\_20260922190350/tasks/task\_008/attempts/attempt\_001/eval\_result.json\\
caller: resolved from session\_id and agent\_id to role task-executor, bound to task\_009\\
capsule: parent task\_007; inspirations task\_001, task\_002, task\_006

\textbf{Pipeline log}\\
19:31:52 INFO pipeline.py[evaluate\_permission] Final decision: DENY, Read of
'.../tasks/task\_008/attempts/attempt\_001/eval\_result.json' denied: under runs/
task-executor may only read runs/circle\_packing\_20260922190350/tasks/task\_009/ and the
artifact\_ref directories returned by select\_parent/select\_inspirations.
(handler: task\_executor\_read\_allowlist\_under\_runs)

\textbf{Hook response (hook $\to$ host)}\\
\{"hookSpecificOutput": \{"hookEventName": "PreToolUse",
"permissionDecision": "deny",
"permissionDecisionReason": "Read of '.../task\_008/.../eval\_result.json' denied:
under runs/ task-executor may only read .../tasks/task\_009/ and the artifact\_ref
directories returned by select\_parent/select\_inspirations."\}\}
\end{hookcard}
\vspace{6pt}
{\captionsetup{hypcap=false}
\captionof{figure}{One interception by the second rule family. The executor
of \texttt{task\_009}, granted \texttt{task\_007} as parent and
\texttt{task\_001}, \texttt{task\_002}, and \texttt{task\_006} as
inspirations, attempted to read the evaluation record of
\texttt{task\_008}; the whitelist admits only its own directory and the
artifact paths Governance bound into its capsule, so the host drops the
call and returns the stated reason.}
\label{fig:hook-denial-runs}}
\medskip

\subsubsection{Host Portability}
\label{app:host-portability}

The ledger's state machine is shared across hosts. The scheduling adapter
maps native events onto the four lifecycle roles of
Table~\ref{tab:hook-events}, and permission interception runs through the
same pre-tool event; Appendix~\ref{app:host-integration}
states the full set a host must supply.

\begin{table}[htbp]
  \centering
  \small
  \begin{tabular}{@{}lll@{}}
    \toprule
    Role & Claude Code & OpenCode \\
    \midrule
    spawn observed  & \texttt{PreToolUse} on the agent tool & \texttt{tool.execute.before} on the task tool \\
    handle returned & \texttt{PostToolUse}                  & \texttt{tool.execute.after} \\
    subagent starts & \texttt{SubagentStart}                & \texttt{session.created} carrying a parent \\
    subagent ends   & \texttt{SubagentStop}                 & \texttt{session.idle} \\
    \bottomrule
  \end{tabular}
  \caption{Native events mapped onto the four lifecycle roles the ledger
  requires, shown for the two hosts used in this paper.}
  \label{tab:hook-events}
\end{table}

Hosts differ in how directly they expose these roles. OpenCode, for
example, supplies the last two only indirectly, and the adapter
reconstructs them. Its subagents are ordinary sessions, so a child announces
itself by carrying a parent session, and the parent identifier resolves the
run. Its post-call payload also reports the arguments as they stood before
the hook rewrote them, so the injected identifier never comes back through
the host. The adapter therefore keeps a registry keyed by call: allocation
records the run and task, and the completion of that same call retrieves
them. A second registry keyed by session lets the idle event recover which
task a stopping subagent held, and retiring the entry on use fires the
terminal transition once. The state machine itself never changes: whatever
events a host emits, the adapter's only job is to translate them into these
four roles.

\subsection{Judge in Existing Systems}
\label{app:judge-baselines}

Protected Judgment (\S\ref{sec:protected-judgment}) asks three things of a
system: evaluator sources stay unreadable from govern and propose, the
normalized outcome is the only channel back, and something outside the
agent holds that boundary. This appendix checks the five baselines of
\S\ref{sec:exp-setup} against these three conditions, reading the public
revisions our experiments start from: SkyDiscover revision \texttt{0d932b6}
(2026-09-14) for OpenEvolve, EvoX, and AdaEvolve~\citep{liu2026skydiscover},
and CORAL revision \texttt{0123dfb} (2026-09-06)~\citep{qu2026coral}. None
of the five meets the first condition, and three of them route the sources
into their own prompts. The closing paragraph states how the main results
close these paths.

\paragraph{Controller-centric systems.}
In SkyDiscover the evaluator is a path passed beside the initial program,
and the controller reads that path freely: no prompt forbids it, no
permission rule covers it, and a plain-Python evaluator is loaded into the
controller's own process. A configuration option
(\texttt{inject\_evaluator\_context}) prepends the evaluator source to the
proposer's system message; it resolves to on when no seed program is given
and to off otherwise, so the seeded Math, ADRS, and Frontier-CS runs of this
paper leave it off. Two of the three controllers read the source on another
path that no option guards. AdaEvolve's paradigm generator, the meta level
that issues breakthrough directions when the search stagnates, receives the
full evaluator source in a prompt section headed ``Evaluator Code (shows
how solutions are scored)'' and is instructed to analyze how solutions are
scored before proposing; the resulting directions are inserted into the
proposer's prompt. EvoX's operator generation receives the same source to
derive its diverge and refine guidance, and its strategy-evolution prompt
asks the model to trace how the combined score is computed; the derived
guidance is rendered into every proposer prompt. In both systems govern
reads judge's sources as a matter of design and forwards what it learns to
propose. OpenEvolve reads nothing on its own, and nothing stops it.

\paragraph{Agent-centric systems.}
CORAL's paper lists evaluator isolation among its execution safeguards,
with grader code held in a private directory that agents cannot inspect or
modify~\citep{qu2026coral}. In the released code the protection is
narrower than that description. The private directory holds the grader
environment and any answer keys, while the grader source itself is
surfaced into every agent worktree with read permission, and the
instruction file rendered for each agent points to it as the exact code
that scores the agent and invites the agent to read it. In the default
host session the private directory is guarded by a read rule on one tool
while shell access stays unrestricted; the optional container and sandbox
sessions mount the grader source read-only. CORAL's isolation therefore
protects the integrity of the grader and the secrecy of answer keys, and
leaves the scoring logic readable in every session type.

\paragraph{What the baselines share.}
In each of the five systems the boundary around judge is a component
boundary. A separate evaluator exists, nothing stops govern or propose
from reading it, and three systems route it into their own prompts.
Protected Judgment differs in what it protects and in who holds the line.
It protects the evidence, so the scoring logic itself stays out of reach,
and it enforces the boundary outside the agent through the permission
hooks of Appendix~\ref{app:hooks}. \S\ref{sec:leakage-analysis} measures
what the difference is worth: with the sources readable, the Original
score rises by 0.093 on average, while the out-of-distribution gain
shrinks to 0.009 and turns into a loss for CORAL-4 and EvoTrias.

\paragraph{Protocol for the main results.}
The scores of \S\ref{sec:main-results} are produced with every one of
these paths closed, so the comparison turns on search design and not on
evaluator access. For OpenEvolve, EvoX, and AdaEvolve we turn evaluator
injection off for the proposer and remove the evaluator source from the
meta-level prompts of AdaEvolve and EvoX. For CORAL we add a strict
prompt-level prohibition on reading evaluator sources to the agent
instruction file and extend the workspace guard described in its paper
into a hook that denies agent operations on the evaluator, including
read, write, and grep. The exposed setting of
\S\ref{sec:leakage-analysis} is the only configuration in which any
system can reach the sources (Appendix~\ref{app:exposure-protocol}).

\subsection{Discussion: What the Boundaries Cost and Buy}
\label{app:discussion}

The main text follows one line: which plane owns each decision. This
section follows three consequences of drawing those boundaries.

\paragraph{Bounded context.}
An agent-centric proposer reads the whole shared memory: every scored
attempt, every lesson, the full history of the run. This access helps in
one measurable way. On Frontier-CS, CORAL-4 leaves fewer near-zero scores
than EvoTrias (\S\ref{sec:main-results}), which the richer cross-trajectory
context plausibly helps to repair. The same access carries the cognitive
cost of \S\ref{sec:introduction}: one attempt's conclusion reaches every
agent unchecked, and the search narrows before the evidence justifies it.
It also carries a computational cost that grows with the run, since each
agent spends more of its budget re-reading a longer history
(\S\ref{sec:efficiency-analysis}). Open Proposal in EvoTrias never sees the
raw history. Structured Governance distills settled outcomes into posterior
evidence and constructs the capsule from it: a parent and inspirations
selected by the current policy, Notes that were validated on admission and
loaded by relevance, and Meta-guidance only when the search stalls
(\S\ref{sec:structured-governance}). What reaches a task has been checked
and scoped before it arrives. The results bear out this trade. EvoTrias
trails CORAL-4 only in repairing zero-scoring attempts, while it ranks
first on nine of the twelve Math and ADRS tasks under each backbone, leads
Frontier-CS in both
mean and median (\S\ref{sec:main-results}), and does so at 64\% lower token
cost (\S\ref{sec:efficiency-analysis}). A task receives a bounded context,
and the separation of govern from propose that bounds it returns far more
than it costs. The Level~1 ablations of Table~\ref{tab:ablation} then
measure how much each component of the capsule matters: removing any one
of them lowers the score on both tasks, so the information Governance
selects is the information Proposal needs. The converse holds as well:
handing the controller-centric baselines an agentic proposer without
changing their govern leaves every one of them below EvoTrias
(Appendix~\ref{app:agentic-baselines}).

\paragraph{Parallelism.}
Concurrency is where much of EvoTrias's efficiency comes from. The
wall-clock advantage in \S\ref{sec:efficiency-analysis} comes mainly from
parallel task execution, and \S\ref{sec:concurrency-analysis} shows that
$c=4$ cuts wall-clock time by 74\% at a fitness loss under 0.009, a
near-linear gain in end-to-end efficiency. How a design adds concurrency
therefore matters. In an agent-centric system, govern lives inside each
agent's trajectory, so adding a parallel proposer means replicating the
whole agent, and every replica re-reads history and plans its own moves. In
EvoTrias, govern and propose are separate: cross-round state lives in
Governance, scheduling in the hook-written ledger
(Appendix~\ref{app:hooks}), and a task executor needs only its capsule.
Adding a slot therefore means spawning one more executor subagent that
constructs one candidate and terminates (\S\ref{sec:open-proposal});
nothing else in the run changes. The two ways of scaling differ sharply in
cost. CORAL-4 spends 3.9$\times$ the tokens of CORAL-1 under the same
wall-clock budget (Figure~\ref{fig:sp-tokens}), whereas EvoTrias's token
cost varies by less than 5\% from $c=1$ to $c=8$. At the same wall-clock
budget and the same four-way width, EvoTrias spends 36\% of CORAL-4's
tokens (\S\ref{sec:efficiency-analysis}). What parallelism does cost is
evidence lag: a task binds its context before its siblings settle, and
Appendix~\ref{app:concurrency-staleness} bounds that lag at $c-1$ outcomes.
Concurrency therefore follows from the allocation itself.

\paragraph{Complementarity.}
The division brings one more benefit: each plane can improve along its own
line without touching the others. Proposal improves with stronger hosts and
backbones (Table~\ref{tab:ablation}, Level~2) and deploys as a plugin on
any host that meets the contracts (Appendix~\ref{app:host-integration}).
Governance improves with better search algorithms: the adaptive controllers
of \S\ref{sec:related-work} are candidate replacements for $G$, and the
default policy $\Pi_0$ already adapts population-based rules from
OpenEvolve and AdaEvolve (\S\ref{sec:structured-governance}). Judgment
improves with better evaluators behind the same request interface.

\subsection{Limitations}
\label{app:limitations}

Two limitations bound the present results. (1) Open Proposal runs on a
general-purpose coding agent, so construction quality tracks the backbone:
EvoTrias scores 0.7329 with GLM-5.2 and 0.8313 with Opus-5 on Signal
Processing, and every result is reported per backbone
(Appendix~\ref{app:backbone-gap}). (2) Coding-agent trajectories cost tokens.
A full 50-iteration run averages \$41.37, which is 64\% below four-agent
CORAL and still above the three controller-centric baselines
(\S\ref{sec:efficiency-analysis}); construction autonomy is what that margin
buys.

%% file: sections/generated/positive_notes_case.tex
\begin{positivenotecard}{P1}{Suppress reversals by post-hoc persistence pruning, not by narrowing bandwidth}
\textbf{Lesson.} Smoothness and accuracy trade off only when bandwidth is the sole knob. Adding persistence pruning of sub-noise extrema plus monotone PAVA projection after the zero-phase order-3 smoother cut reversals further than narrowing bandwidth, without the lag cost.

\textbf{Actionable guidance.} Keep the smoother bandwidth wide (corner coefficients \textasciitilde{}1.0-5.0 over W) and get smoothness from a separate persistence-pruning + PAVA projection stage; select bandwidth, adaptivity and pruning threshold jointly against a surrogate whose accuracy term is a Cp risk on the final pruned output.

\textbf{Evidence.} task\_015 fitness 0.8144 vs parent task\_007 0.7249 (+12.35\%), best inspiration task\_011 0.7705. slope\_changes 18.0 -\textgreater{} 5.4, false\_reversals 13.4 -\textgreater{} 3.2, lag\_error 0.208 -\textgreater{} 0.226, correlation 0.872 -\textgreater{} 0.817. Bandwidth-only task\_011 got slope\_changes 7.8 but lag\_error 0.294. The winning internal candidate (corner, gamma, tau) was not isolated.

\textbf{Confidence.} medium.
\end{positivenotecard}

\begin{positivenotecard}{P2}{Cut slope changes with a shape-domain post-stage, not a narrower band}
\textbf{Lesson.} Because smoothness\_score = 1/(1 + slope\_changes/20), reversals dominate fitness. After Whittaker smoothing, 4-sigma persistence pruning and PAVA cut reversals at near-zero correlation cost, unlike spectral narrowing.

\textbf{Actionable guidance.} Keep the adaptive Whittaker core and its \_C\_NOMINAL=2.2 corner unchanged; tune smoothness via the stage-2 persistence threshold (\_PERSIST\_K=4.0 times post-filter sigma) instead of lowering the cutoff.

\textbf{Evidence.} task\_016 attempt\_001 fitness 0.8007 vs parent task\_007 0.7249 (+10.46\%). slope\_changes 18.0-\textgreater{}7.8, false\_reversals 13.4-\textgreater{}5.0, smoothness 0.526-\textgreater{}0.719, correlation 0.872-\textgreater{}0.869, eval\_time 0.13s-\textgreater{}0.27s. Stage 2 runs unconditionally, no fallback branch.

\textbf{Confidence.} medium.
\end{positivenotecard}

\begin{positivenotecard}{P3}{Select L1 trend-filter lambda by surrogate sweep}
\textbf{Lesson.} On the same L1/ADMM + knot-refit estimator, replacing the parent's fixed discrepancy-principle lambda with a 12-rung geometric lambda ladder scored by an internal surrogate (0.58*smoothness + 0.34*agreement + 0.08*(1-lag)) raised fitness 0.654 -\textgreater{} 0.720. The idea was adapted from inspiration task\_014. Tightened knot extraction shipped in the same attempt, so the sweep is not isolated.

\textbf{Actionable guidance.} For a single-knob estimator, sweep a wide geometric ladder and keep the rung maximising a saturating-smoothness surrogate scored on the emitted slice; use the noise-matched rung only as the accuracy reference.

\textbf{Evidence.} task\_017/attempt\_001 fitness 0.7198 vs parent task\_009 0.6540 (+10.1\%). slope\_changes 20.0-\textgreater{}10.6, smoothness 0.500-\textgreater{}0.654, correlation 0.746-\textgreater{}0.756. Still below task\_014 (0.778).

\textbf{Confidence.} medium.
\end{positivenotecard}

\begin{positivenotecard}{P4}{Select reversal-pruning level by surrogate objective}
\textbf{Lesson.} Replacing the fixed kappa=3 pruning threshold with an explicit selector likely drove the gain: enumerate the full smallest-swing-first pruning path of a Kalman/RTS track over wide bandwidths (lam\_ml x \{1,4,16\}) and pick the pair maximizing 0.40*corr\_proxy + 0.32/(1+0.05*slope\_changes), with corr\_proxy a Mallows Cp estimate and a floor forbidding corr drops \textgreater{}0.06 below the unpruned track.

\textbf{Actionable guidance.} When the objective mixes observable slope changes with unobservable clean-signal correlation, enumerate pruning once and select with exact slope counts, Cp risk, and an accuracy floor; do not hand-tune a significance constant.

\textbf{Evidence.} task\_019/attempt\_001 fitness 0.7730 vs parent task\_004 0.5664 (+36.5\%) and best inspiration task\_008 0.7574. slope\_changes 24.4 -\textgreater{} 13.8, smoothness\_score 0.450 -\textgreater{} 0.592, correlation 0.976 -\textgreater{} 0.959. Costs: lag\_error 0.128 -\textgreater{} 0.436, noise\_reduction 0.869 -\textgreater{} 0.585, eval\_time 1.9s -\textgreater{} 3.8s.

\textbf{Confidence.} medium.
\end{positivenotecard}

\begin{positivenotecard}{P5}{Jointly sweep bandwidth and pruning with a Cp surrogate}
\textbf{Lesson.} Freeing the smoothing corner as a second knob alongside pruning level, rather than fixing it at C=3.55, improved every axis at once: slope changes 25.6 -\textgreater{} 13.0, false reversals 19 -\textgreater{} 7.4, correlation 0.911 -\textgreater{} 0.957, lag error 0.223 -\textgreater{} 0.155. The Cp accuracy term likely prevented the over-pruning seen in the pruning-only inspiration.

\textbf{Actionable guidance.} With a cheap O(n) core, jointly sweep bandwidth (1.60-7.50 over W) and the pruning path; score the emitted z[W-1:] slice by Cp accuracy, not against the candidate's stage-1 track.

\textbf{Evidence.} task\_026/attempt\_001 fitness 0.8047 vs parent task\_005 0.6976 (+15.3\%) and best inspiration task\_014 0.7776. The grid has 5 bandwidths and at most 26 pruning rungs; output is the argmax.

\textbf{Confidence.} medium.
\end{positivenotecard}

%% file: sections/generated/negative_notes_case.tex
\begin{negativenotecard}{N1}{Do not swap the L2 roughness post-filter for this unguarded L1 dead-zone + knot refit}
\textbf{Lesson.} Swapping the parent's L2 roughness post-filter for an L1 curvature dead-zone on the Kalman/RTS track (0.25-sigma displacement budget, knot refit at rel tol 2e-3 and 20\% knot cap) lost accuracy and smoothness simultaneously instead of trading one for the other; the permissive knot set added slope changes rather than removing them.

\textbf{Actionable guidance.} Avoid this unguarded variant; if trying L1 curvature post-filtering, use a much stricter knot tolerance and smaller knot cap, size the budget from the smoothed track's own residual scale, and keep stage 2 only if smoothness and correlation both improve.

\textbf{Evidence.} task\_013 attempt\_001 fitness 0.6261 vs parent task\_010 0.7137 (-12.3\%). Correlation 0.854 -\textgreater{} 0.688, smoothness 0.575 -\textgreater{} 0.372, slope\_changes 14.8 -\textgreater{} 33.8, false\_reversals 11.6 -\textgreater{} 26.2. Best reference is still pure Kalman/RTS with no post-filter (task\_003, 0.7381).

\textbf{Confidence.} medium.
\end{negativenotecard}

\begin{negativenotecard}{N2}{Gate L0 changepoint branches on reversal count, not just a 1.30x Cp-risk bar}
\textbf{Lesson.} Admitting L0/PELT segmented branches whenever their Cp risk stayed within 1.30x branch A's risk let a many-segment fit win on accuracy while tripling reversals; smoothness\_score = 1/(1+slope\_changes/20) dominates this objective.

\textbf{Actionable guidance.} If pairing a changepoint branch with the smoother, also require it to match branch A on slope\_changes/false\_reversals, or route it through the same persistence pruning; do not admit it on a Cp-risk ratio alone.

\textbf{Evidence.} task\_029/attempt\_001 fitness 0.7673 vs parent task\_015 0.8144 (-5.8\%). correlation 0.817-\textgreater{}0.943, avg\_error 0.414-\textgreater{}0.340, but slope\_changes 5.4-\textgreater{}15.0, false\_reversals 3.2-\textgreater{}12.8, smoothness\_score 0.787-\textgreater{}0.571, eval\_time 2.6s-\textgreater{}6.7s. No per-record branch tag in eval output, so the executed branch is unverified.

\textbf{Confidence.} medium.
\end{negativenotecard}

\begin{negativenotecard}{N3}{Guard cross-branch surrogate against reversal inflation when adding a changepoint branch}
\textbf{Lesson.} Adding an L0 changepoint branch beside the penalised smoother raised correlation (0.817 -\textgreater{} 0.943) but quadrupled slope\_changes (5.4 -\textgreater{} 21.4) and false\_reversals (3.2 -\textgreater{} 19.2), so smoothness\_score fell 0.787 -\textgreater{} 0.483 and fitness regressed 7.5\%. The cross-branch surrogate under-weighted the reversal terms.

\textbf{Actionable guidance.} If adding a changepoint branch, raise the reversal weights in the cross-branch surrogate (\_W\_SMOOTH \textgreater{}= 0.45, \_W\_FALSE\_REV \textgreater{}= 0.20) or veto branch B when its slope\_changes exceed the smoother branch's; do not rely on the correlation term alone to arbitrate.

\textbf{Evidence.} Child 0.754 vs parent 0.814. slope\_changes 21.4 vs 5.4, false\_reversals 19.2 vs 3.2, correlation 0.943 vs 0.817. Executed branch not recorded in eval\_result, so branch attribution is unverified.

\textbf{Confidence.} medium.
\end{negativenotecard}

\begin{negativenotecard}{N4}{Validate replacement selection criteria against realised correlation and noise\_reduction}
\textbf{Lesson.} A five-term plug-in criterion (reversals 0.36, purity 0.18, noise-corrected correlation 0.20, linear 1-risk/sigma\textasciicircum{}2 0.14, noise-floor-saturated tracking 0.12) replacing the lineage surrogate picked over-smoothed shapes: reversals halved while correlation and noise\_reduction both fell, fitness -5.29\%. Smoother and DP were unchanged, so the criterion configuration is implicated.

\textbf{Actionable guidance.} Before adopting a replacement selection criterion, check its realised slope\_changes, correlation and noise\_reduction against the parent; reuse an existing smoothed rung and cost table for any reference track rather than adding a separate DP pass.

\textbf{Evidence.} task\_044/attempt\_001 vs parent task\_027: fitness 0.7820 vs 0.8257 (-5.29\%); slope\_changes 2.5 vs 5.0; correlation 0.744 vs 0.821; noise\_reduction 0.464 vs 0.641; success\_rate 0.8 vs 1.0. Estimator code identical to parent.

\textbf{Confidence.} medium.
\end{negativenotecard}

%% file: sections/appendix/b_experimental_setup.tex
\section{Experimental Setup Details}
\label{app:experiment-details}

\subsection{Evaluation Protocol}
\label{app:eval-protocol}

Every method scores its candidates through the same evaluator for each
task, so a reported fitness means the same thing across systems. The
evaluator accepts an artifact and returns the normalized outcome of
Equation~\ref{eq:search-objective}. The protected setting of
\S\ref{sec:protected-judgment} is the default for the main results, so no
method may read evaluator sources; the exposed setting of
\S\ref{sec:leakage-analysis} is the deliberate exception, and it is the only
configuration in which any method can read them.

The search itself optimizes one scalar fitness per task, while
Tables~\ref{tab:combined-mean-results} and~\ref{tab:mean-best-results} report
each task in the unit its own benchmark states. The two coincide on most tasks
and differ on those whose evaluator normalizes against a published reference
value or inverts a cost. Table~\ref{tab:fitness-map} gives the conversion, so
the internal fitness values quoted in the case studies of
Appendix~\ref{app:additional-results} can be read against the main tables.

\begin{table}[htbp]
  \centering
  \caption{Reported quantity and internal fitness for each task. The arrow
  marks the direction of the reported quantity $R$; the internal fitness $f$ is
  always larger-is-better. The constants are the published reference values the
  evaluator normalizes against.}
  \label{tab:fitness-map}
  \footnotesize
  \setlength{\tabcolsep}{8pt}
  \renewcommand{\arraystretch}{1.08}
  \begin{tabular}{@{}lll@{}}
    \toprule
    Task & Reported quantity $R$ & Internal fitness $f$ \\
    \midrule
    \multicolumn{3}{@{}l}{\textit{Math}} \\
    \quad Circle Packing (square)    & sum of 26 radii $\uparrow$ & $R/2.635$ \\
    \quad Circle Packing (rectangle) & sum of 21 radii $\uparrow$ & $R/2.365832$ \\
    \quad Heilbronn Triangles        & normalized minimum triangle area $\uparrow$ & $R/0.0365299$ \\
    \quad Heilbronn Convex           & normalized minimum triangle area $\uparrow$ & $R/0.0309369$ \\
    \quad MinMax Distance            & squared min-to-max distance ratio $\uparrow$ & $R/0.2400471$ \\
    \quad Signal Processing          & composite filtering score $\uparrow$ & $R$ \\
    \quad Third Autocorrelation      & third autocorrelation constant $\downarrow$ & $1.4556428/R$ \\
    \addlinespace
    \multicolumn{3}{@{}l}{\textit{ADRS}} \\
    \quad Cloudcast                  & total transfer cost $\downarrow$ & $1/(1+R)$ \\
    \quad EPLB                       & expert-balance and speed score $\uparrow$ & $R$ \\
    \quad LLM-SQL                    & cache-hit and runtime score $\uparrow$ & $R$ \\
    \quad Prism                      & inverse KV-cache pressure score $\uparrow$ & $R$ \\
    \quad TXN                        & inverse-makespan schedule score $\uparrow$ & $R$ \\
    \addlinespace
    \multicolumn{3}{@{}l}{\textit{Frontier-CS}} \\
    \quad All 30 problems            & judge score on a 0--100 scale $\uparrow$ & $R$ \\
    \bottomrule
  \end{tabular}
\end{table}
\FloatBarrier

\subsection{Budget Alignment Across Baselines}
\label{app:coral-budget}

For the controller-centric baselines, we match the outer search budget exactly:
EvoTrias, OpenEvolve, EvoX, and AdaEvolve each receive 50 proposal slots. A slot
can add at most one valid candidate. After the first attempt, generation or
evaluation failure can trigger up to two repair-and-re-evaluation retries, but
every method stops at the first valid result; retries are never used to improve
an already valid score. These rules match the 50 opportunities that can add
evolved candidates to the search state; retries only recover invalid executions.

Table~\ref{tab:evaluator-calls} verifies that this rule holds in the runs
behind the main results by counting the evaluator calls each method actually
made. Every method lands within a narrow band above the 50-slot floor, the
excess being the retries that recovered invalid executions, and EvoTrias sits
at or below the baselines' band on every suite. A proposal slot therefore
buys the same number of official evaluations for every iteration-based
method; what differs is only how the candidate inside that slot is
constructed, which is the design decision under test.

\begin{table}[htbp]
  \centering
  \caption{Mean evaluator calls per run under the shared 50-slot budget
  (GLM-5.2; three runs per Math and ADRS task, one run per Frontier-CS
  problem). Each slot admits at most one valid candidate and at most two
  repair retries, so a run makes between 50 and 150 calls; the realized
  means show that retries are rare and comparable across methods.}
  \label{tab:evaluator-calls}
  \footnotesize
  \setlength{\tabcolsep}{10pt}
  \renewcommand{\arraystretch}{1.12}
  \begin{tabular}{@{}lcccc@{}}
    \toprule
    Suite & OpenEvolve & EvoX & AdaEvolve & EvoTrias \\
    \midrule
    Math ($n=21$ runs)         & 55.47 & 57.81 & 58.76 & 55.81 \\
    ADRS ($n=15$ runs)         & 57.60 & 57.40 & 54.53 & 52.87 \\
    Frontier-CS ($n=30$ runs)  & 61.13 & 56.37 & 55.13 & 56.80 \\
    \bottomrule
  \end{tabular}
\end{table}
\FloatBarrier

CORAL has no outer proposal loop. For each task and backbone, we therefore
match CORAL-1 and CORAL-4 to EvoTrias's mean wall-clock runtime under the
50-slot budget. Table~\ref{tab:coral-budget} reports these shared limits.

\begin{table}[htbp]
  \centering
  \caption{Per-task wall-clock budgets shared by EvoTrias, CORAL-1, and
  CORAL-4.}
  \label{tab:coral-budget}
  \footnotesize
  \setlength{\tabcolsep}{12pt}
  \renewcommand{\arraystretch}{1.12}
  \begin{tabular}{@{}lcc@{}}
    \toprule
    Problem & \textbf{GLM-5.2 (h)} & \textbf{Opus-5 (h)} \\
    \midrule
    \multicolumn{3}{@{}l}{\textit{Math}} \\
    \quad Circle Packing Square    & 1.87 & 2.67 \\
    \quad Circle Packing Rectangle & 1.84 & 2.11 \\
    \quad Heilbronn Triangles      & 1.81 & 1.90 \\
    \quad Heilbronn Convex         & 2.54 & 3.16 \\
    \quad MinMax Distance          & 3.04 & 3.09 \\
    \quad Signal Processing        & 2.18 & 3.01 \\
    \quad Third Autocorrelation    & 2.70 & 2.59 \\
    \addlinespace
    \multicolumn{3}{@{}l}{\textit{ADRS}} \\
    \quad Cloudcast                & 1.94 & 2.88 \\
    \quad EPLB                     & 2.37 & 2.39 \\
    \quad LLM-SQL                  & 3.18 & 3.75 \\
    \quad Prism                    & 3.39 & 3.90 \\
    \quad TXN                      & 2.03 & 2.48 \\
    \addlinespace
    \multicolumn{3}{@{}l}{\textit{Frontier-CS}} \\
    \quad Average ($n=30$) & 4.81 & -- \\
    \bottomrule
  \end{tabular}
\end{table}
\FloatBarrier

The two CORAL configurations bracket the concurrency comparison. CORAL-1 uses
one top-level trajectory, whereas EvoTrias can keep $c=4$ bounded task
executors in flight, so proposal width favors EvoTrias in that comparison.
CORAL-4 removes this asymmetry and makes the resource allocation conservative
for EvoTrias: it runs four complete top-level agents in parallel and returns
their best final score, while EvoTrias retains one top-level orchestrator and
implements four-way concurrency only through its task-executor subagents.
Under the same wall-clock limit and nominal four-way width, CORAL-4 consumed
about $2.8\times$ as many tokens (Figure~\ref{fig:sp-tokens}). This ratio is the
realized cost of the matched protocol, not an intrinsic constant of CORAL; it
shows that CORAL-4 received the larger aggregate agent-compute allocation.

\paragraph{What CORAL-4 is a comparison against.}
The asymmetry deserves to be stated plainly, because it decides how the
paired statistics of Appendix~\ref{app:mean-best-results} and
\ref{app:frontier-analysis} should be read. CORAL-4 is not one evolution
system with four workers. It is four complete evolution systems, each a
full top-level CORAL agent with its own trajectory, its own read and write
access to the shared memory, and its own freedom to spawn subagents, run
for the entire wall-clock budget, after which the best of the four final
scores is reported. That is a best-of-four ensemble over independent runs,
and it enjoys two advantages that no single system in the comparison has:
four times the top-level agent compute, and a final max over four outcomes
that suppresses the run-to-run variance every other entry carries in full.
EvoTrias, in contrast, is one system: one orchestrator, one Governance
state, one budget of 50 proposal slots, and a single final score with no
selection over repeated runs. Its subagents are not four agents; they are
bounded task executors that construct one candidate each inside a capsule
and terminate.

Under this protocol, any lead over CORAL-4 is a lead over a four-way
ensemble at $2.8\times$ the token cost, and any tie is a tie against it.
EvoTrias leads CORAL-4 on 9 of 12 Math and ADRS tasks with no ties, and
achieves a geometric-mean gain of 0.640\% with a positive confidence
interval (Table~\ref{tab:math-adrs-paired-stats}); on
Frontier-CS it raises the mean by 3.49 points and the median by 13.53
(Tables~\ref{tab:frontier-perproblem} and~\ref{tab:frontier-paired-stats}). The exact tests against CORAL-4
are the only ones in the paper that do not reach $p<0.05$, and that is what
one should expect when a single system is measured against a best-of-four
ensemble of a strong competitor: the protocol was chosen to stack the
comparison against EvoTrias, not to produce a significant margin. The
result to take from these tables is that one EvoTrias run matches or beats
the best of four CORAL runs while spending 36\% of their combined tokens.
A comparison at equal top-level agent count is CORAL-1, and there EvoTrias
wins all 12 Math and ADRS tasks and 15 of 30 Frontier-CS problems with 2
losses, with every confidence interval positive.

\subsection{Hyperparameter Configuration}
\label{app:hyperparameters}

Table~\ref{tab:hyperparams} lists EvoTrias's default operating settings,
grouped by module; controlled ablations override only the setting named in
their description. Every value was fixed before the benchmark runs and is
shared by all runs in this paper; no constant was tuned per task or per
benchmark. A dash indicates that the setting has no separate
mathematical symbol. Annealed parameters are written as start~$\rightarrow$~end
over the task budget.

\begin{center}
\begin{minipage}{\textwidth}
  \centering
  \captionsetup{hypcap=false}
  \captionof{table}{Default EvoTrias operating settings, grouped by module.}
  \label{tab:hyperparams}
  \footnotesize
  \begin{tabular}{@{}llc@{}}
    \toprule
    Parameter & Symbol & Value \\
    \midrule
    \multicolumn{3}{@{}l}{\textit{Scheduling and budget}} \\
    \quad Initial exploration directions (roots) & $n_{\mathrm{dir}}$ & 4 \\
    \quad Concurrent task slots & $c$ & 4 \\
    \quad Task budget (iterations) & $T$ & 50 \\
    \addlinespace
    \multicolumn{3}{@{}l}{\textit{Lineage and parent selection}} \\
    \quad Lineage temperature (annealed) & $\tau_t$ & $1.5\rightarrow0.5$ \\
    \quad Exploit probability & $p_{\mathrm{exploit}}$ & 0.7 \\
    \quad Quality bonus strength & $\eta$ & 0.3 \\
    \quad Local exploration (annealed) & $c_{\mathrm{local},t}$ & $1.5\rightarrow0.5$ \\
    \quad Elite rank fraction & $\kappa$ & 0.15 \\
    \quad Elite selection shield & $s$ & 0.6 \\
    \quad Potential prior & $\alpha,\beta$ & $(1,3)$ \\
    \quad Lineage potential weight & $\lambda_L$ & 0.10 \\
    \quad Parent potential weight & $\lambda_P$ & 0.03 \\
    \addlinespace
    \multicolumn{3}{@{}l}{\textit{Inspiration selection}} \\
    \quad Maximum inspiration slots & $|I_t|$ & 3 \\
    \quad Cross-lineage rate & $\mu$ & 0.2 \\
    \quad Foreign elite pool per lineage & $n_e$ & 2 \\
    \quad Sampling temperature (annealed) & $T_t$ & $1.5\rightarrow0.5$ \\
    \quad Rebalance entropy threshold & $H^\star$ & 0.9 \\
    \quad Rebalance entropy floor & $H_{\min}$ & 0.5 \\
    \quad Rebalance max strength & $\gamma_{\max}$ & 2.0 \\
    \addlinespace
    \multicolumn{3}{@{}l}{\textit{Notes}} \\
    \quad Delta quantile (annealed) & $q$ & $0.80\rightarrow0.60$ \\
    \quad Relative-delta threshold range & $\tau_t^+,\tau_t^-$ & $[2\%,5\%]$ \\
    \quad Near-best gate (annealed) & $r_t$ & $0.80\rightarrow1.00$ \\
    \addlinespace
    \multicolumn{3}{@{}l}{\textit{Meta-guidance}} \\
    \quad Plateau window & $w$ & 5 \\
    \quad Plateau patience & $P_{\mathrm{stall}}$ & 4 \\
    \quad Maximum directions per request & -- & 3 \\
    \bottomrule
  \end{tabular}
\end{minipage}
\end{center}

\FloatBarrier
\subsection{Baseline Configurations}
\label{app:baseline-configs}

OpenEvolve, EvoX, and AdaEvolve run from the reference implementations and
default hyperparameters of the SkyDiscover
framework~\citep{liu2026skydiscover}, with one change: we set the island count
to four, so that every controller-centric baseline opens with the same number
of independent search branches as EvoTrias's $n_{\mathrm{dir}}=4$ initial directions
(Appendix~\ref{app:init-directions}). CORAL-1 and CORAL-4 keep their published
agent design and run on the same coding-agent host as EvoTrias, Claude Code
v2.1.201, so the comparison between the two agent-centric configurations and
EvoTrias turns on the architecture rather than on the agent; Appendix~\ref{app:coral-budget}
details their budget and concurrency controls. The host ablations of
\S\ref{sec:ablation} replace that host with OpenCode v1.18.25
and change nothing else. Every method receives the same task statement, the
same initial reference program, and the same evaluator
(Appendix~\ref{app:eval-protocol}), and none of them may read evaluator
sources outside the exposed setting of \S\ref{sec:leakage-analysis}.

%% file: sections/appendix/c_additional_results.tex
\section{Additional Results and Analysis}
\label{app:additional-results}

\subsection{Math and ADRS: Statistical Results}
\label{app:mean-best-results}

Table~\ref{tab:mean-best-results} adds the best run and the sample standard
deviation to Table~\ref{tab:combined-mean-results} for the five strongest
methods; OpenEvolve is omitted, since it places last on almost every task of
Table~\ref{tab:combined-mean-results} and its spread adds little to the
comparison. Figure~\ref{fig:variance-spread}
rescales its Std rows into the coefficient of variation, so that tasks orders of
magnitude apart share one axis; the 20 of the 120 task--method pairs whose three
runs returned an identical score have zero Std and sit in the lane at the left.
Under
GLM-5.2, EvoTrias is the most stable of the five methods, with median
$1.8\times10^{-3}$ against $3.8\times10^{-3}$ for the next. Under Opus-5 the
comparison loses its grip: all three runs return the same score in 19 of that
backbone's 60 pairs, leaving too little spread to rank the methods by.

\input{sections/mean_best_results_table}

\vspace{3pt}
\begin{center}
\begin{minipage}{\textwidth}
  \centering
  \includegraphics[width=0.985\textwidth]{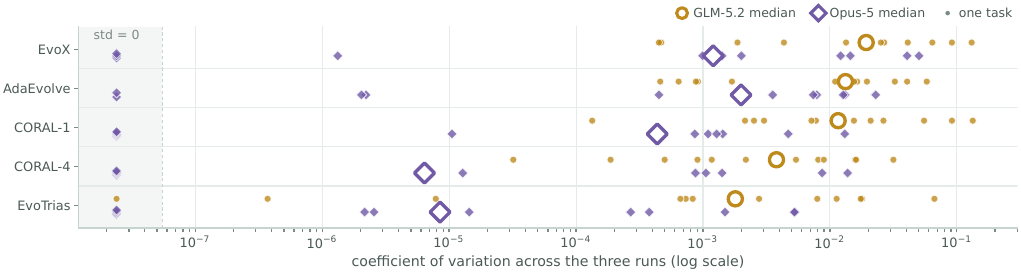}
  \captionsetup{hypcap=false}
  \captionof{figure}{Run-to-run spread behind the Std rows of
  Table~\ref{tab:mean-best-results}, as the coefficient of variation. Small
  marks are the 12 tasks, large open marks their median.}
  \label{fig:variance-spread}
\end{minipage}
\end{center}

\paragraph{Paired task-level analysis.}
We complement the raw-score table with a scale-invariant comparison across all
12 tasks. For task $t$, backbone $b$, and baseline $m$, let
$e_{tbm}=d_t\log(\bar{s}_{tb,\mathrm{EvoTrias}}/
\bar{s}_{tbm})$, where $d_t=1$ for higher-is-better tasks and $d_t=-1$ for
Third Autocorrelation and Cloudcast. We average $e_{tbm}$ over the two
backbones within each task and report
$100[\exp(\operatorname{mean}_t e_{tm})-1]$ as the geometric-mean relative
gain. The 95\% confidence intervals resample the 12 paired tasks 200,000 times.
We also apply an exact two-sided sign-flip test to the task effects and use
Holm correction across the five baselines; the confidence intervals are
per-comparison intervals.

\begin{table}[htbp]
  \centering
  \caption{Paired task-level comparison across Math and ADRS. Positive gains
  favor EvoTrias. W/T/L counts the signs of the 12 backbone-averaged task
  effects.}
  \label{tab:math-adrs-paired-stats}
  \footnotesize
  \setlength{\tabcolsep}{7.0pt}
  \renewcommand{\arraystretch}{1.12}
  \begin{tabular*}{0.92\textwidth}{@{\extracolsep{\fill}}lrrrrr@{}}
    \toprule
    Baseline & Rel. gain & 95\% CI & W/T/L & Exact $p$ & Holm $p$ \\
    \midrule
    OpenEvolve & +6.403\% & $[+2.926,+10.270]\%$ & 12/0/0 & 0.000488 & 0.00244 \\
    EvoX       & +3.010\% & $[+1.502,+4.596]\%$  & 12/0/0 & 0.000488 & 0.00244 \\
    AdaEvolve  & +2.047\% & $[+0.845,+3.422]\%$  & 12/0/0 & 0.000488 & 0.00244 \\
    CORAL-1    & +1.704\% & $[+0.784,+2.756]\%$  & 12/0/0 & 0.000488 & 0.00244 \\
    CORAL-4    & +0.640\% & $[+0.010,+1.382]\%$  & 9/0/3  & 0.1646   & 0.1646 \\
    \bottomrule
  \end{tabular*}
\end{table}

EvoTrias improves on OpenEvolve, EvoX, AdaEvolve, and CORAL-1 on all 12 tasks.
The gains range from 1.704\% to 6.403\%, every confidence
interval is positive, and all four comparisons remain significant after Holm
correction ($p=0.00244$). EvoTrias also leads CORAL-4 in aggregate by 0.640\%
and on 9 of 12 tasks. This is a strong efficiency result: CORAL-4 selects the
best outcome from four parallel top-level agents, while EvoTrias uses one
top-level agent and coordinates native subagents within it.

\FloatBarrier

\subsection{Where Backbone Strength Shows Up}
\label{app:backbone-gap}

Table~\ref{tab:mean-best-results} shows that the two backbones separate on some
tasks and not others. Comparing final scores alone cannot say why. This
subsection follows one GLM-5.2 run and one Opus-5 run through the budget on two
tasks of differing scope, using the split that \S\ref{sec:ablation} draws:
Circle Packing targets one fixed 26-circle arrangement, while Signal Processing
optimizes a filtering algorithm over a family of signals. Fitness in this
subsection is the internal score of Table~\ref{tab:fitness-map}; on Circle
Packing the internal 1.000373 is 2.63598 on the raw scale of
Table~\ref{tab:combined-mean-results}.

\begin{center}
\begin{minipage}{\textwidth}
  \centering
  \includegraphics[width=0.985\textwidth]{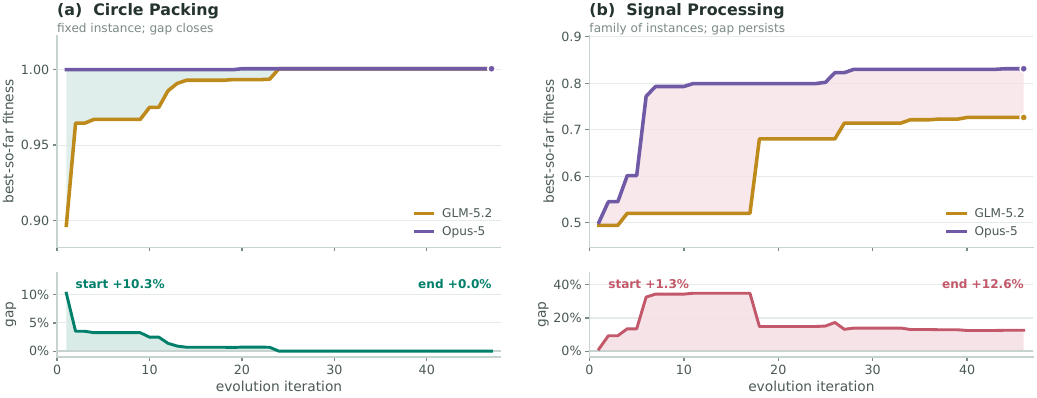}
  \captionsetup{hypcap=false}
  \captionof{figure}{Best-so-far fitness for one GLM-5.2 run and one Opus-5 run
  on each task, with the gap between them shaded and repeated below each panel
  as a share of the leading run's score at that iteration.}
  \label{fig:backbone-gap}
\end{minipage}
\end{center}

\paragraph{Two opposite regimes.}
Throughout this subsection a gap is reported as the difference between the two
runs divided by the leading run's score at that iteration.
On Circle Packing the stronger backbone opens with a large lead: its first
settled candidate already scores 0.9997 against 0.897, a gap of 10.3\%.
Evolution then removes that lead. The gap falls below 1\% by iteration 18 and
reaches zero by iteration 24, and both runs finish on the same value of
1.000373. On Signal Processing the two runs instead start level, 0.4945 against
0.5010, a gap of 1.3\%. Here the gap is opened by the search rather than
inherited from it: it widens to 34.9\% by iteration 12, and although GLM-5.2
recovers part of it later, 12.6\% still remains at the end.

\paragraph{Why scope decides this.}
Circle Packing has a reachable optimum, and both backbones arrive within the
budget, so backbone strength buys a better starting point and nothing more; the
search absorbs the difference. Signal Processing offers no such ceiling within
50 iterations, so a run keeps finding headroom, and how much of it the run
captures tracks the backbone. Backbone strength therefore acts on where a run
begins in the first case, and on how far it climbs in the second.

\paragraph{Corroboration.}
One run per cell cannot separate these trajectories from run-to-run noise, so we
read them against Table~\ref{tab:mean-best-results}. Over three runs each,
EvoTrias scores 0.7329 with GLM-5.2 and 0.8313 with Opus-5 on Signal Processing,
an 11.8\% gap, against 2.63484 and 2.63599 on Circle Packing, a gap of 0.043\%.
Both figures agree with the single-run trajectories that the backbone gap is
large on Signal Processing and negligible on Circle Packing.

\subsection{Frontier-CS: Statistical Results}
\label{app:frontier-analysis}

Table~\ref{tab:frontier-perproblem} lists the score of every method on each
of the 30 Frontier-CS problems; cell shading follows the score, from pale teal
(low) to deep teal (high). The corresponding distribution-level
comparison appears in Figures~\ref{fig:frontier-dist}
and~\ref{fig:frontier-ecdf}.

\begin{table}[t]
  \centering
  \caption{Per-problem Frontier-CS scores (GLM-5.2). Cell shading follows
  the shared 0--100 scale shown at right.}
  \label{tab:frontier-perproblem}
  \scriptsize
  \setlength{\tabcolsep}{6.4pt}
  \renewcommand{\arraystretch}{1.32}
  \makebox[\textwidth][c]{%
    \begin{minipage}[t]{0.84\textwidth}
      \vspace{0pt}
      \centering
      \input{sections/frontier_scores_table}    \end{minipage}%
    \hspace{0.5em}%
    \begin{minipage}[t]{0.085\textwidth}
      \vspace{0.16in}
      \includegraphics[height=4.71in]{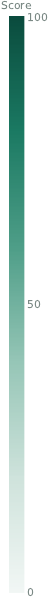}%
    \end{minipage}%
  }
\end{table}

\paragraph{Paired problem-level analysis.}
Unlike Math and ADRS, all Frontier-CS outcomes share the same 0--100 scale and
each method has one run per problem. We therefore use the raw paired
difference on each of the 30 problems. Mean-difference confidence intervals
resample complete problem rows 200,000 times, preserving all method scores in
each pair. The reported $p$-values come from two-sided paired label swaps with
1,000,000 samples and are Holm-corrected across the five baselines.

\begin{table}[htbp]
  \centering
  \caption{Paired Frontier-CS comparisons. Positive score differences favor
  EvoTrias; W/T/L counts the 30 problem-level outcomes.}
  \label{tab:frontier-paired-stats}
  \footnotesize
  \setlength{\tabcolsep}{7.0pt}
  \renewcommand{\arraystretch}{1.12}
  \begin{tabular*}{0.92\textwidth}{@{\extracolsep{\fill}}lrrrrr@{}}
    \toprule
    Baseline & Mean $\Delta$ & 95\% CI & W/T/L & Paired $p$ & Holm $p$ \\
    \midrule
    OpenEvolve & +9.44  & $[+3.65,+16.10]$ & 14/10/6 & 0.00348  & 0.01045 \\
    EvoX       & +8.76  & $[+1.58,+16.86]$ & 11/11/8 & 0.03377  & 0.06755 \\
    AdaEvolve  & +10.13 & $[+4.44,+16.79]$ & 13/12/5 & 0.000866 & 0.00433 \\
    CORAL-1    & +8.50  & $[+3.34,+14.57]$ & 15/13/2 & 0.00247  & 0.00989 \\
    CORAL-4    & +3.49  & $[-0.44,+8.06]$  & 10/12/8 & 0.1310   & 0.1310 \\
    \bottomrule
  \end{tabular*}
\end{table}

The paired analysis confirms EvoTrias's distribution-level lead. Its mean
gain is 8.50--10.13 points over OpenEvolve, AdaEvolve, and CORAL-1, with
positive confidence intervals and Holm-corrected $p\leq0.01045$. The gain over
EvoX is 8.76 points with a positive confidence interval. Against CORAL-4,
EvoTrias raises the mean by 3.49 points and the median by 13.53 points, despite
using one top-level agent rather than CORAL-4's best-of-four top-level-agent
selection. Together with the 65.42 mean and 80.67 median in
Figure~\ref{fig:frontier-dist}, these results show that governed subagent
parallelism converts the available agent compute into stronger high-scoring
solutions.

\FloatBarrier

\subsection{Best-so-Far Optimization Trajectories}
\label{app:iteration-trajectories}

\begin{figure*}[t]
  \centering
  \captionsetup[subfigure]{font=footnotesize,skip=1pt}
  \begin{subfigure}[t]{0.485\textwidth}
    \centering
    \includegraphics[width=\linewidth]{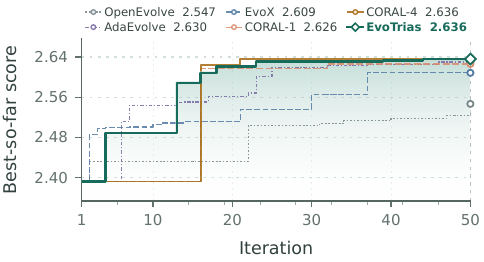}
    \caption{Circle Packing (square).}
  \end{subfigure}\hfill
  \begin{subfigure}[t]{0.485\textwidth}
    \centering
    \includegraphics[width=\linewidth]{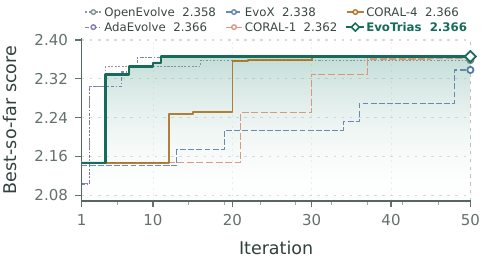}
    \caption{Circle Packing (rectangle).}
  \end{subfigure}

  \vspace{0.15em}
  \begin{subfigure}[t]{0.485\textwidth}
    \centering
    \includegraphics[width=\linewidth]{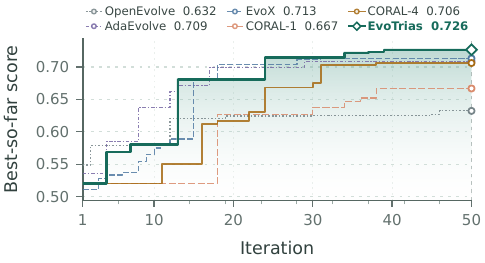}
    \caption{Signal Processing.}
  \end{subfigure}\hfill
  \begin{subfigure}[t]{0.485\textwidth}
    \centering
    \includegraphics[width=\linewidth]{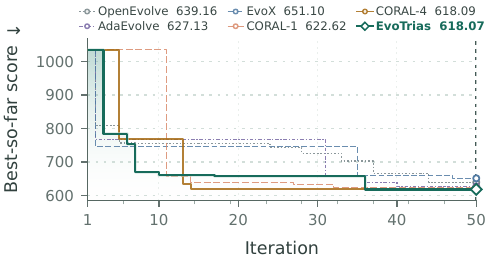}
    \caption{Cloudcast ($\downarrow$).}
  \end{subfigure}

  \vspace{0.15em}
  \begin{subfigure}[t]{0.485\textwidth}
    \centering
    \includegraphics[width=\linewidth]{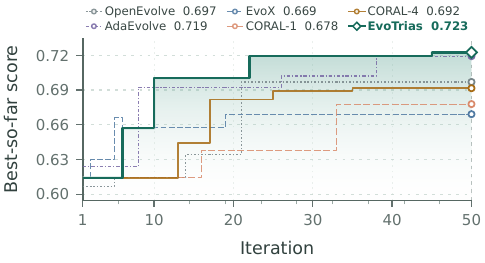}
    \caption{LLM-SQL.}
  \end{subfigure}\hfill
  \begin{subfigure}[t]{0.485\textwidth}
    \centering
    \includegraphics[width=\linewidth]{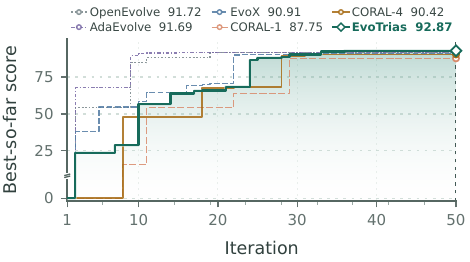}
    \caption{Frontier-CS 258.}
  \end{subfigure}

  \vspace{0.15em}
  \begin{subfigure}[t]{0.485\textwidth}
    \centering
    \includegraphics[width=\linewidth]{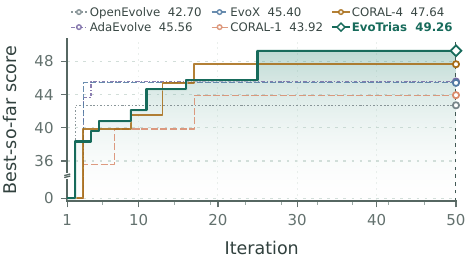}
    \caption{Frontier-CS 255.}
  \end{subfigure}\hfill
  \begin{subfigure}[t]{0.485\textwidth}
    \centering
    \includegraphics[width=\linewidth]{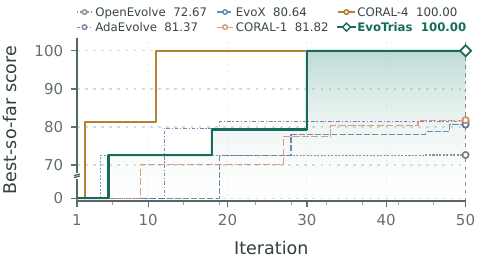}
    \caption{Frontier-CS 252.}
  \end{subfigure}

  \caption{Best-so-far fitness over 50 evolution iterations. The Math and ADRS
  panels show the median-final-score run among three; the Frontier-CS panels
  show the single run of that problem. Higher
  is better except for Cloudcast. CORAL observations are aligned by elapsed
  time to EvoTrias iterations. The Frontier-CS panels preserve zero-score
  prefixes while compressing the unused score interval.}
  \label{fig:iteration-trajectories}
\end{figure*}

Figure~\ref{fig:iteration-trajectories} plots best-so-far fitness over 50
evolution iterations for eight tasks. For the Math and ADRS panels we select
the run with the median final score among the three; the Frontier-CS panels
have a single run per problem (\S\ref{sec:exp-setup}). Because CORAL has no
outer iteration counter, its observations are aligned to the EvoTrias iteration
with matching elapsed time.

\FloatBarrier
\subsection{Controller-Centric Baselines with an Agentic Proposer}
\label{app:agentic-baselines}

The controller-centric baselines of \S\ref{sec:main-results} propose with a
single model call. This subsection asks how much of their gap to EvoTrias
closes once they, too, receive an agentic proposer. SkyDiscover ships such a
mode: its \texttt{--agentic} flag replaces the single call with a multi-turn
tool-calling loop that reads and searches files before it outputs the
candidate program. We enable this flag for OpenEvolve, AdaEvolve, and EvoX
and change nothing else. The controller, the task statement, the seed
program, the evaluation service, and the 50-iteration budget are those of
\S\ref{sec:exp-setup}, so the two configurations of each system differ only
in how propose is realized. Table~\ref{tab:agentic-baselines} reports three
runs per cell with GLM-5.2 on the two ablation tasks of
Table~\ref{tab:ablation} and one further Math task, Heilbronn Triangles.

\input{sections/appendix/c_agentic_baselines_table}

Two patterns follow. The first is that the gain from the agentic proposer
depends on the controller behind it. OpenEvolve improves on every task, by
2.9\% on Circle Packing, 6.8\% on Signal Processing, and 20.1\% on
Heilbronn Triangles. For AdaEvolve and EvoX the same switch moves the score
by between $-1.2\%$ and $+1.9\%$, with no consistent sign. OpenEvolve
leaves headroom that a proposer able to inspect the seed program and its
context can take on its own. The two adaptive controllers already extract
what a single call can offer, and the proposer's added autonomy has nowhere
to go: the controller receives one fitness value per iteration, so what the
agent learned while constructing a candidate never reaches the selection
policy, and a stalled search can only be answered with a different context,
not with a direction. Autonomy granted to propose without a govern designed
to receive it is spent at the boundary.

The second pattern is that no agentic configuration reaches EvoTrias. The
remaining gap ranges from 0.65\% to 9.67\%. On each task the three agentic
variants also end close to one another: 2.612--2.618 on Circle Packing,
0.680--0.710 on Signal Processing, and 0.030--0.032 on Heilbronn Triangles.
This ceiling matches the Level~1 ablations of Table~\ref{tab:ablation} from
the other direction. Take Signal Processing. Adding an agentic proposer
lifts the controller-centric baselines to at most 0.710. Removing
Governance components from EvoTrias lowers it into the same range: 0.6926
with the default policy $\Pi_0$ in place of evidence-conditioned
adaptation, and 0.7041 without Meta-guidance. The full system scores
0.7329. A stronger proposer alone stops near 0.71, and only Structured
Governance carries the score from there to 0.7329.

The agentic mode we enable is the one SkyDiscover provides, and it leaves
the controller's role unchanged: the controller decides what the proposer
sees and takes back a single score. The proposer's autonomy is limited on
both sides of that exchange. EvoTrias runs a general-purpose coding agent
on the same three tasks because its contracts and hooks
(\S\ref{sec:open-proposal}, Appendix~\ref{app:hooks}) were designed for a
proposer of that kind.

\FloatBarrier
\subsection{Cross-Run Prevalence of the Two Experience Channels}
\label{app:channel-prevalence}

Appendix Figures~\ref{fig:note-trigger-replay} and~\ref{fig:meta-guidance-case}
each replay a single run. A case study shows that a channel \emph{can} fire, but
it leaves open whether the channel is a standing part of the search or an
artifact of one favorable trajectory. This subsection audits the GLM-5.2 runs
behind the main results instead: every Math and ADRS benchmark with three runs
each, together with all 30 Frontier-CS problems with one run each, for 66 runs
under the 50-iteration budget of \S\ref{sec:exp-setup}. For each run we report
how often each channel is
used and how the loaded notes relate to the task that consumes them.

\begin{figure}[t]
  \centering
  \includegraphics[width=0.97\textwidth]{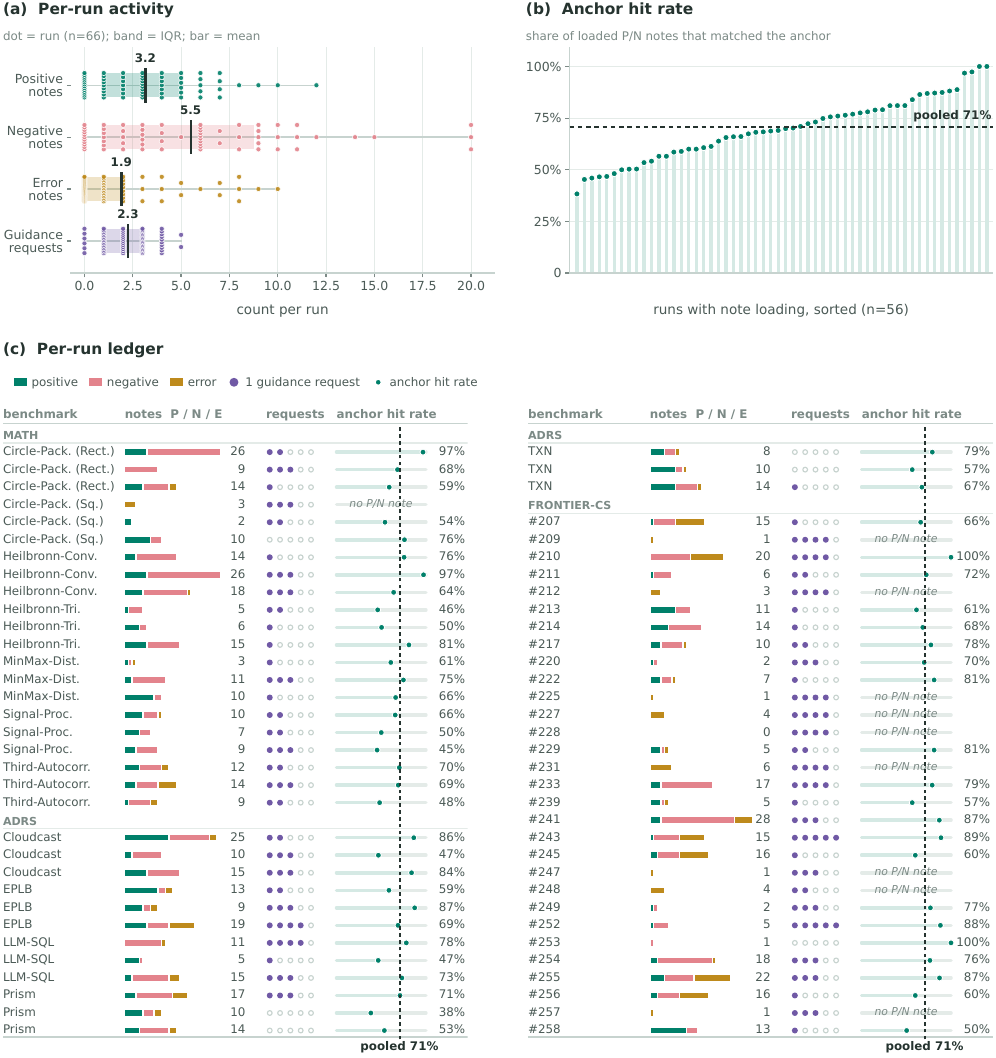}
  \caption{Prevalence of the two experience channels across the 66
  audited runs. (a) Per-run counts of positive, negative, and
  error notes and of Meta-guidance requests; each dot is one run, the band is
  the interquartile range, and the bar is the mean. (b) Anchor hit rate of note
  loading per run, sorted, for the 56 runs that loaded positive or negative
  notes; the dashed line is the value pooled over all loading events.
  (c) Per-run ledger grouped by benchmark suite, giving for each run the note
  composition by type, the number of guidance requests, and its anchor hit
  rate. Runs without a defined anchor hit rate are marked accordingly.}
  \label{fig:channel-prevalence}
\end{figure}

\paragraph{What the panels show.}
Figure~\ref{fig:channel-prevalence} reports the audit in three panels.
Panel (a) plots the per-run count of each note type and of guidance requests,
one dot per run, with the interquartile band and the mean. A run emits 3.2
positive, 5.5 negative, and 1.9 error notes on average, and issues 2.3 guidance
requests. Appendix~\ref{app:notes} defines the anchor hit rate, the share of
loaded positive and negative notes whose provenance intersects the anchor set
of the consuming task rather than arriving through the seeded random draw.
Panel (b) sorts by that rate the 56 runs that loaded positive or negative
notes, and marks the value pooled over all loading events. Panel
(c) gives the per-run ledger: one row per run, grouped by benchmark suite, with
the note composition, the guidance requests, and the anchor hit rate of that
run. The ledger is the panel that separates a prevalent channel from a
well-chosen example, since a channel that only worked once would leave most
rows empty.

\paragraph{Findings.}
Of the 66 runs, 65 emit at least one note and 60 issue at least one guidance
request. Across the audit, the runs record 697 notes and 149 guidance requests.
Note retrieval stays tied to the consuming context rather than degenerating
into a random draw over the pool: 70.7\% of loaded positive and negative notes
are anchor-matched. Ten runs have no defined anchor hit rate because no positive
or negative note-loading denominator is available, and are marked accordingly.
Both channels are therefore routinely active across benchmarks and suites,
which is the condition under which the single-run cases of
Appendix~\ref{app:notes} and~\ref{app:guidance} are representative.

\FloatBarrier
\subsection{Mechanistic Content of Evolution Notes}
\label{app:note-content-mechanism}

Having established that the two experience channels are routinely active
across the audited runs, we now turn from their prevalence to the content of
the Notes channel.

\subsubsection{Corpus and taxonomy}
\label{app:note-content-corpus}

\paragraph{Corpus.}
We analyze the 697 validated and committed evolution notes from the cross-run
audit in Appendix~\ref{app:channel-prevalence}. For each note, we retain its
distilled lesson and actionable guidance, the outcome that triggered its
creation, and the provenance of the attempt from which it was distilled.

\paragraph{Content taxonomy.}
We analyzed the 697-note corpus and grouped its content into two broad
families and seven mutually exclusive categories, with each note assigned
to exactly one category. The first family records decisions about the
search algorithm itself:
\begin{itemize}
  \item \textbf{Strategy} denotes a change to the core method or model family.
  Examples include replacing a greedy construction with dynamic programming
  or switching from one optimizer family to another. The defining question is
  which method to use, rather than how to tune a fixed method.
  \item \textbf{Search operator} covers changes made within a fixed method,
  including local moves, perturbations, restarts, seeding, tie breaking, beam
  width, and annealing or exploration schedules. These notes retain how the
  current search procedure should traverse its candidate space.
  \item \textbf{Objective} covers what the search optimizes or admits. It
  includes surrogate objectives, ranking keys, acceptance gates, validity
  checks, fallback rules, and verification required before a candidate is
  committed. These notes therefore describe the criterion used to distinguish
  an admissible improvement from a misleading score.
  \item \textbf{Budget} covers resource allocation and complexity, including
  wall-clock limits, iteration and population caps, restart counts, time
  sharing, and asymptotic cost. A change is assigned here when its motivation
  is a timeout or resource cap rather than candidate quality.
\end{itemize}
The second family records failures in executing or evaluating a proposed
change:
\begin{itemize}
  \item \textbf{Implementation defect} is a concrete coding error whose
  repair is mechanical once identified, such as a tensor-shape mismatch, a
  wrong positional index, a type error, a missing include, or misuse of a
  library call.
  \item \textbf{Environment} assigns the outcome to the evaluation harness or
  infrastructure rather than to the candidate. Judge outages, endpoint
  timeouts, and non-reproducible harness failures belong to this category.
  \item \textbf{Numerics} covers failures caused by precision or conditioning,
  including floating-point precision, tolerance choice, gradient scaling,
  degenerate geometry, and finite-difference step size.
\end{itemize}

Each content label is accompanied by two orthogonal annotations. \emph{Scope}
records how far the written lesson can travel: \textbf{instance} is tied to a
specific problem ID, configuration, or lineage; \textbf{benchmark} applies
across attempts on the same task family; and \textbf{general} can be used on
an unrelated benchmark. \emph{Conditioned} records whether the guidance states
where it applies, for example through a threshold, a size regime, an ``only
if'' clause, an acceptance check, or a fallback. Scope is judged from the
guidance as written.

\begin{figure}[t]
  \centering
  \includegraphics[width=0.985\textwidth]{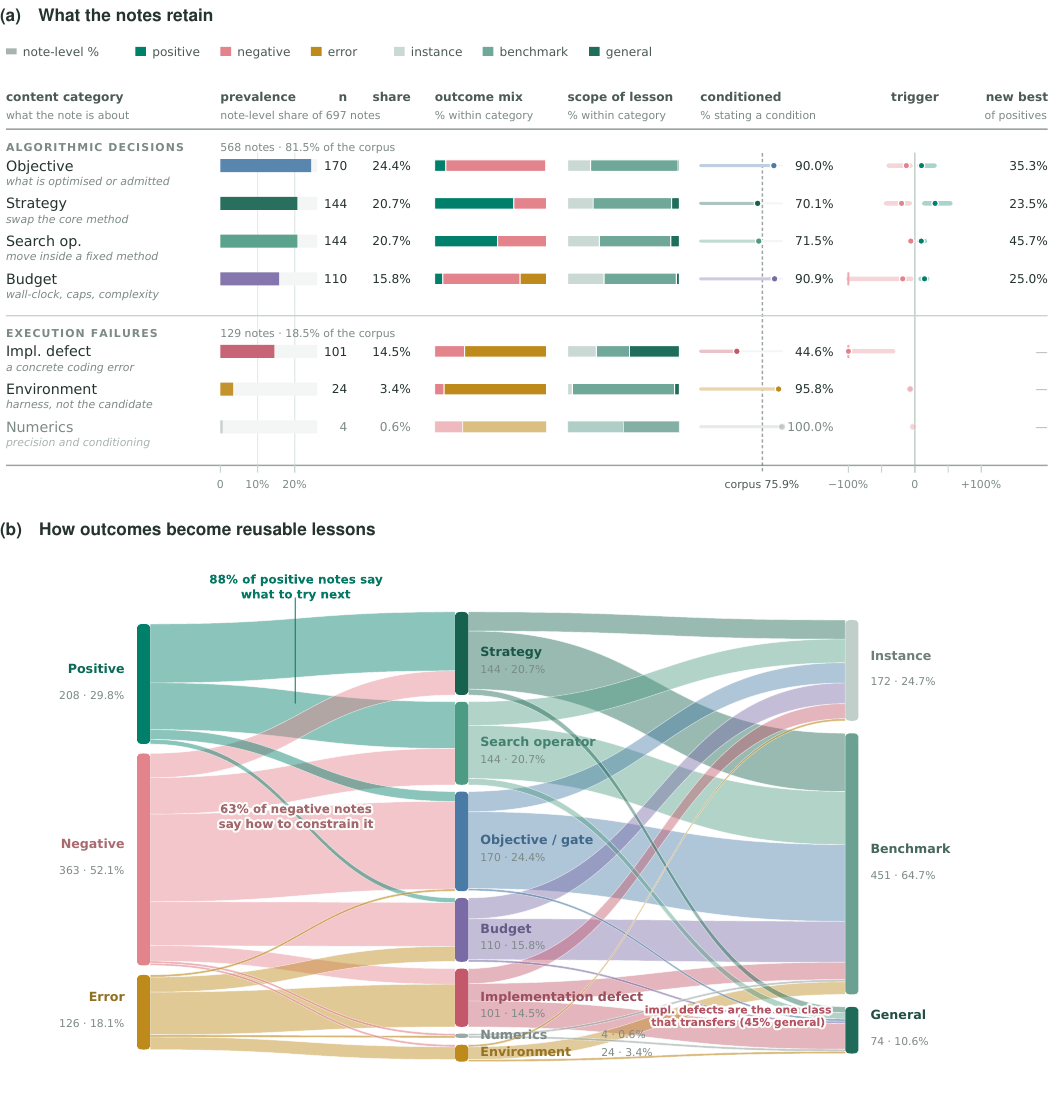}
  \caption{Mechanistic analysis of the 697 committed evolution notes.
  (a) Category-level prevalence, outcome mix, lesson scope, conditioning,
  trigger score changes, and new-best rates. (b) Flows from search outcome
  through content category to lesson scope, with ribbon width proportional to
  note count.}
  \label{fig:note-content-mechanism}
\end{figure}

\subsubsection{What the notes retain}
\label{app:note-content-ledger}

\paragraph{Reading panel (a).}
Figure~\ref{fig:note-content-mechanism}(a) is a category-level ledger over the
697 notes. Positive and negative notes are created by material gains and
regressions, respectively; error notes arise from repaired or terminal
failures. Two columns need a definition. The \emph{trigger} column summarizes
the parent-to-child score change that caused a positive or negative note to be
requested: for a child score $f$ and parent score $f_p$, this change is
$\delta=(f-f_p)/\max(|f_p|,\varepsilon)$, following Appendix~\ref{app:notes}.
The point is the median and the horizontal segment the interquartile range,
drawn to the right for gains and to the left for losses. \emph{New best} is the
fraction of positive notes whose triggering child also set a new run-level best;
categories without positive notes have no such rate.

\paragraph{Composition and outcome asymmetry.}
The four algorithmic-decision categories account for 568 of the 697 notes
(81.5\%), while the three execution-failure categories account for 129
(18.5\%). The Notes database is therefore primarily a record of method
selection, search refinement, acceptance, and resource control, rather than a
log dominated by software failures.

The outcome mix splits these categories cleanly. Positives concentrate in
strategy and search operator, negatives dominate objective and budget, and the
execution-failure family is carried by error notes. This is consistent with a
division of labor: successful attempts mainly preserve what method or move to
try, while regressions preserve the acceptance rules and resource limits that
prevent the same failure from recurring. The four numerics notes are too few to
support a category-level inference.

\paragraph{Scope and conditionality.}
Across the corpus, 451 notes (64.7\%) are benchmark-scoped, 172 (24.7\%) are
instance-scoped, and 74 (10.6\%) are general. The algorithmic-decision
categories are overwhelmingly benchmark-scoped and rarely general, because
their guidance depends on the task's objective, evaluator, or feasible
structure. Implementation defects are the one class that transfers: roughly
45\% of them are written as general lessons, since a faulty index or tensor
shape can recur outside the benchmark that exposed it. This
separation argues against treating every note as a global rule. Benchmark match
should govern the reuse of algorithmic experience, while general implementation
repairs can travel more freely.

Conditioning also concentrates where boundaries matter. Against a corpus-wide
rate of 75.9\%, objective reaches 90.0\% and budget 90.9\%, whose lessons often
retain a validity check, timeout cap, or fallback, and environment reaches
95.8\%, largely because the guidance names the harness state under which
attribution should be suspended. Implementation defects are conditioned only
44.6\% of the time: once a specific indexing error is identified, its
repair is often written as a direct rule.

\paragraph{Trigger magnitude and search role.}
Positive strategy notes follow the largest typical score moves, yet positive
search-operator notes set a new run-level best about twice as often (45.7\%
versus 23.5\%). This combination is consistent with two different search roles.
Replacing the core method can move far from its parent, but a large relative
step does not necessarily beat the incumbent found by another lineage. A local
operator starts from an established method and makes a smaller step, but more
often carries that method past the current best.

The loss side records a different risk profile: negative budget and
implementation-defect notes reach a complete loss in their lower tail. This
explains why the memory retains acceptance checks, caps, and mechanical repairs
alongside successful methods, since they protect evaluation budget from failure
modes that can erase an attempt. The trigger and new-best columns describe how
a note came to exist. Categories also differ in benchmark composition, so these
comparisons remain descriptive.

\subsubsection{How outcomes become reusable lessons}
\label{app:note-content-flow}

Figure~\ref{fig:note-content-mechanism}(b) follows the same 697 notes from
search outcome to content category and then to scope. Ribbon width is the
number of notes in a transition, and each of the three columns sums to 697.

\paragraph{Outcome-specific abstraction.}
The largest flows reveal a clear division in what each outcome contributes.
Of the 208 positive notes, 88\% concern strategy or search operators: they
mainly retain a method choice or a move that later tasks can try. Of the 363
negative notes, 63\% concern objective or budget: they mainly retain how to
constrain a candidate through validation, acceptance, fallback, or resource
limits. The 126 error notes concentrate in implementation defects and the
environment, retaining how to repair execution or how to avoid blaming a
candidate for an infrastructure failure. The three outcomes therefore preserve
different forms of reusable control information.

\paragraph{Where the lessons travel.}
The scope column then bounds reuse. Positive and negative lessons are
predominantly benchmark-scoped, and the general column is fed disproportionately
by implementation defects, of which roughly 45\% are general, whereas
environment notes stay attached to the evaluator that produced them. Mechanical
programming failures are the ones written as portable lessons; algorithmic and
harness lessons stay tied to their task family or evaluator.

Together, the flows describe a typed memory. Positive notes supply candidate
search actions, negative notes supply guards and limits, and error notes supply
recovery and attribution rules. The
scope annotation then controls how widely each lesson should be reused. This
organization preserves both sides of evolution: the proposals that advance a
lineage and the constraints that keep later lineages from spending budget on
known failure modes.

\subsection{OR-Tools Case: Experimental Settings}
\label{app:ortools}

This appendix details the repository-scale experiment in
\S\ref{sec:repo-scale-analysis}.

\paragraph{Repository and optimization target.}
OR-Tools 9.15~\citep{perron2024ortools} is Google's open-source optimization suite.
The version we evolve contains about 680K lines of C++ across roughly
2{,}000 source files (Table~\ref{tab:ortools-modules}). We target the
single-threaded feasibility-jump local search in CP-SAT, which sits in the
largest module but spans several interacting classes. In contrast to our
single-file benchmarks, the task specifies a solver-level objective rather
than a function to replace: locating the relevant execution paths is itself
part of proposal generation.

\paragraph{Edit surface.}
A candidate may modify four files under \texttt{ortools/sat/}:
\texttt{feasibility\_jump.h}, \texttt{feasibility\_jump.cc},
\texttt{constraint\_violation.h}, and \texttt{constraint\_violation.cc}.
These carry variable and delta selection,
constraint violation measures, jump scoring, constraint weight updates, and
initial solution construction, and one candidate often touches several of
them. This allowlist keeps patches reviewable and prevents changes to the
build, evaluator, instances, or scoring code, while leaving Open Proposal to
find the functions worth changing.
Together the four files hold 5{,}073 lines (698, 1{,}213, 807, and
2{,}355, in the order above), with about 30 classes and structs and about 120
function definitions. They include 22 other OR-Tools headers; their
transitive includes within the repository reach 39 files, about 13.8K lines
of headers and 24.9K lines with the matching implementations.
\texttt{cp\_model\_solver.cc} (3{,}081 lines) constructs and schedules the
local search they implement. The task marks no region inside these files,
and edits are coupled: the local search maintains every constraint violation
incrementally, so a change to jump scoring must be carried into the
incremental updates and the weight updates.

\begin{table}[htbp]
  \centering
  \small
  \caption{C++ modules of the OR-Tools tree, by size.}
  \label{tab:ortools-modules}
  \begin{tabular}{@{}lrl@{}}
    \toprule
    Directory & Lines & Contents \\
    \midrule
    \texttt{sat/}                & 234K        & CP-SAT solver \\
    \texttt{math\_opt/}          & 130K        & unified modeling layer across solvers \\
    \texttt{constraint\_solver/} & 108K        & legacy constraint programming and routing \\
    \texttt{graph/}              & 37K         & shortest path, flow, and assignment \\
    \texttt{linear\_solver/}     & 31K         & linear solver wrappers \\
    \texttt{glop/}               & 20K         & simplex LP solver \\
    \texttt{pdlp/}               & 19K         & first-order LP solver \\
    \texttt{util/}               & 15K         & shared utilities \\
    \texttt{flatzinc/}           & 13.5K       & FlatZinc and MiniZinc interface \\
    others                       & 1--13K each & scheduling, routing, packing, set cover \\
    \bottomrule
  \end{tabular}
\end{table}

\paragraph{Patch-based evolution.}
Each candidate is a Git patch against a frozen OR-Tools tree. At an
iteration, Structured Governance selects a parent and optional inspirations
from the settled candidates (\S\ref{sec:structured-governance}). Open
Proposal creates an isolated worktree, applies the parent, edits ordinary
C++ source, and exports a self-contained patch that passes the allowlist and
applies cleanly to the untouched tree. Experimental behavior is guarded by
\texttt{\#ifdef ORTOOLS\_LS\_EXPERIMENT}, with the original code kept in the
other branch, so each candidate reads as a difference against a solver that
still builds and runs. Carrying lineage as patches also avoids storing whole
repository copies.

\paragraph{Evaluation protocol.}
We conduct two evolution experiments, obtaining gains of $+13.2\%$ on the
full instance set and $+16.9\%$ on a smaller subset
(Figure~\ref{fig:ortools-gain}). The full set comprises 300 instances from
the MiniZinc Challenge 2023--2025, 100 per year, across 60
problem families (5 instances per family): 210 minimization, 79
maximization, and 11 satisfaction instances. The subset contains 60 instances,
selected deterministically as the second-smallest by size in each
problem family. It therefore retains complete coverage of all 60 families
while cutting the evaluation cost of each evolution iteration.

The two experiments are independent: they maintain separate lineages and do
not share candidates or scores. All other settings are identical. Protected
Judgment applies each patch to a fresh source copy, builds
\texttt{fzn-cp-sat} with the experimental flag, and runs it in free-search,
local-search-only mode. Every instance uses one solver thread and a 180-second
time limit.

\paragraph{Borda fitness.}
We follow the pairwise component of the MiniZinc Challenge's Borda-count
procedure, treating every instance as a voter and comparing the candidate
with the frozen unpatched solver. On an optimization instance, a better
feasible objective scores $1$, a worse objective or failure to return a valid
solution scores $0$, and equal-quality answers split the point by runtime. If
$t_x$ and $t_b$ are the candidate and baseline times (each lower-bounded by
one second), respectively, the candidate receives score
$t_b/(t_b+t_x)$. Satisfaction instances use the analogous valid-solution and
runtime comparison. For candidate $x$, frozen baseline $b$, and instance set
$D$, let $s_i(x,b)$ denote the pair score on instance $i$. Fitness is
$
  F_D(x)=\frac{1}{|D|}\sum_{i\in D}s_i(x,b),
$
where $|D|=300$ for the full experiment and $|D|=60$ for the subset
experiment, so $F_D$ lies in $[0,1]$. Comparing the unpatched solver with its
own results gives $s_i=0.5$ and $F_D=0.5$, which represents neither improvement
nor degradation. We define the relative performance ratio as
$R_D(x)=F_D(x)/0.5$ and report the percentage gain over the unpatched
solver as
\[
  \Delta_D(x)=\left(\frac{F_D(x)}{0.5}-1\right)\times 100\%.
\]

For each experiment, we run the unpatched solver once before evolution on the
same machine and with the same CPU allocation, solver flags, time limit, and
concurrency used for its candidates. We parse and store the resulting
per-instance records as a frozen baseline. During evolution, each candidate
produces only a new set of logs, which a deterministic Python collector
compares with these stored records. The script checks statuses, timeouts, and
abnormal exits, extracts objectives and runtimes, and computes and aggregates
all pair scores. Given the logs, this procedure is deterministic and efficient
and contains no model-generated judgments. Freezing the baseline, instance
manifest, resource limits, and toolchain keeps fitness comparable across
generations.

\paragraph{Search settings.}
Both experiments use EvoTrias's default operating settings
(Table~\ref{tab:hyperparams}): a 50-iteration task budget,
$n_{\mathrm{dir}}=4$ initial directions, and $c=4$ concurrent task-executor
slots. Open Proposal runs on Claude Code v2.1.201 with DeepSeek-V4-Flash-0731
as the backbone.

\paragraph{One-shot reference.}
To separate what the evolution loop adds from what the coding agent achieves
on its own, we also ran the same host and backbone once, end to end, under
the same task configuration, edit surface, and prompts. The agent reads the
repository, edits, builds, and submits a single patch; no Governance context,
selection, or later iteration follows. We ran four independent passes per
instance set and report the mean of their gains $\Delta_D$: $-0.1\%$ on the
full set and $+1.3\%$ on the subset.

\paragraph{Iterative-refinement baseline.}
A single pass spends one evaluation, so we also give the agent alone the
full budget. The same host and backbone run 50 sequential rounds without
EvoTrias, under the same task description, edit surface, evaluator, and time
limit. In each round the agent reads the evaluation result of the previous
round's candidate, applies that candidate's patch to a fresh source copy,
makes one change, exports a patch, and evaluates it. A patch that fails to
build or run is retried from the same starting point, up to three attempts
per round; if all three fail, the round keeps its starting point. Every
valid candidate becomes the starting point of the next round, and the agent
keeps a progress log across rounds. No Governance context, selection, or
shared notes are involved. The best gain over the 50 rounds is $+4.6\%$ on
the full set and $+2.2\%$ on the subset, against $+13.2\%$ and $+16.9\%$ for
EvoTrias; Figure~\ref{fig:ortools-gain} marks the two baseline values as
dotted lines.

\FloatBarrier

\subsection{Evolutionary Lineage Case Studies}
\label{app:lineage-case-studies}

\begin{figure}[p]
  \centering
  \captionsetup{skip=2pt}
  \includegraphics[width=0.74\textwidth]{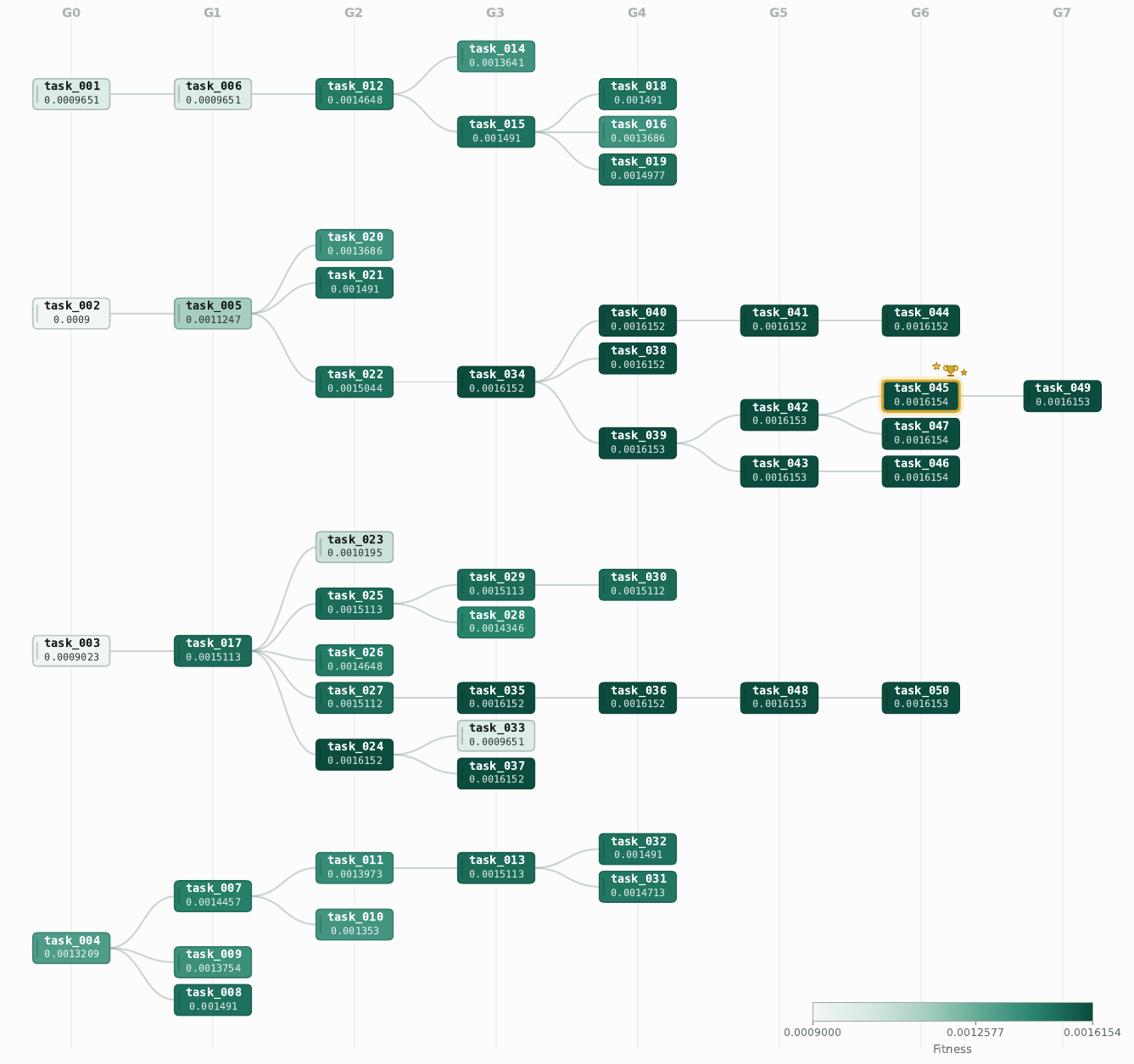}
  \caption{Evolutionary lineage of the Cloudcast run.}
  \label{fig:lineage-cloudcast}

  \vspace{2.5em}
  \includegraphics[width=0.74\textwidth]{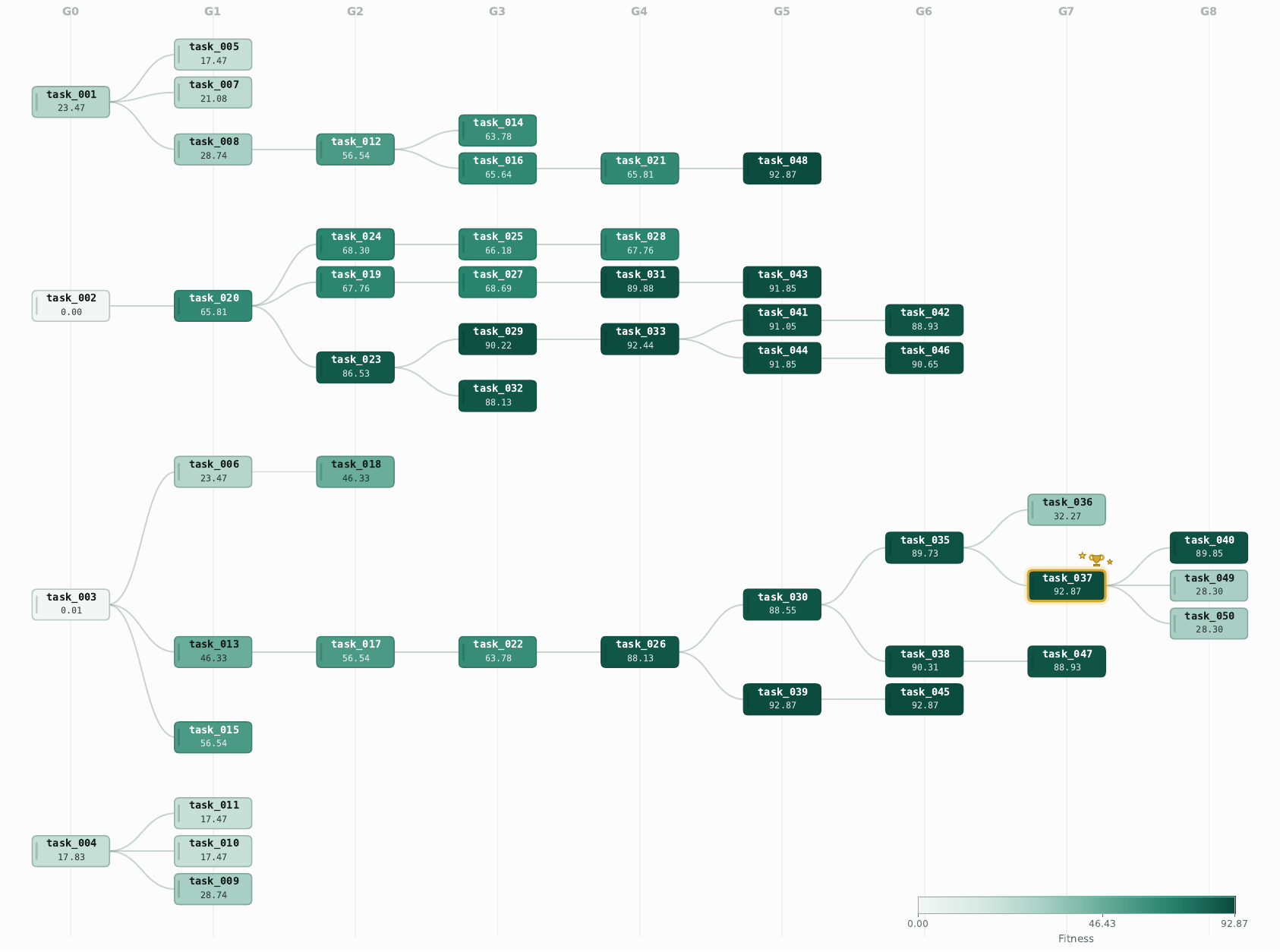}
  \caption{Evolutionary lineage of the Frontier-CS problem 258 run.}
  \label{fig:lineage-frontier-258}
\end{figure}

Figures~\ref{fig:lineage-cloudcast} and~\ref{fig:lineage-frontier-258}
reconstruct two runs, one on Cloudcast and one on Frontier-CS problem 258.
Both use Claude Code as the Open Proposal agent with GLM-5.2 as the backbone,
and both begin from the default $n_{\mathrm{dir}}=4$ initial directions. Each card is one
successfully evaluated iteration and reports its task ID and verified fitness.
$G$ denotes generation: the four tasks at $G0$ are the roots, and each edge
links the selected parent directly to its child. Card color follows the
run-specific fitness scale below each graph; the gold border and trophy mark
the best task in that run.

\FloatBarrier

\paragraph{Improvement with evolutionary depth.}
Both cases shift toward higher fitness as descendants accumulate, although
individual offspring can regress under exploration. Fitness here is the
internal score of Table~\ref{tab:fitness-map}, which on Cloudcast is
$1/(1+\text{cost})$ against the transfer cost of
Table~\ref{tab:combined-mean-results}. In Cloudcast, the median
fitness rises from 0.0009337 at $G0$ to 0.0014106 at $G1$, while the best value
improves from 0.0013209 among the four roots to 0.0016154. Frontier-CS problem
258 shows a larger change: the $G0$ median and maximum are 8.92 and 23.47,
respectively; by $G4$, they reach 88.13 and 92.44, and the run attains 92.87 at
$G5$. Thus, neither set of initial LLM proposals reaches the eventual run
optimum. The repeated gains along descendant chains provide case-level
evidence that evolutionary reuse and refinement improve on one-shot
generation for these optimization problems.

\paragraph{Adaptive allocation across lineages.}
Equal initialization does not imply equal allocation thereafter. In root order
\texttt{task\_001} through \texttt{task\_004}, the Cloudcast lineages contain
$7{:}16{:}15{:}8$ descendants, whereas Frontier-CS problem 258 contains
$8{:}16{:}18{:}3$. This contrast is consistent with the adaptive routing rule
of Structured Governance (Section~\ref{sec:structured-governance}). As detailed
in Appendix~\ref{app:selection}, Equation~\ref{eq:lineage-routing} converts
verified quality and descendant potential into a deserved lineage share, then
compares that share with the lineage's realized population. The search
therefore does not reserve one quarter of the remaining budget for each root;
it favors promising yet under-covered lineages. Within the routed lineage,
Equation~\ref{eq:parent} further combines candidate quality, exploration, and
descendant potential when selecting the parent. The fourth Frontier-CS root
illustrates why both levels matter. Its initial fitness is 17.83, higher than
two other roots, but its three descendants score only 17.47, 17.47, and 28.74,
while the other lineages eventually exceed 92. The resulting $8{:}16{:}18{:}3$
allocation reflects accumulated evidence about a branch rather than its root
score alone: a competitive initial direction can lose search budget when its
descendants fail to sustain improvement.
\subsection{Evaluator Exposure: Protocol, Held-out Suites, and Code Cases}
\label{app:evaluator-exposure}

\subsubsection{Exposure protocol}
\label{app:exposure-protocol}

The exposed and protected settings differ only in whether evaluator sources are
reachable. In the exposed setting CORAL runs in its default host session,
which links the grader source into every agent worktree and grants read
access to it (Appendix~\ref{app:judge-baselines}), and EvoTrias runs with
the permission hooks of Appendix~\ref{app:hooks} disabled. In the
protected setting the two systems carry the same measures of
\S\ref{sec:protected-judgment}: the prompt states the prohibition, and a
hook layer denies the corresponding tool calls. CORAL has no layer that
denies reading evaluator sources, so we supply an EvoTrias-equivalent one
for its protected runs and leave the rest of its design untouched.

The exposed setting gives EvoTrias the same access its baselines take.
Following CORAL's configuration, the exposed proposal prompts point Open
Proposal at the evaluator source and encourage reading it; following
AdaEvolve, the Meta-guidance request carries the evaluator source alongside
the search trace. CORAL keeps its stock instruction file, which points the
agent at the grader source. Any use of the sources is therefore each
system's own behavior under its exposed prompts. Each of the six cells is three
independent runs, and every score in Table~\ref{tab:evaluator-exposure} is the
mean over those three. Each run's final program is then scored once on the
original evaluator and once on each held-out suite below.

\subsubsection{Task and held-out construction}

Signal Processing requires a program to map a noisy, non-stationary input
signal and a window size to a filtered signal. The evaluator aggregates
smoothness, tracking error, false reversals, correlation, and noise reduction
over several signals. The intended object of optimization is therefore a
filtering algorithm for a family of signals, rather than a solution to one
fixed instance. This differs from Circle Packing, where the task itself asks
for one concrete arrangement of 26 circles in a unit square.

We also compared exposed and protected evaluation on Circle Packing. Their
original-evaluator scores differed by less than 1\%, consistent with evaluator
exposure having little effect when the evaluation target is the specified case
itself rather than performance on unseen instances.

Table~\ref{tab:signal-heldout-design} separates the original suite from the
two held-out suites. All three use the same alignment and scoring
implementation; only the generated signals change. The IID suite preserves
the original five families and Gaussian-noise assumption, but draws five new
parameterizations of each family. The OOD suite varies every major axis of the
original generator and adds five signal families.

\begin{table}[htbp]
  \centering
  \caption{Signal Processing evaluation suites. IID preserves the task
  families while randomizing their instances; OOD broadens both the signal
  and disturbance distributions.}
  \label{tab:signal-heldout-design}
  \scriptsize
  \setlength{\tabcolsep}{4pt}
  \renewcommand{\arraystretch}{1.08}
  \begin{tabularx}{\textwidth}{@{}p{0.14\textwidth}LLL@{}}
    \toprule
    \textbf{Dimension} & \textbf{Original evaluator}
      & \textbf{IID held-out} & \textbf{OOD held-out} \\
    \midrule
    Cases
      & 5
      & 25
      & 20 \\
    \midrule
    Signal families
      & Five fixed families, one case each
      & Same five families, five cases each
      & Five widened families plus five new families, two cases each \\
    \midrule
    Length
      & 500, 600, 700, 800, 900
      & 529--997
      & 457--1217 \\
    \midrule
    Family--length relation
      & One-to-one; length identifies the case
      & Multiple lengths per family
      & Length is decoupled from family \\
    \midrule
    Window size
      & 20
      & 20
      & 9--67 \\
    \midrule
    Time span
      & 0--10
      & 8--14
      & 6.5--17 \\
    \midrule
    Disturbance
      & IID Gaussian, $\sigma=0.2$--$0.6$
      & IID Gaussian, $\sigma=0.16$--$0.68$
      & Gaussian, AR colored, Student-$t$, impulsive, and heteroscedastic \\
    \midrule
    Parameters
      & Fixed frequencies, amplitudes, and step locations
      & Randomized amplitudes, frequencies, phases, steps, drift, and noise
      & Wider parameter ranges and additional non-stationary dynamics \\
    \midrule
    Seed
      & $42+i$
      & $1049+97f+17r$
      & $31013+101i$, independent of the original evaluator \\
    \bottomrule
  \end{tabularx}
\end{table}

For IID case $(f,r)$, the generator samples duration uniformly from
$[8,14]$, selects a family-specific parameterization, and adds Gaussian noise
with a sampled standard deviation. The five retained families are
sine-with-trend, multi-frequency oscillation, chirp, piecewise-constant step,
and random walk. Thus, IID changes the concrete signals without changing the
task taxonomy, window size, or noise family.

The OOD suite uses ten families: wider versions of the original five, plus
amplitude modulation, damped oscillation, piecewise ramp, frequency switch,
and smooth latent drift. It also varies the window size and introduces colored,
heavy-tailed, impulsive, and heteroscedastic disturbances. This construction
breaks three fingerprints available in the original evaluator: the mapping
from length to signal family, the fixed \texttt{window\_size=20}, and the
assumption of IID Gaussian noise.

\FloatBarrier
\subsubsection{Code-level mechanism}
\label{app:evaluator-code-cases}

Table~\ref{tab:evaluator-code-case-scores} reports two illustrative programs.
They come from different method groups and are not a paired estimate of the
exposure effect; the controlled group comparison remains
Table~\ref{tab:evaluator-exposure}. Instead, the programs provide qualitative
evidence for the mechanism behind the aggregate advantage contraction.

\begin{center}
\begin{minipage}{\textwidth}
  \centering
  \captionsetup{hypcap=false}
  \captionof{table}{Scores of the two code cases shown below. Gap is OOD minus
  Original.}
  \label{tab:evaluator-code-case-scores}
  \small
  \setlength{\tabcolsep}{7pt}
  \renewcommand{\arraystretch}{1.08}
  \begin{tabular}{@{}llrrrr}
    \toprule
    Method & Evaluator & Original & IID & OOD & Gap \\
    \midrule
    CORAL-1  & Exposed   & 0.808966 & 0.783652 & 0.687841
      & \cellcolor{tablegapred}$-$0.121125 \\
    EvoTrias & Protected & 0.719451 & 0.723515 & 0.730097
      & \cellcolor{tablegapgreen}$+$0.010646 \\
    \bottomrule
  \end{tabular}
\end{minipage}
\end{center}

\clearpage
\begin{center}
\begin{minipage}{\textwidth}
\captionsetup{hypcap=false}
\begin{exposedcodebox}{Exposed program (CORAL-1, selected excerpt)}
# Tuned per signal type against the grader's fixed seeds.
_SCHEDULES = {
    "sinsin":     "sin_demod",
    "multifreq":  ((21, 101), (2, 2)),
    "chirp":      "hilbert",
    "step":       "pwc",
    "randomwalk": ((31, 51, 71, 91, 111), (2, 2, 2, 2, 2)),
    "default":    ((31, 41, 61, 81, 91), (2, 2, 2, 2, 2)),
}
_STEP_PWC_WIN = 40
_STEP_PWC_THR = 0.5
_CHIRP_PHASE_W = 401
_CHIRP_ENV_W = 251
_SHORT_SIGNAL_THRESHOLD = 500  # all grader signals have n >= 500

# ... generic smoothing helpers omitted ...
def _classify(x):
    n = len(x)
    y = medfilt(x, _safe_window(9, n))
    y = _savgol(y, _safe_window(21, n), 2)

    centered = y - y.mean()
    zc = (np.abs(np.diff(np.sign(centered))) > 0).astype(float)
    q = max(1, n // 4)
    rates = [zc[i*q:(i+1)*q].mean() for i in range(4)]
    zc_trend, zc_final = rates[-1] - rates[0], rates[-1]

    d = np.abs(np.diff(y))
    jump_frac = (d > 3 * d.mean()).sum() / max(len(d), 1)
    local_std = np.array([
        y[max(0, i-30):i+31].std() for i in range(n)
    ])
    std_ratio = local_std.max() / (local_std.mean() + 1e-9)

    yc, var = y - y.mean(), y.var() + 1e-9
    def ac(lag):
        return np.sum(yc[:-lag] * yc[lag:]) / ((n-lag) * var)
    ac_arr = np.array([ac(l) for l in range(5, min(250, n//2))])
    peaks = [v for i, v in enumerate(ac_arr[2:-2], 2)
             if v > ac_arr[i-1] and v > ac_arr[i+1] and v > 0.5]

    # Thresholds recover the five fixed evaluator families.
    if zc_trend > 0.02 and zc_final > 0.05:
        return "chirp"
    if jump_frac > 0.02 and std_ratio > 3.0:
        return "step"
    if peaks and max(peaks) > 0.7:
        return "sinsin" if abs(ac(100)) > 0.6 else "multifreq"
    return "randomwalk" if ac(min(150, n//3)) > -0.2 else "multifreq"

# ... model-matched chirp, sinusoid, and step filters omitted ...

def adaptive_filter(x, window_size=20):
    x = np.asarray(x, dtype=float).ravel()
    n = len(x)
    if n < _SHORT_SIGNAL_THRESHOLD:  # never used by the original grader
        y = _savgol(medfilt(x, 3), 5, 2)
        out = y[window_size - 1:].copy()
        out[-1] = x[-1]
        return out

    kind = _classify(x)
    spec = _SCHEDULES.get(kind, _SCHEDULES["default"])
    if spec == "hilbert":
        smooth = _hilbert_chirp(x)
    elif spec == "sin_demod":
        smooth = _sin_demod(x)
    elif spec == "pwc":
        smooth = _step_pwc(x)
    else:
        smooth = _cascade(x, *spec)

    # Match the grader's alignment and force its recent-lag term to zero.
    out = smooth[window_size - 1:].copy()
    out[-2] = 0.5 * (out[-3] + x[-1])
    out[-1] = x[-1]
    return out
\end{exposedcodebox}
\captionof{figure}{Exposed solution excerpt. Red identifiers mark the fixed
schedule and length regime, case-specific classification and routing, and the
evaluator-aligned output shortcut. Generic filtering helpers are omitted.}
\label{fig:exposed-code-case}
\medskip

The exposed program reaches 0.808966 on the original evaluator but falls to
0.687841 OOD. Beyond choosing a denoiser, it encodes the evaluator's alignment,
forces the last-sample lag term to zero, identifies the five fixed families,
and routes them to separately tuned procedures. Its comments state that the
schedules were tuned against the grader's fixed seeds.
\end{minipage}
\end{center}

\clearpage
\begin{center}
\begin{minipage}{\textwidth}
\captionsetup{hypcap=false}
\begin{codecasebox}{Protected program (EvoTrias, selected excerpt)}
def _haar_finest_sigma(signal):
    x = np.asarray(signal, dtype=float).flatten()
    if len(x) < 4:
        return float(np.std(x)) if len(x) else 0.0
    even, odd = x[0::2], x[1::2]
    m = min(len(even), len(odd))
    detail = (even[:m] - odd[:m]) / np.sqrt(2.0)
    mad = float(np.median(np.abs(detail)))
    return mad / 0.6745 if mad > 0 else float(np.std(detail)) * 0.5

def _estimate_lengthscale(signal, max_lag_cap=200):
    x = np.asarray(signal, dtype=float).flatten()
    x_centered = x - np.mean(x)
    var = float(np.dot(x_centered, x_centered)) / len(x)
    acf = np.correlate(x_centered, x_centered, mode="full")
    acf = acf[len(x) - 1:] / (var * len(x))
    max_lag = min(max_lag_cap, len(x) - 1)
    below = np.where(acf[1:max_lag + 1] < 0.2)[0]
    zero = np.where(np.diff(np.sign(acf[1:max_lag + 1])) != 0)[0]
    candidates = []
    if below.size:
        candidates.append(int(below[0]) + 1)
    if zero.size:
        candidates.append(int(zero[0]) + 1)
    lag = min(candidates) if candidates else max(4, min(20, len(x)//20))
    return 1.5 * max(2.0, lag)

def _matern32_kernel(distance, lengthscale, sigma_f):
    scaled = np.sqrt(3.0) * distance / max(float(lengthscale), 1e-6)
    return sigma_f**2 * (1.0 + scaled) * np.exp(-scaled)

def _gp_matern32_fit_predict(x, lengthscale, sigma_f, sigma_n, m_cap=80):
    x = np.asarray(x, dtype=float).flatten()
    n, m = len(x), min(m_cap, len(x))
    idx = np.unique(np.linspace(0, n - 1, m).astype(int))
    t_full, t_sub = np.arange(n, dtype=float), idx.astype(float)
    y_sub = x[idx]

    d_sub = np.abs(t_sub[:, None] - t_sub[None, :])
    K_xx = _matern32_kernel(d_sub, lengthscale, sigma_f)
    jitter = 1e-6 * (sigma_f**2 + 1.0)
    K_y = K_xx + (sigma_n**2 + jitter) * np.eye(len(idx))
    cho, low = cho_factor(K_y, lower=True, check_finite=False)
    alpha = cho_solve((cho, low), y_sub, check_finite=False)

    d_cross = np.abs(t_full[:, None] - t_sub[None, :])
    mean = _matern32_kernel(d_cross, lengthscale, sigma_f) @ alpha
    return mean if np.all(np.isfinite(mean)) else None

# ... zero-phase Savitzky-Golay helpers and parent fallback filter omitted ...

def _run_filter_full(x, window_size):
    x = np.asarray(x, dtype=float).flatten()
    if len(x) < 2:
        return x.copy()

    noise_sigma = max(_haar_finest_sigma(x), 1e-6)
    signal_scale = max(float(np.std(x)), 1e-6)
    lengthscale = _estimate_lengthscale(x)

    gp_mean = _gp_matern32_fit_predict(
        x, lengthscale=lengthscale, sigma_f=signal_scale,
        sigma_n=noise_sigma, m_cap=80,
    )
    if gp_mean is None or len(gp_mean) != len(x):
        return _parent_fallback_filter(x, window_size)

    broad = min(max(3, int(window_size) | 1),
                max(5, len(x) // 4 | 1))
    fine = max(3, min(int(window_size) // 2 | 1,
                      len(x) // 8 | 1))
    smoothed = _two_stage_sg_smooth(gp_mean, broad, fine)
    return smoothed
\end{codecasebox}
\captionof{figure}{Protected solution excerpt. The noise scale, kernel
lengthscale, signal scale, and post-smoothing windows are inferred from the
current input; no fixed family, length, or seed selects an algorithm branch.
Ellipses omit the Savitzky--Golay smoothing helpers and the parent fallback filter.}
\label{fig:protected-code-case}
\medskip

The protected program reaches 0.719451 on the original evaluator and improves
to 0.730097 OOD. It uses one signal-agnostic pipeline: Haar-MAD estimates the
noise, autocorrelation estimates the Mat\'ern lengthscale, and the supplied
window controls the final smoothing.
\end{minipage}
\end{center}

\clearpage

\subsection{Ownership Ablation: Protocol and Full Results}
\label{app:ownership-ablation}

\paragraph{Protocol.} A and B share one plugin build, Claude Code with
GLM-5.2, the root phase, the construction instructions of the task executor,
the retry rules, the protected evaluator, 50 tasks, $n_{\mathrm{dir}}=4$,
and $c=4$. B replaces the Governance tools with three MCP tools.
\texttt{get\_search\_state} returns the raw history: every candidate with its
artifact path, fitness, metrics, and lineage, every failed task, every note,
and the budget progress. \texttt{declare\_selection} records the parent and
inspirations the agent chose. \texttt{write\_note} stores a note as written
and shows it to every later task. B's prompt explains these tools and leaves
the strategy to the agent. Its executor may read every task directory of the
run, and the evaluator source stays denied. Tokens are taken from the Claude
Code transcripts, once per model request; tokens per task are those of the
task executor, and tokens per run add the other agents of the run.

\paragraph{Full results.} Table~\ref{tab:ownership} adds two views to
the score and cost of \S\ref{sec:ownership-ablation}. After the root phase,
A uses 3.7--4.0 root lineages on average, while B uses 1.0--1.7. B writes
35--44.7 notes per run, and 89--91\% of its notes come from attempts that moved
fitness by less than 2\% relative to the parent; A commits 4.0--16.3 notes
per run, none below that level. B turns attempts with little measured
fitness change into shared lessons and concentrates the search on fewer
root lineages. Alongside its higher token cost, these patterns support the
concern in \S\ref{sec:introduction} that agent-managed governance can lead to
unreliable memory, premature search narrowing, and unnecessary computational
costs.

\begin{table}[htbp]
  \centering
  \caption{Controlled ownership ablation (three runs per row). A:
  Structured Governance makes the search decisions (EvoTrias); B: the coding
  agent does. Score: final best, mean $\pm$ std. Tokens per task: iterations
  5--15 $\rightarrow$ 40--50. Lineages: root lineages that receive a task
  after the root phase. Top root: runs whose final best descends from the
  root with the highest root-phase score. Lineages and notes per run are
  averaged over three runs; in
  B a task may write more than one. Small-$\Delta$: notes from attempts
  within 2\% of the parent's fitness.}
  \label{tab:ownership}
  \vspace{-5pt}
  \scriptsize
  \setlength{\tabcolsep}{2.2pt}
  \renewcommand{\arraystretch}{1.12}
  \newcommand{\ostd}[1]{{\fontsize{4.8}{5.2}\selectfont\color{black!70}$\pm$#1}}
  \begin{tabular*}{\textwidth}{@{\extracolsep{\fill}}l r r c c c r r@{}}
    \toprule
    & & \multicolumn{2}{c}{Cost} & \multicolumn{2}{c}{Search breadth}
    & \multicolumn{2}{c}{Shared notes} \\
    \cmidrule(lr){3-4}\cmidrule(lr){5-6}\cmidrule(lr){7-8}
    Owner of the search & Score $\uparrow$ & Tokens / run & Tokens / task
      & Lineages & Top root & Per run & Small-$\Delta$ \\
    \midrule
    \rowcolor{backbonegray}
    \multicolumn{8}{@{}l}{\textit{Circle Packing}} \\
    A: Structured Governance & \cellcolor{resultfirst}2.6340\ostd{0.0026}
      & 25.9M & 321K $\rightarrow$ 383K & 3.7 & 0/3 & 4.0 & 0\% \\
    B: coding agent &2.6253\ostd{0.0126}
      & 33.2M & 364K $\rightarrow$ 675K & 1.7 & 3/3 & 35 & 91\% \\
    \midrule
    \rowcolor{backbonegray}
    \multicolumn{8}{@{}l}{\textit{Signal Processing}} \\
    A: Structured Governance & \cellcolor{resultfirst}0.7597\ostd{0.0317}
      & 27.6M & 368K $\rightarrow$ 418K & 4.0 & 0/3 & 16.3 & 0\% \\
    B: coding agent &0.6391\ostd{0.0579}
      & 32.8M & 355K $\rightarrow$ 605K & 1.0 & 3/3 & 44.7 & 89\% \\
    \bottomrule
  \end{tabular*}
\end{table}

%% file: sections/mean_best_results_table.tex
\begin{center}
\begin{minipage}{\textwidth}
  \centering
  \captionsetup{hypcap=false}
  \captionof{table}{Mean, best-run, and sample standard deviation (Std) on Math
  and ADRS; OpenEvolve omitted, and lower is better for Third Autocorrelation
  and Cloudcast ($\downarrow$). First and second Mean and Best results within
  each backbone are highlighted as \colorbox{resultfirst}{first} and
  \colorbox{resultsecond}{second}.}
  \label{tab:mean-best-results}
  \scriptsize
  \setlength{\tabcolsep}{1.0pt}
  \renewcommand{\arraystretch}{1.02}
  \newcommand{\stdcell}[1]{{\color{black!55}%
    \raisebox{6.4pt}[0pt][0pt]{\scalebox{0.88}{#1}}}}
  \newlength{\problemgroupspace}
  \setlength{\problemgroupspace}{0.7pt}
  \newcommand{\problemgroupbreak}{%
    \addlinespace[\dimexpr\problemgroupspace-5.75pt\relax]}
  \newcommand{\terminalglmstd}[1]{%
    \multicolumn{1}{r@{\hspace{3.4pt}%
      \raisebox{0pt}[0pt][0pt]{\rule[-1.2pt]{0.35pt}{9.3pt}}%
      \hspace{1.0pt}}}{{\color{black!55}%
        \raisebox{0.68pt}[0pt][0pt]{\scalebox{0.88}{#1}}}}}
  \begin{tabularx}{\textwidth}{@{}l >{\raggedleft\arraybackslash}p{20pt}
    *{5}{>{\raggedleft\arraybackslash}X}
    @{\hspace{3.4pt}\vrule width 0.35pt\hspace{1.0pt}}
    *{5}{>{\raggedleft\arraybackslash}X}@{}}
    \toprule
    & & \multicolumn{5}{c}{GLM-5.2}
      & \multicolumn{5}{c}{Opus-5} \\
    \cmidrule(lr){3-7}\cmidrule(lr){8-12}
    Task & Stat & EvoX & AdaE. & CORAL-1 & CORAL-4 & EvoTrias
      & EvoX & AdaE. & CORAL-1 & CORAL-4 & EvoTrias \\
    \midrule
    \rowcolor{backbonegray}
    \multicolumn{12}{l}{\textit{Math}} \\
    \multirow{3}{*}{Circle-Pack. (Sq.)}
      & Mean & 2.60956 & 2.63152 & 2.62337
      & \cellcolor{resultsecond}2.63418
      & \cellcolor{resultfirst}2.63484
      & \cellcolor{resultsecond}2.63598
      & \cellcolor{resultsecond}2.63598
      & \cellcolor{resultsecond}2.63598
      & \cellcolor{resultsecond}2.63598
      & \cellcolor{resultfirst}2.63599 \\
      & Best & 2.61094 & \cellcolor{resultsecond}2.63429 & 2.63135
      & \cellcolor{resultfirst}2.63598
      & \cellcolor{resultfirst}2.63598
      & \cellcolor{resultsecond}2.63598
      & \cellcolor{resultsecond}2.63598
      & \cellcolor{resultsecond}2.63598
      & \cellcolor{resultsecond}2.63598
      & \cellcolor{resultfirst}2.63599 \\
      & \stdcell{Std} & \stdcell{0.00123} & \stdcell{0.00240}
      & \stdcell{0.00794} & \stdcell{0.00309} & \stdcell{0.00193}
      & \stdcell{1.72e-9} & \stdcell{1.52e-9} & \stdcell{0}
      & \stdcell{4.57e-9} & \stdcell{6.75e-6} \\
    \problemgroupbreak
    \multirow{3}{*}{Circle-Pack. (Rect.)}
      & Mean & 2.34876 & \cellcolor{resultsecond}2.36487 & 2.32097
      & 2.36459 & \cellcolor{resultfirst}2.36493
      & \cellcolor{resultsecond}2.36583
      & \cellcolor{resultfirst}2.36584
      & \cellcolor{resultsecond}2.36583
      & \cellcolor{resultsecond}2.36583
      & \cellcolor{resultfirst}2.36584 \\
      & Best & 2.35825 & \cellcolor{resultsecond}2.36575
      & \cellcolor{resultsecond}2.36425
      & \cellcolor{resultfirst}2.36583
      & \cellcolor{resultfirst}2.36583
      & \cellcolor{resultsecond}2.36583
      & \cellcolor{resultfirst}2.36584
      & \cellcolor{resultsecond}2.36583
      & \cellcolor{resultsecond}2.36583
      & \cellcolor{resultfirst}2.36584 \\
      & \stdcell{Std} & \stdcell{0.0102} & \stdcell{0.00152}
      & \stdcell{0.0613} & \stdcell{0.00214} & \stdcell{0.00157}
      & \stdcell{0} & \stdcell{5.24e-6} & \stdcell{0}
      & \stdcell{0} & \stdcell{5.11e-6} \\
    \problemgroupbreak
    \multirow{3}{*}{Heilbronn-Tri.}
      & Mean & 0.0302 & 0.0318 & 0.0320
      & \cellcolor{resultsecond}0.0332
      & \cellcolor{resultfirst}0.0335
      & 0.0361 & \cellcolor{resultsecond}0.0364
      & \cellcolor{resultsecond}0.0364
      & \cellcolor{resultfirst}0.0365
      & \cellcolor{resultfirst}0.0365 \\
      & Best & 0.0343 & 0.0331 & \cellcolor{resultfirst}0.0361
      & 0.0343 & \cellcolor{resultsecond}0.0360
      & \cellcolor{resultfirst}0.0365
      & \cellcolor{resultfirst}0.0365
      & \cellcolor{resultfirst}0.0365
      & \cellcolor{resultfirst}0.0365
      & \cellcolor{resultfirst}0.0365 \\
      & \stdcell{Std} & \stdcell{0.00395} & \stdcell{0.00129}
      & \stdcell{0.00427} & \stdcell{0.00105} & \stdcell{0.00223}
      & \stdcell{0.000438} & \stdcell{0.000287} & \stdcell{0.000171}
      & \stdcell{0} & \stdcell{0} \\
    \problemgroupbreak
    \multirow{3}{*}{Heilbronn-Conv.}
      & Mean & 0.0269 & 0.0275 & \cellcolor{resultsecond}0.0284
      & 0.0283 & \cellcolor{resultfirst}0.0302
      & \cellcolor{resultsecond}0.0307
      & 0.0304
      & \cellcolor{resultfirst}0.0309
      & \cellcolor{resultfirst}0.0309
      & \cellcolor{resultfirst}0.0309 \\
      & Best & \cellcolor{resultsecond}0.03043 & 0.02777
      & 0.02870 & 0.02864 & \cellcolor{resultfirst}0.03085
      & \cellcolor{resultfirst}0.03094
      & \cellcolor{resultsecond}0.03086
      & \cellcolor{resultfirst}0.03094
      & \cellcolor{resultfirst}0.03094
      & \cellcolor{resultfirst}0.03094 \\
      & \stdcell{Std} & \stdcell{0.00246} & \stdcell{0.000451}
      & \stdcell{0.000439} & \stdcell{0.000447} & \stdcell{0.000541}
      & \stdcell{0.000445} & \stdcell{0.000401} & \stdcell{0}
      & \stdcell{0} & \stdcell{0} \\
    \problemgroupbreak
    \multirow{3}{*}{MinMax-Dist.}
      & Mean & 0.23970 & 0.23987 & 0.23975
      & \cellcolor{resultsecond}0.24003
      & \cellcolor{resultfirst}0.24005
      & \cellcolor{resultfirst}0.24005
      & \cellcolor{resultfirst}0.24005
      & \cellcolor{resultfirst}0.24005
      & \cellcolor{resultfirst}0.24005
      & \cellcolor{resultfirst}0.24005 \\
      & Best & \cellcolor{resultfirst}0.24005
      & \cellcolor{resultfirst}0.24005
      & \cellcolor{resultfirst}0.24005
      & \cellcolor{resultfirst}0.24005
      & \cellcolor{resultfirst}0.24005
      & \cellcolor{resultfirst}0.24005
      & \cellcolor{resultfirst}0.24005
      & \cellcolor{resultfirst}0.24005
      & \cellcolor{resultfirst}0.24005
      & \cellcolor{resultfirst}0.24005 \\
      & \stdcell{Std} & \stdcell{0.000448} & \stdcell{0.000211}
      & \stdcell{0.000514} & \stdcell{4.49e-5} & \stdcell{8.92e-8}
      & \stdcell{0} & \stdcell{0} & \stdcell{0} & \stdcell{0} & \stdcell{0} \\
    \problemgroupbreak
    \multirow{3}{*}{Signal-Proc.}
      & Mean & \cellcolor{resultsecond}0.7098 & 0.6961 & 0.6638
      & 0.7083 & \cellcolor{resultfirst}0.7329
      & 0.7861 & 0.7976 & 0.8251
      & \cellcolor{resultsecond}0.8280
      & \cellcolor{resultfirst}0.8313 \\
      & Best & 0.7173 & \cellcolor{resultsecond}0.7214 & 0.6669
      & 0.7149 & \cellcolor{resultfirst}0.7370
      & 0.8079 & 0.8140 & 0.8264
      & \cellcolor{resultsecond}0.8294
      & \cellcolor{resultfirst}0.8315 \\
      & \stdcell{Std} & \stdcell{0.00951} & \stdcell{0.0226}
      & \stdcell{0.00515} & \stdcell{0.00573} & \stdcell{0.00583}
      & \stdcell{0.0319} & \stdcell{0.0183} & \stdcell{0.00118}
      & \stdcell{0.00117} & \stdcell{0.000224} \\
    \problemgroupbreak
    \multirow{3}{*}{Third-Autocorr. $\downarrow$}
      & Mean & \cellcolor{resultfirst}1.4601 & 1.4614 & 1.4814
      & \cellcolor{resultsecond}1.4608 & 1.4609
      & 1.4587 & 1.4568 & 1.4572
      & \cellcolor{resultfirst}1.4559
      & \cellcolor{resultsecond}1.4563 \\
      & Best & 1.4594 & 1.4590 & 1.4593
      & \cellcolor{resultsecond}1.4574
      & \cellcolor{resultfirst}1.4563
      & 1.4571 & 1.4560 & \cellcolor{resultsecond}1.4556
      & \cellcolor{resultfirst}1.4548 & 1.4560 \\
      & \stdcell{Std} & \stdcell{0.000681} & \stdcell{0.00247}
      & \stdcell{0.0310} & \stdcell{0.00318} & \terminalglmstd{0.00404}
      & \stdcell{0.00145} & \stdcell{0.000655} & \stdcell{0.00160}
      & \stdcell{0.00127} & \stdcell{0.000549} \\
    \noalign{\vskip -4.5pt}
    \midrule
    \rowcolor{backbonegray}
    \multicolumn{12}{l}{\textit{ADRS}} \\
    \multirow{3}{*}{Cloudcast $\downarrow$}
      & Mean & 646.3036 & 631.1295 & 622.5701
      & \cellcolor{resultsecond}618.0869
      & \cellcolor{resultfirst}618.0671
      & \cellcolor{resultfirst}618.0603 & 618.0707
      & 618.0657 & 618.0684
      & \cellcolor{resultsecond}618.0648 \\
      & Best & 627.1164 & 627.1193 & 618.0928
      & \cellcolor{resultsecond}618.0648
      & \cellcolor{resultfirst}618.0615
      & \cellcolor{resultfirst}618.0596 & 618.0699
      & \cellcolor{resultsecond}618.0616
      & \cellcolor{resultfirst}618.0596
      & \cellcolor{resultfirst}618.0596 \\
      & \stdcell{Std} & \stdcell{17.30} & \stdcell{6.939}
      & \stdcell{4.454} & \stdcell{0.0198} & \stdcell{0.00485}
      & \stdcell{0.000820} & \stdcell{0.00126} & \stdcell{0.00651}
      & \stdcell{0.00791} & \stdcell{0.00890} \\
    \problemgroupbreak
    \multirow{3}{*}{EPLB}
      & Mean & 0.1328 & 0.143327 & 0.1423
      & \cellcolor{resultfirst}0.1437
      & \cellcolor{resultsecond}0.143332
      & 0.1437 & 0.1439 & 0.1445
      & \cellcolor{resultfirst}0.1450
      & \cellcolor{resultsecond}0.1448 \\
      & Best & 0.1426 & 0.143403 & 0.1426
      & \cellcolor{resultfirst}0.1442
      & \cellcolor{resultsecond}0.143415
      & 0.1439 & 0.1443 & 0.1446
      & \cellcolor{resultfirst}0.1452
      & \cellcolor{resultsecond}0.1449 \\
      & \stdcell{Std} & \stdcell{0.00851} & \stdcell{6.60e-5}
      & \stdcell{0.000359} & \stdcell{0.000777} & \stdcell{0.000119}
      & \stdcell{0.000203} & \stdcell{0.000509} & \stdcell{0.000185}
      & \stdcell{0.000153} & \stdcell{0.000216} \\
    \problemgroupbreak
    \multirow{3}{*}{LLM-SQL}
      & Mean & 0.6794 & \cellcolor{resultsecond}0.7148 & 0.6742
      & 0.6890 & \cellcolor{resultfirst}0.7206
      & 0.7229 & 0.7232 & 0.7240
      & \cellcolor{resultsecond}0.7318
      & \cellcolor{resultfirst}0.7398 \\
      & Best & 0.7110 & \cellcolor{resultsecond}0.7228 & 0.7094
      & 0.6934 & \cellcolor{resultfirst}0.7276
      & 0.7245 & 0.7329 & 0.7245
      & \cellcolor{resultsecond}0.7365
      & \cellcolor{resultfirst}0.7437 \\
      & \stdcell{Std} & \stdcell{0.0279} & \stdcell{0.0111}
      & \stdcell{0.0372} & \stdcell{0.00617} & \stdcell{0.00814}
      & \stdcell{0.00145} & \stdcell{0.00920} & \stdcell{0.000624}
      & \stdcell{0.00636} & \stdcell{0.00386} \\
    \problemgroupbreak
    \multirow{3}{*}{Prism}
      & Mean & 26.246 & 25.359
      & \cellcolor{resultsecond}26.252 & 26.248
      & \cellcolor{resultfirst}26.256
      & \cellcolor{resultfirst}26.256
      & \cellcolor{resultfirst}26.256
      & \cellcolor{resultfirst}26.256
      & \cellcolor{resultfirst}26.256
      & \cellcolor{resultfirst}26.256 \\
      & Best & \cellcolor{resultfirst}26.256
      & \cellcolor{resultsecond}26.233
      & \cellcolor{resultfirst}26.256
      & \cellcolor{resultfirst}26.256
      & \cellcolor{resultfirst}26.256
      & \cellcolor{resultfirst}26.256
      & \cellcolor{resultfirst}26.256
      & \cellcolor{resultfirst}26.256
      & \cellcolor{resultfirst}26.256
      & \cellcolor{resultfirst}26.256 \\
      & \stdcell{Std} & \stdcell{0.0118} & \stdcell{1.469}
      & \stdcell{0.00352} & \stdcell{0.0131} & \stdcell{0}
      & \stdcell{0} & \stdcell{0} & \stdcell{0} & \stdcell{0} & \stdcell{0} \\
    \problemgroupbreak
    \multirow{3}{*}{TXN}
      & Mean & 3{,}843 & 3{,}928 & 4{,}078
      & \cellcolor{resultfirst}4{,}280
      & \cellcolor{resultsecond}4{,}203
      & 4{,}281 & 4{,}292 & 4{,}498
      & \cellcolor{resultsecond}4{,}532
      & \cellcolor{resultfirst}4{,}573 \\
      & Best & 3{,}906 & 4{,}016
      & \cellcolor{resultfirst}4{,}386
      & \cellcolor{resultsecond}4{,}329 & 4{,}292
      & 4{,}405 & 4{,}310
      & \cellcolor{resultsecond}4{,}566
      & \cellcolor{resultfirst}4{,}587
      & \cellcolor{resultfirst}4{,}587 \\
      & \stdcell{Std} & \stdcell{96.89} & \stdcell{76.90}
      & \stdcell{373.0} & \stdcell{68.76} & \terminalglmstd{73.54}
      & \stdcell{215.4} & \stdcell{31.77} & \stdcell{58.85}
      & \stdcell{62.48} & \stdcell{24.08} \\
    \noalign{\vskip -4.5pt}
    \bottomrule
  \end{tabularx}
\end{minipage}
\end{center}

%% file: sections/frontier_scores_table.tex
\begin{tabular}{c!{\color{white}\vrule width 0.8pt}*{5}{c!{\color{white}\vrule width 0.8pt}}c}
\noalign{\color{black}\hrule height 0.8pt}
Problem ID & OpenEvolve & EvoX & AdaEvolve & CORAL-1 & CORAL-4 & EvoTrias \\
\noalign{\color{black}\hrule height 0.6pt}
207 & \cellcolor[HTML]{3B917A}\textcolor{white}{\fontfamily{phv}\selectfont 67.42} & \cellcolor[HTML]{F2F7F5}\fontfamily{phv}\selectfont 0.00 & \cellcolor[HTML]{F2F7F5}\fontfamily{phv}\selectfont 0.00 & \cellcolor[HTML]{3A9079}\textcolor{white}{\fontfamily{phv}\selectfont 67.90} & \cellcolor[HTML]{82BBAA}\fontfamily{phv}\selectfont 45.32 & \cellcolor[HTML]{58A38E}\textcolor{white}{\fontfamily{phv}\selectfont 57.13} \\
\noalign{\color{white}\hrule height 0.8pt}
209 & \cellcolor[HTML]{F2F7F5}\fontfamily{phv}\selectfont 0.00 & \cellcolor[HTML]{F2F7F5}\fontfamily{phv}\selectfont 0.00 & \cellcolor[HTML]{F2F7F5}\fontfamily{phv}\selectfont 0.00 & \cellcolor[HTML]{F2F7F5}\fontfamily{phv}\selectfont 0.00 & \cellcolor[HTML]{DFECE7}\fontfamily{phv}\selectfont 10.99 & \cellcolor[HTML]{F2F7F5}\fontfamily{phv}\selectfont 0.00 \\
\noalign{\color{white}\hrule height 0.8pt}
210 & \cellcolor[HTML]{A6CEC1}\fontfamily{phv}\selectfont 34.67 & \cellcolor[HTML]{7EB9A7}\fontfamily{phv}\selectfont 46.72 & \cellcolor[HTML]{B8D7CD}\fontfamily{phv}\selectfont 28.35 & \cellcolor[HTML]{C3DDD4}\fontfamily{phv}\selectfont 24.30 & \cellcolor[HTML]{9CC9BB}\fontfamily{phv}\selectfont 38.04 & \cellcolor[HTML]{B9D8CD}\fontfamily{phv}\selectfont 27.98 \\
\noalign{\color{white}\hrule height 0.8pt}
211 & \cellcolor[HTML]{156A56}\textcolor{white}{\fontfamily{phv}\selectfont 86.95} & \cellcolor[HTML]{156A56}\textcolor{white}{\fontfamily{phv}\selectfont 87.10} & \cellcolor[HTML]{166C58}\textcolor{white}{\fontfamily{phv}\selectfont 85.68} & \cellcolor[HTML]{4A9B85}\textcolor{white}{\fontfamily{phv}\selectfont 61.58} & \cellcolor[HTML]{166C58}\textcolor{white}{\fontfamily{phv}\selectfont 85.60} & \cellcolor[HTML]{166C58}\textcolor{white}{\fontfamily{phv}\selectfont 85.89} \\
\noalign{\color{white}\hrule height 0.8pt}
212 & \cellcolor[HTML]{F2F7F5}\fontfamily{phv}\selectfont 0.00 & \cellcolor[HTML]{F2F7F5}\fontfamily{phv}\selectfont 0.00 & \cellcolor[HTML]{F2F7F5}\fontfamily{phv}\selectfont 0.00 & \cellcolor[HTML]{C0DCD3}\fontfamily{phv}\selectfont 25.00 & \cellcolor[HTML]{C0DCD3}\fontfamily{phv}\selectfont 25.00 & \cellcolor[HTML]{C0DCD3}\fontfamily{phv}\selectfont 25.00 \\
\noalign{\color{white}\hrule height 0.8pt}
213 & \cellcolor[HTML]{18705B}\textcolor{white}{\fontfamily{phv}\selectfont 84.25} & \cellcolor[HTML]{A9D0C3}\fontfamily{phv}\selectfont 33.33 & \cellcolor[HTML]{18705B}\textcolor{white}{\fontfamily{phv}\selectfont 84.25} & \cellcolor[HTML]{5FA793}\textcolor{white}{\fontfamily{phv}\selectfont 55.09} & \cellcolor[HTML]{0E5445}\textcolor{white}{\fontfamily{phv}\selectfont 96.23} & \cellcolor[HTML]{0D5243}\textcolor{white}{\fontfamily{phv}\selectfont 97.26} \\
\noalign{\color{white}\hrule height 0.8pt}
214 & \cellcolor[HTML]{A8CFC3}\fontfamily{phv}\selectfont 33.94 & \cellcolor[HTML]{439680}\textcolor{white}{\fontfamily{phv}\selectfont 64.35} & \cellcolor[HTML]{4B9C86}\textcolor{white}{\fontfamily{phv}\selectfont 61.01} & \cellcolor[HTML]{1E7C65}\textcolor{white}{\fontfamily{phv}\selectfont 78.66} & \cellcolor[HTML]{2B856F}\textcolor{white}{\fontfamily{phv}\selectfont 73.74} & \cellcolor[HTML]{19745E}\textcolor{white}{\fontfamily{phv}\selectfont 82.74} \\
\noalign{\color{white}\hrule height 0.8pt}
217 & \cellcolor[HTML]{439680}\textcolor{white}{\fontfamily{phv}\selectfont 64.11} & \cellcolor[HTML]{469882}\textcolor{white}{\fontfamily{phv}\selectfont 63.14} & \cellcolor[HTML]{ADD2C6}\fontfamily{phv}\selectfont 31.70 & \cellcolor[HTML]{AAD0C4}\fontfamily{phv}\selectfont 33.08 & \cellcolor[HTML]{A9D0C3}\fontfamily{phv}\selectfont 33.33 & \cellcolor[HTML]{388E78}\textcolor{white}{\fontfamily{phv}\selectfont 68.54} \\
\noalign{\color{white}\hrule height 0.8pt}
220 & \cellcolor[HTML]{348C75}\textcolor{white}{\fontfamily{phv}\selectfont 70.00} & \cellcolor[HTML]{0B4C3E}\textcolor{white}{\fontfamily{phv}\selectfont 100.00} & \cellcolor[HTML]{348C75}\textcolor{white}{\fontfamily{phv}\selectfont 70.00} & \cellcolor[HTML]{0B4C3E}\textcolor{white}{\fontfamily{phv}\selectfont 100.00} & \cellcolor[HTML]{0B4C3E}\textcolor{white}{\fontfamily{phv}\selectfont 100.00} & \cellcolor[HTML]{0B4C3E}\textcolor{white}{\fontfamily{phv}\selectfont 100.00} \\
\noalign{\color{white}\hrule height 0.8pt}
222 & \cellcolor[HTML]{D7E8E2}\fontfamily{phv}\selectfont 15.52 & \cellcolor[HTML]{D7E8E2}\fontfamily{phv}\selectfont 15.41 & \cellcolor[HTML]{D7E8E2}\fontfamily{phv}\selectfont 15.52 & \cellcolor[HTML]{D7E8E2}\fontfamily{phv}\selectfont 15.37 & \cellcolor[HTML]{55A28D}\textcolor{white}{\fontfamily{phv}\selectfont 57.91} & \cellcolor[HTML]{1E7C65}\textcolor{white}{\fontfamily{phv}\selectfont 78.59} \\
\noalign{\color{white}\hrule height 0.8pt}
225 & \cellcolor[HTML]{F2F7F5}\fontfamily{phv}\selectfont 0.00 & \cellcolor[HTML]{F2F7F5}\fontfamily{phv}\selectfont 0.00 & \cellcolor[HTML]{F2F7F5}\fontfamily{phv}\selectfont 0.00 & \cellcolor[HTML]{F2F7F5}\fontfamily{phv}\selectfont 0.00 & \cellcolor[HTML]{F2F7F5}\fontfamily{phv}\selectfont 0.00 & \cellcolor[HTML]{F2F7F5}\fontfamily{phv}\selectfont 0.00 \\
\noalign{\color{white}\hrule height 0.8pt}
227 & \cellcolor[HTML]{EEF5F2}\fontfamily{phv}\selectfont 2.45 & \cellcolor[HTML]{EEF5F2}\fontfamily{phv}\selectfont 2.45 & \cellcolor[HTML]{EEF5F2}\fontfamily{phv}\selectfont 2.45 & \cellcolor[HTML]{EEF5F2}\fontfamily{phv}\selectfont 2.45 & \cellcolor[HTML]{EEF5F2}\fontfamily{phv}\selectfont 2.46 & \cellcolor[HTML]{EEF5F2}\fontfamily{phv}\selectfont 2.45 \\
\noalign{\color{white}\hrule height 0.8pt}
228 & \cellcolor[HTML]{0B4C3E}\textcolor{white}{\fontfamily{phv}\selectfont 100.00} & \cellcolor[HTML]{0B4C3E}\textcolor{white}{\fontfamily{phv}\selectfont 100.00} & \cellcolor[HTML]{0B4C3E}\textcolor{white}{\fontfamily{phv}\selectfont 100.00} & \cellcolor[HTML]{0B4C3E}\textcolor{white}{\fontfamily{phv}\selectfont 100.00} & \cellcolor[HTML]{0B4C3E}\textcolor{white}{\fontfamily{phv}\selectfont 100.00} & \cellcolor[HTML]{0B4C3E}\textcolor{white}{\fontfamily{phv}\selectfont 100.00} \\
\noalign{\color{white}\hrule height 0.8pt}
229 & \cellcolor[HTML]{0B4C3E}\textcolor{white}{\fontfamily{phv}\selectfont 100.00} & \cellcolor[HTML]{0B4C3E}\textcolor{white}{\fontfamily{phv}\selectfont 99.67} & \cellcolor[HTML]{0B4C3E}\textcolor{white}{\fontfamily{phv}\selectfont 100.00} & \cellcolor[HTML]{0B4C3E}\textcolor{white}{\fontfamily{phv}\selectfont 100.00} & \cellcolor[HTML]{0B4C3E}\textcolor{white}{\fontfamily{phv}\selectfont 100.00} & \cellcolor[HTML]{0C4E3F}\textcolor{white}{\fontfamily{phv}\selectfont 99.00} \\
\noalign{\color{white}\hrule height 0.8pt}
231 & \cellcolor[HTML]{F2F7F5}\fontfamily{phv}\selectfont 0.00 & \cellcolor[HTML]{F2F7F5}\fontfamily{phv}\selectfont 0.00 & \cellcolor[HTML]{F2F7F5}\fontfamily{phv}\selectfont 0.00 & \cellcolor[HTML]{F2F7F5}\fontfamily{phv}\selectfont 0.00 & \cellcolor[HTML]{F2F7F5}\fontfamily{phv}\selectfont 0.00 & \cellcolor[HTML]{F2F7F5}\fontfamily{phv}\selectfont 0.00 \\
\noalign{\color{white}\hrule height 0.8pt}
233 & \cellcolor[HTML]{AAD0C4}\fontfamily{phv}\selectfont 33.20 & \cellcolor[HTML]{B5D6CB}\fontfamily{phv}\selectfont 29.11 & \cellcolor[HTML]{4B9C86}\textcolor{white}{\fontfamily{phv}\selectfont 61.25} & \cellcolor[HTML]{B2D4C9}\fontfamily{phv}\selectfont 30.20 & \cellcolor[HTML]{B0D3C8}\fontfamily{phv}\selectfont 31.13 & \cellcolor[HTML]{51A08A}\textcolor{white}{\fontfamily{phv}\selectfont 59.11} \\
\noalign{\color{white}\hrule height 0.8pt}
239 & \cellcolor[HTML]{62A995}\fontfamily{phv}\selectfont 54.56 & \cellcolor[HTML]{459881}\textcolor{white}{\fontfamily{phv}\selectfont 63.39} & \cellcolor[HTML]{358C76}\textcolor{white}{\fontfamily{phv}\selectfont 69.89} & \cellcolor[HTML]{28836C}\textcolor{white}{\fontfamily{phv}\selectfont 74.67} & \cellcolor[HTML]{4D9D87}\textcolor{white}{\fontfamily{phv}\selectfont 60.54} & \cellcolor[HTML]{0B4C3E}\textcolor{white}{\fontfamily{phv}\selectfont 100.00} \\
\noalign{\color{white}\hrule height 0.8pt}
241 & \cellcolor[HTML]{105A49}\textcolor{white}{\fontfamily{phv}\selectfont 93.81} & \cellcolor[HTML]{0E5344}\textcolor{white}{\fontfamily{phv}\selectfont 96.67} & \cellcolor[HTML]{0F5948}\textcolor{white}{\fontfamily{phv}\selectfont 94.28} & \cellcolor[HTML]{166B57}\textcolor{white}{\fontfamily{phv}\selectfont 86.67} & \cellcolor[HTML]{105A49}\textcolor{white}{\fontfamily{phv}\selectfont 93.81} & \cellcolor[HTML]{105B4A}\textcolor{white}{\fontfamily{phv}\selectfont 93.33} \\
\noalign{\color{white}\hrule height 0.8pt}
243 & \cellcolor[HTML]{27826C}\textcolor{white}{\fontfamily{phv}\selectfont 75.00} & \cellcolor[HTML]{0B4C3E}\textcolor{white}{\fontfamily{phv}\selectfont 100.00} & \cellcolor[HTML]{27826C}\textcolor{white}{\fontfamily{phv}\selectfont 75.00} & \cellcolor[HTML]{27826C}\textcolor{white}{\fontfamily{phv}\selectfont 75.00} & \cellcolor[HTML]{27826C}\textcolor{white}{\fontfamily{phv}\selectfont 75.00} & \cellcolor[HTML]{27826C}\textcolor{white}{\fontfamily{phv}\selectfont 75.00} \\
\noalign{\color{white}\hrule height 0.8pt}
245 & \cellcolor[HTML]{136452}\textcolor{white}{\fontfamily{phv}\selectfont 89.17} & \cellcolor[HTML]{136350}\textcolor{white}{\fontfamily{phv}\selectfont 89.96} & \cellcolor[HTML]{136350}\textcolor{white}{\fontfamily{phv}\selectfont 89.96} & \cellcolor[HTML]{136452}\textcolor{white}{\fontfamily{phv}\selectfont 89.17} & \cellcolor[HTML]{0B4C3E}\textcolor{white}{\fontfamily{phv}\selectfont 100.00} & \cellcolor[HTML]{136351}\textcolor{white}{\fontfamily{phv}\selectfont 89.64} \\
\noalign{\color{white}\hrule height 0.8pt}
247 & \cellcolor[HTML]{EFF5F3}\fontfamily{phv}\selectfont 2.10 & \cellcolor[HTML]{DDEBE6}\fontfamily{phv}\selectfont 12.07 & \cellcolor[HTML]{F2F7F5}\fontfamily{phv}\selectfont 0.00 & \cellcolor[HTML]{F2F7F5}\fontfamily{phv}\selectfont 0.00 & \cellcolor[HTML]{E1EDE9}\fontfamily{phv}\selectfont 9.91 & \cellcolor[HTML]{F2F7F5}\fontfamily{phv}\selectfont 0.00 \\
\noalign{\color{white}\hrule height 0.8pt}
248 & \cellcolor[HTML]{0B4C3E}\textcolor{white}{\fontfamily{phv}\selectfont 100.00} & \cellcolor[HTML]{0B4C3E}\textcolor{white}{\fontfamily{phv}\selectfont 100.00} & \cellcolor[HTML]{0B4C3E}\textcolor{white}{\fontfamily{phv}\selectfont 100.00} & \cellcolor[HTML]{0B4C3E}\textcolor{white}{\fontfamily{phv}\selectfont 100.00} & \cellcolor[HTML]{0B4C3E}\textcolor{white}{\fontfamily{phv}\selectfont 100.00} & \cellcolor[HTML]{0B4C3E}\textcolor{white}{\fontfamily{phv}\selectfont 100.00} \\
\noalign{\color{white}\hrule height 0.8pt}
249 & \cellcolor[HTML]{0B4C3E}\textcolor{white}{\fontfamily{phv}\selectfont 100.00} & \cellcolor[HTML]{0B4C3E}\textcolor{white}{\fontfamily{phv}\selectfont 100.00} & \cellcolor[HTML]{0B4C3E}\textcolor{white}{\fontfamily{phv}\selectfont 100.00} & \cellcolor[HTML]{0B4C3E}\textcolor{white}{\fontfamily{phv}\selectfont 100.00} & \cellcolor[HTML]{0B4C3E}\textcolor{white}{\fontfamily{phv}\selectfont 100.00} & \cellcolor[HTML]{0B4C3E}\textcolor{white}{\fontfamily{phv}\selectfont 100.00} \\
\noalign{\color{white}\hrule height 0.8pt}
252 & \cellcolor[HTML]{2D8770}\textcolor{white}{\fontfamily{phv}\selectfont 72.67} & \cellcolor[HTML]{1A7862}\textcolor{white}{\fontfamily{phv}\selectfont 80.64} & \cellcolor[HTML]{1A7660}\textcolor{white}{\fontfamily{phv}\selectfont 81.37} & \cellcolor[HTML]{19755F}\textcolor{white}{\fontfamily{phv}\selectfont 81.82} & \cellcolor[HTML]{0B4C3E}\textcolor{white}{\fontfamily{phv}\selectfont 100.00} & \cellcolor[HTML]{0B4C3E}\textcolor{white}{\fontfamily{phv}\selectfont 100.00} \\
\noalign{\color{white}\hrule height 0.8pt}
253 & \cellcolor[HTML]{0B4C3E}\textcolor{white}{\fontfamily{phv}\selectfont 100.00} & \cellcolor[HTML]{0B4C3E}\textcolor{white}{\fontfamily{phv}\selectfont 100.00} & \cellcolor[HTML]{0B4C3E}\textcolor{white}{\fontfamily{phv}\selectfont 100.00} & \cellcolor[HTML]{0B4C3E}\textcolor{white}{\fontfamily{phv}\selectfont 100.00} & \cellcolor[HTML]{0B4C3E}\textcolor{white}{\fontfamily{phv}\selectfont 100.00} & \cellcolor[HTML]{0B4C3E}\textcolor{white}{\fontfamily{phv}\selectfont 100.00} \\
\noalign{\color{white}\hrule height 0.8pt}
254 & \cellcolor[HTML]{BEDAD1}\fontfamily{phv}\selectfont 25.91 & \cellcolor[HTML]{A6CEC1}\fontfamily{phv}\selectfont 34.68 & \cellcolor[HTML]{BEDAD1}\fontfamily{phv}\selectfont 25.91 & \cellcolor[HTML]{B1D4C8}\fontfamily{phv}\selectfont 30.49 & \cellcolor[HTML]{A1CCBE}\fontfamily{phv}\selectfont 36.23 & \cellcolor[HTML]{A7CFC2}\fontfamily{phv}\selectfont 34.06 \\
\noalign{\color{white}\hrule height 0.8pt}
255 & \cellcolor[HTML]{8CC1B1}\fontfamily{phv}\selectfont 42.70 & \cellcolor[HTML]{82BBAA}\fontfamily{phv}\selectfont 45.40 & \cellcolor[HTML]{82BBAA}\fontfamily{phv}\selectfont 45.56 & \cellcolor[HTML]{88BEAE}\fontfamily{phv}\selectfont 43.92 & \cellcolor[HTML]{7BB7A5}\fontfamily{phv}\selectfont 47.64 & \cellcolor[HTML]{74B3A1}\fontfamily{phv}\selectfont 49.26 \\
\noalign{\color{white}\hrule height 0.8pt}
256 & \cellcolor[HTML]{98C7B8}\fontfamily{phv}\selectfont 39.44 & \cellcolor[HTML]{85BDAC}\fontfamily{phv}\selectfont 44.80 & \cellcolor[HTML]{85BDAC}\fontfamily{phv}\selectfont 44.80 & \cellcolor[HTML]{85BDAC}\fontfamily{phv}\selectfont 44.59 & \cellcolor[HTML]{85BDAC}\fontfamily{phv}\selectfont 44.80 & \cellcolor[HTML]{85BDAC}\fontfamily{phv}\selectfont 44.80 \\
\noalign{\color{white}\hrule height 0.8pt}
257 & \cellcolor[HTML]{0B4C3E}\textcolor{white}{\fontfamily{phv}\selectfont 100.00} & \cellcolor[HTML]{0B4C3E}\textcolor{white}{\fontfamily{phv}\selectfont 100.00} & \cellcolor[HTML]{0B4C3E}\textcolor{white}{\fontfamily{phv}\selectfont 100.00} & \cellcolor[HTML]{0B4C3E}\textcolor{white}{\fontfamily{phv}\selectfont 100.00} & \cellcolor[HTML]{0B4C3E}\textcolor{white}{\fontfamily{phv}\selectfont 100.00} & \cellcolor[HTML]{0B4C3E}\textcolor{white}{\fontfamily{phv}\selectfont 100.00} \\
\noalign{\color{white}\hrule height 0.8pt}
258 & \cellcolor[HTML]{125F4D}\textcolor{white}{\fontfamily{phv}\selectfont 91.72} & \cellcolor[HTML]{12614F}\textcolor{white}{\fontfamily{phv}\selectfont 90.91} & \cellcolor[HTML]{125F4D}\textcolor{white}{\fontfamily{phv}\selectfont 91.69} & \cellcolor[HTML]{156854}\textcolor{white}{\fontfamily{phv}\selectfont 87.75} & \cellcolor[HTML]{13624F}\textcolor{white}{\fontfamily{phv}\selectfont 90.42} & \cellcolor[HTML]{115C4B}\textcolor{white}{\fontfamily{phv}\selectfont 92.87} \\
\noalign{\color{black}\hrule height 0.6pt}
{\fontfamily{phv}\selectfont Mean} & {\fontfamily{phv}\selectfont 55.99} & {\fontfamily{phv}\selectfont 56.66} & {\fontfamily{phv}\selectfont 55.29} & {\fontfamily{phv}\selectfont 56.92} & {\fontfamily{phv}\selectfont 61.94} & \makebox[5.1em][l]{{\fontfamily{phv}\selectfont 65.42}\hspace{0.2em}\raisebox{-0.45ex}{\fontfamily{phv}\selectfont\fontsize{4.8}{5.2}\selectfont\bfseries\textcolor[HTML]{178A4B}{+3.49}}} \\
{\fontfamily{phv}\selectfont Median} & {\fontfamily{phv}\selectfont 65.77} & {\fontfamily{phv}\selectfont 63.27} & {\fontfamily{phv}\selectfont 65.57} & {\fontfamily{phv}\selectfont 64.74} & {\fontfamily{phv}\selectfont 67.14} & \makebox[5.1em][l]{{\fontfamily{phv}\selectfont 80.67}\hspace{0.2em}\raisebox{-0.45ex}{\fontfamily{phv}\selectfont\fontsize{4.8}{5.2}\selectfont\bfseries\textcolor[HTML]{178A4B}{+13.53}}} \\
\noalign{\color{black}\hrule height 0.8pt}
\end{tabular}

%% file: sections/appendix/c_agentic_baselines_table.tex
\begin{table}[t]
  \centering
  \captionsetup{hypcap=false}
  \caption{Controller-centric baselines with SkyDiscover's built-in agentic
  proposer (GLM-5.2, mean $\pm$ sample std over three runs). Each group lists
  the published single-call configuration, the same controller with
  \texttt{--agentic} enabled, the relative change this switch produces, and the
  gap that remains to EvoTrias. Green/red cells mark gains/losses relative to
  the single-call configuration; saturation encodes magnitude.}
  \label{tab:agentic-baselines}
  \vspace{2pt}
  \footnotesize
  \setlength{\tabcolsep}{7pt}
  \renewcommand{\arraystretch}{1.12}
  \newcommand{\agstd}[1]{{\fontsize{6.2}{6.6}\selectfont\color{black!70}$\pm$#1}}
  \newcommand{\agsub}[1]{\hspace{0.9em}#1}
  \begin{tabularx}{\textwidth}{l *{3}{>{\raggedleft\arraybackslash}X}}
    \toprule
    Method & Circle Packing (Square) & Signal Processing & Heilbronn Triangles \\
    \midrule
    \rowcolor{oursblue}
    \textbf{EvoTrias} (reference)
      & \textbf{2.6348\agstd{0.0019}} & \textbf{0.7329\agstd{0.0058}} & \textbf{0.0335\agstd{0.0022}} \\
    \midrule
    OpenEvolve
      & 2.5432\agstd{0.0030} & 0.6373\agstd{0.0131} & 0.0253\agstd{0.0012} \\
    \agsub{$+$ agentic proposer}
      & 2.6178\agstd{0.0267} & 0.6804\agstd{0.0140} & 0.0304\agstd{0.0019} \\
    \agsub{$\Delta$ vs.\ single-call}
      & \cellcolor[RGB]{181,230,193}$+$2.93\% & \cellcolor[RGB]{153,220,170}$+$6.76\% & \cellcolor[RGB]{126,211,147}$+$20.12\% \\
    \agsub{$\Delta$ vs.\ EvoTrias}
      & \textcolor{black!60}{\textit{$-$0.65\%}} & \textcolor{black!60}{\textit{$-$7.17\%}} & \textcolor{black!60}{\textit{$-$9.06\%}} \\
    \midrule
    AdaEvolve
      & 2.6315\agstd{0.0024} & 0.6961\agstd{0.0226} & 0.0318\agstd{0.0013} \\
    \agsub{$+$ agentic proposer}
      & 2.6150\agstd{0.0144} & 0.7096\agstd{0.0353} & 0.0322\agstd{0.0010} \\
    \agsub{$\Delta$ vs.\ single-call}
      & \cellcolor[RGB]{251,210,212}$-$0.63\% & \cellcolor[RGB]{192,233,202}$+$1.93\% & \cellcolor[RGB]{202,237,211}$+$1.10\% \\
    \agsub{$\Delta$ vs.\ EvoTrias}
      & \textcolor{black!60}{\textit{$-$0.75\%}} & \textcolor{black!60}{\textit{$-$3.19\%}} & \textcolor{black!60}{\textit{$-$3.82\%}} \\
    \midrule
    EvoX
      & 2.6096\agstd{0.0012} & 0.7098\agstd{0.0095} & 0.0302\agstd{0.0040} \\
    \agsub{$+$ agentic proposer}
      & 2.6118\agstd{0.0148} & 0.7012\agstd{0.0112} & 0.0302\agstd{0.0016} \\
    \agsub{$\Delta$ vs.\ single-call}
      & \cellcolor[RGB]{226,245,231}$+$0.08\% & \cellcolor[RGB]{250,200,202}$-$1.22\% & \cellcolor[RGB]{232,247,236}$+$0.01\% \\
    \agsub{$\Delta$ vs.\ EvoTrias}
      & \textcolor{black!60}{\textit{$-$0.88\%}} & \textcolor{black!60}{\textit{$-$4.33\%}} & \textcolor{black!60}{\textit{$-$9.67\%}} \\
    \bottomrule
  \end{tabularx}
\end{table}

%% file: sections/appendix/d_prompts.tex
\clearpage
\section{Prompts}
\label{app:prompts}

This appendix reproduces the prompts of the four agent roles in a run: the
orchestrator and the three subagents it dispatches (init-explorer, task
executor, and meta-guidance generator), together with the two fixed prompts
of the Notes channel that the task executor follows in place of a fifth.
Omitted passages are marked
[omitted: ...], and [x] marks values filled at dispatch time. The sources call
the evaluator a \emph{verifier}, and the quoted wording is kept as written.

\input{sections/appendix/d_prompts/orchestrator.tex}
\input{sections/appendix/d_prompts/initial_direction.tex}
\input{sections/appendix/d_prompts/task_executor.tex}
\input{sections/appendix/d_prompts/notes.tex}

\input{sections/appendix/d_prompts/meta-guidance.tex}

%% file: sections/appendix/d_prompts/orchestrator.tex
\subsection{Orchestrator}
\label{app:prompt-orchestrator}

\begin{promptcard}{Orchestrator}{evolve master skill prompt, verbatim excerpts}

You are now running the \textbf{self-evolution master skill}, the
orchestrator. Your responsibilities are: initialize a run, maintain the
task queue, react to MCP convergence/meta-evolve actions, delegate
requested meta-guidance generation, stop replenishment only on hard
budget/external stops, wait for already launched tasks to finish, and
call Evolution Controller MCP to generate a deterministic run summary
report.

\textbf{Orchestrator Boundary Constraint}
[omitted: the orchestrator only schedules, calls MCP, and dispatches
templates; it never touches task contents, candidates, evaluators, or
meta-guidances]

\textbf{Dispatch Template Constraint (ALL subagent dispatches)}
[omitted: subagent dispatch prompts are fixed templates of parameter lines; 
subagents are dispatched in the shared working directory]

\textbf{task\_description Fidelity Constraint}
[omitted: task\_description must be identical to the user's original submitted
input]

\textbf{max\_attempts Dispatch Constraint}
[omitted: max\_attempts is set explicitly on every task-executor dispatch,
with a maximum of 3 attempts for every task, including root tasks]

\textbf{Architecture Roles}

\begin{tabularx}{\linewidth}{@{}>{\raggedright\arraybackslash}p{0.3\linewidth}L@{}}
\toprule
Component & Responsibility \\
\midrule
This skill (orchestrator) & Initialize run, maintain task queue, route
MCP action, dispatch meta-guidance requests by ID, finalization \\
meta-guidance-generator agent & Fetch one pending request, inspect
generation evidence, generate and commit guidance, return a compact
result \\
task-executor agent & Execute the full lifecycle of individual tasks \\
Evolution Controller MCP & Parent/inspiration selection, attempt
recording, scheduler-state persistence, convergence/meta-guidance state,
run summary generation \\
\bottomrule
\end{tabularx}

\textbf{Step 1: Confirm Run Parameters}

Before any run creation or MCP call, determine and confirm the three run
parameters. They are independent knobs - never derive one from another:

\begin{tabularx}{\linewidth}{@{}>{\raggedright\arraybackslash}p{0.3\linewidth}Lc@{}}
\toprule
Parameter & Meaning & Default \\
\midrule
LINEAGE\_ROOT\_NUMBER & Number of initial exploration directions
generated at initialization; equals the number of lineage root tasks the
run establishes & 4 \\
MAX\_CONCURRENT\_TASKS & Maximum number of task-executors running in
parallel & 4 \\
MAX\_TOTAL\_TASKS & Hard budget: total number of tasks the run may create
before stopping & 50 \\
\bottomrule
\end{tabularx}

1. Extract each value from the user's original request when stated:
[omitted: example phrasings for each of the three parameters]\\
2. For any value not stated, use its default from the table.\\
3. Present all three to the user in one compact message - name, chosen
value, whether it was extracted or defaulted, and a one-line meaning -
and ask for confirmation or adjustments.\\
4. Only after the user confirms, proceed to Step 2 with the confirmed
values. These confirmed values are referenced by name throughout the
rest of this skill.

\textbf{Step 2: Run Initialization}

1. Derive task\_slug from the user's original input. [omitted:
slug-derivation and normalization rules]\\
2. Generate run\_id: [task\_slug]\_YYYYMMDDHHMMSS.\\
\hspace*{1.2em}- Use the current local start time with second-level
precision. Use the Bash tool to execute date +\%Y\%m\%d\%H\%M\%S.\\
3. Register the confirmed parameters with the controller:\\
\hspace*{1.2em}- Call configure\_run(run\_id,
max\_total\_tasks=[MAX\_TOTAL\_TASKS],
lineage\_root\_number=[LINEAGE\_ROOT\_NUMBER],
max\_concurrent\_tasks=[MAX\_CONCURRENT\_TASKS],
task\_description=[original user input]) BEFORE any other controller MCP
call for this run. The controller records all three values; they are
locked for the rest of the run.\\
\hspace*{1.2em}- [omitted: dashboard startup banner]\\
4. Create the run workspace directory and persistent state files.
All runs/ paths are relative to the project root (cwd), not to the skill's base
directory. Use Bash to execute: mkdir -p "runs/[run\_id]"

\textbf{Step 3: Generate Initial Exploration Directions}

Invoke the evotrias\_plugin:init-explorer agent using the
\textbf{init-explorer dispatch template}. [omitted: parameters to fill]

Wait for evotrias\_plugin:init-explorer  to complete before proceeding to 
Step 4.

\textbf{Step 4: Task Queue Scheduling}

MCP owns batch sizing: check\_convergence tells you how many tasks to
create via tasks\_to\_create, and you dispatch exactly that many - never
more, never fewer, never inferred. Any returned value, including 0, is a
normal scheduling decision; on 0 dispatch nothing and keep looping.

Task completion data is persisted by the controller as subagents record
their attempts; notifications merely prompt the orchestrator to call
check\_convergence again.

\textbf{Data Flow}
[omitted: the per-run directory runs/[run\_id]/, relative to the project root]

\textbf{Scheduling Pseudocode}

first\_batch = true\\
just\_after\_meta = false\\

WHILE true:\\
\hspace*{1.2em}// Wait-gate: every check\_convergence call (except the
first and the one right after a meta-\\
\hspace*{1.2em}// guidance commit) must be preceded by at least one
task-executor finishing. Without this gate\\
\hspace*{1.2em}// the loop would hammer the MCP in a tight cycle.\\
\hspace*{1.2em}IF not first\_batch and not just\_after\_meta:\\
\hspace*{2.4em}wait for at least one task-executor to finish\\
\hspace*{1.2em}just\_after\_meta = false

\hspace*{1.2em}// The first check\_convergence call is mandatory: on
first\_batch the wait-gate is skipped, but\\
\hspace*{1.2em}// the call itself must NOT be. Every batch's task count
comes only from c.tasks\_to\_create -\\
\hspace*{1.2em}// never inferred or hardcoded. tasks\_to\_create == 0 with
action "continue" is normal - dispatch\\
\hspace*{1.2em}// nothing and keep looping.\\
\hspace*{1.2em}c = check\_convergence(run\_id,
concurrency=MAX\_CONCURRENT\_TASKS)

\hspace*{1.2em}IF c.action == "budget\_exhausted":\\
\hspace*{2.4em}BREAK WHILE LOOP

\hspace*{1.2em}IF c.action == "request\_meta\_guidance":\\
\hspace*{2.4em}request\_id = c.request\_id\\
\hspace*{2.4em}Dispatch to meta-guidance-generator with
run\_in\_background: false using the\\
\hspace*{3.6em}meta-guidance-generator dispatch
template (fill: run\_id, request\_id)\\
\hspace*{2.4em}Wait for the meta-guidance-generator compact result.

\hspace*{2.4em}just\_after\_meta = true\\
\hspace*{2.4em}CONTINUE \hspace*{0.6em}// re-check convergence; the
second check sees active guidance and goes to continue

\hspace*{1.2em}// action == "continue"\\
\hspace*{1.2em}// Every task, including a root task, gets at most three attempts.\\
\hspace*{1.2em}max\_attempts = 3\\
\hspace*{1.2em}REPEAT c.tasks\_to\_create TIMES:\\
\hspace*{2.4em}Dispatch to task-executor with run\_in\_background: true
using the task-executor dispatch\\
\hspace*{3.6em}template (fill: task\_description,
max\_attempts=max\_attempts).\\
\hspace*{1.2em}END REPEAT\\
\hspace*{1.2em}first\_batch = false

\hspace*{1.2em}[omitted: when and what to print to the terminal]

END WHILE

// After the while loop exits (action == budget\_exhausted), wait for
any still-running task-executors to finish, finally do finalization (see
Step 5).

[omitted: compact terminal progress rules; the orchestrator never prints
Notes or meta-guidance content in the main terminal]

\textbf{Critical Constraints}
[omitted: critical rules. e.g. stop conditions, workflow guidelines, etc.]

\textbf{Meta-Guidance Worker Dispatch}
[omitted: passes run\_id and request\_id to a meta-guidance subagent, waits
for its compact result, and loops back to check\_convergence]

\textbf{Step 5 run finalization} [omitted: call finalize\_run and generate 
final report]

[omitted: error-handling table]

\end{promptcard}

%% file: sections/appendix/d_prompts/initial_direction.tex
\subsection{Initial Directions}
\label{app:prompt-init}

\begin{promptcard}{Init-explorer}{skill prompt, verbatim excerpts}

Generate initial exploration directions for evolutionary tasks.

First, confirm these critical identifiers from the context: run\_id;
task\_description; num\_directions. CRITICAL: If any of these are missing,
ask immediately before proceeding.

Then execute the following steps in EXACT ORDER:

\textbf{Step 1: Understand Task}

1. Locate and read the problem SKILL.md if it exists. The problem
SKILL.md is typically found in the project root or in a benchmark
directory.\\
2. Follow the problem SKILL.md's public instance-selection instructions
using the original task\_description. If the skill requires an instance
or problem ID, validate it and read the public statement/metadata it
declares. Never infer a benchmark instance from an evolution task\_id.\\
3. Resolve the initial reference declared by the problem SKILL.md.
[omitted: how the initial program file is located] Read the resolved
file as initial\_program\_reference.\\
4. Understand the verification method, what constitutes a successful
solution, and the reference program's main algorithmic approach if one
exists.\\
5. Do not read verifier/evaluator scripts or judge-side resources. Only
read the problem SKILL.md and the public problem resources and initial
reference it explicitly declares.

\textbf{Step 2: Analyze Problem Structure}

Before generating directions, analyze the problem to identify its key
structural decision points - the places where a solver can meaningfully
diverge into different approaches.

Ask yourself:\\
- What are the \textbf{core decisions} a solution must make? (e.g.,
one-shot vs iterative, exact vs heuristic, deterministic vs random)\\
- What \textbf{information is available} to the solver? (e.g., feedback
after each step, no feedback at all, partial observability)\\
- What \textbf{constraints shape the solution space}? (e.g., fixed budget
of attempts, single attempt only, time limit)\\
- What \textbf{failure modes} are possible? (e.g., wrong on first try,
converges slowly, overshoots)

Write a brief problem analysis (3-5 lines) covering:\\
1. Problem type (one-shot, iterative, optimization, construction, etc.)\\
2. Available information and feedback\\
3. Key decision axes where approaches can diverge

This analysis is internal - do not save it to a file. It guides direction
generation.

\textbf{Step 3: Generate Directions}

Generate exactly [num\_directions] different directions. Each direction
must be:\\
- One line, 8-25 words in English (SHORT and PRECISE)\\
- Clearly distinct from the other directions\\
- Actionable, focusing on a specific approach\\
- Feasible and meaningful for the given task

If an initial\_program\_reference exists, use it as important context
for the task interface, baseline algorithm, and known feasible starting
route. Directions may improve, specialize, simplify, or deliberately
depart from that reference. Keep the set diverse; do not generate four
directions that merely restate the initial program's current approach.

Directions MUST be diverse enough so that following them produces
significantly different solutions. Use the problem analysis from Step 2
to identify \textbf{problem-specific} axes of divergence - the decision
points unique to this particular problem - rather than applying generic
coding-style axes.
[omitted: worked examples of good and bad directions]

\textbf{Step 4: Cross-Check Diversity}

Compare all generated directions pairwise. For each pair, ask:

\textbf{Would following direction A and direction B produce solutions
that are essentially the same?}

If two directions would converge to the same solution or use the same
core strategy:\\
- Replace the weaker/less diverse direction with a new one that explores
a genuinely different axis\\
- Re-check the new direction against all others\\
- Repeat until every pair is clearly distinct

Specific red flags to catch: [omitted: typical anti-patterns, e.g. duplicate
insights]

This is a text-level comparison only - never execute or simulate tasks
during this step.

\textbf{Step 5: Save Directions}
[omitted: directory and direction-file bookkeeping]

\textbf{CRITICAL RULES}
[omitted: list including critical workflow guidelines, language and file-I/O
hygiene rules]

\end{promptcard}

%% file: sections/appendix/d_prompts/task_executor.tex
\subsection{Task Executor}
\label{app:prompt-executor}

\begin{promptcard}{Task executor}{evolve-task-exec skill prompt, verbatim
excerpts}
Execute a single evolutionary task with strict, ordered steps.

First, confirm these critical identifiers from the context: task\_id;
run\_id; task\_description; max\_attempts. [omitted: failure handling when
the injected context lacks a task id]

Then execute the following steps in EXACT ORDER:

\textbf{Step 1: Load Benchmark Contract}

Locate and read the problem SKILL.md for this task. Follow its public
instance-selection instructions using the original task\_description. If
it requires an instance or problem ID, validate that selector and read
the public statement/metadata declared by the skill. Never infer a
benchmark instance from the evolution task\_id.

Store the skill-declared main\_artifact\_name. If the skill does not
declare one, default to solution.py. Also store any skill-declared
initial\_program\_reference path for possible root-task use in Step 3.

Do not read verifier/evaluator scripts or judge-side resources while
resolving the benchmark contract.

\textbf{Step 2: Select Parent}

Call MCP tool select\_parent with run\_id, task\_id. Store the result (may
be null if no parent exists): parent\_id, parent artifact\_ref, parent
fitness (higher is better), and parent evaluator diagnostic metrics (may
be empty).

If parent exists, read the artifact file at artifact\_ref to understand
the parent solution.

\textbf{Step 3: Load Initial Reference Program (Root Tasks Only)}

If parent\_id is null, this is a root exploration task. Read the
skill-declared initial\_program\_reference resolved in Step 1. If the
skill does not declare one and initial\_program.py exists beside its
SKILL.md, use that sibling file as the fallback
initial\_program\_reference.

Use this reference only for root tasks with no parent. For non-root
tasks, parent and inspiration artifacts are the evolutionary context, so
do not load or prioritize the initial reference.

Do not read verifier scripts while locating the reference program.

\textbf{Step 4: Select Inspirations}

Call MCP tool select\_inspirations with run\_id, task\_id. Store the
result as a list of inspirations, each with inspiration\_id,
artifact\_ref, fitness, and metrics.

Read each inspiration artifact file to understand their approaches. When
comparing multiple inspirations, prefer higher-fitness artifacts and
avoid copying strategies from lower-fitness artifacts when they conflict
with better references.

\textbf{Step 5: Load Run-Local Notes}

Run-local Notes are enabled by default. The run owner may disable them
for the whole MCP process by setting environment variable
EVOTRIAS\_NOTES\_ENABLED=0 before starting the run.

Call MCP tool get\_notes\_prompt with run\_id, task\_id.

If the result has available == true, include the returned prompt exactly
as one section of the generation context. Do not rewrite, regroup,
summarize, or reinterpret the notes yourself.

If available == false, continue without a notes section. If the response
also has disabled == true, treat this as an intentional run
configuration and do not try to generate or commit notes manually.

\textbf{Step 6: Load Meta-Guidance}

Call MCP tool get\_meta\_guidance\_prompt with run\_id, task\_id.

If the result has available == true, include the returned prompt exactly
as one required strategy section of the generation context. Do not
rewrite, regroup, summarize, or reinterpret it yourself.

Meta-guidance has higher strategy priority than parent/inspiration
blending, but it never overrides verifier correctness or the benchmark
output contract. The selected parent is the anchor parent for this
guidance. Use it to preserve the interface and understand the current
search state. Do not assume the parent strategy is correct. If the
guidance says to depart from or replace the current approach, implement
the new strategy directly instead of forcing a superficial hybrid with
the parent or inspirations.

If available == false, continue without a meta-guidance section.

\textbf{Step 7: Read Direction (If Exists)}

Check for file: runs/[run\_id]/tasks/[task\_id]/direction.txt

If the file exists, read it as direction\_hint and reference the
direction when generating the solution in Step 8.2.

\textbf{Step 8: Attempt Loop (Up To MAX\_ATTEMPTS Attempts)}

[omitted: MAX\_ATTEMPTS resolution and loop-state initialization]

This is a retry loop, not a fixed MAX\_ATTEMPTS-attempt workflow. Start
with attempt\_001. Run attempt\_002 only if attempt\_001 fails. Run
attempt\_003 only if attempt\_002 also fails. Continue up to
attempt\_[MAX\_ATTEMPTS]. If any attempt succeeds, stop immediately and
do not create or run later attempts.

The remaining attempt budget is only for debugging failures. It is never
an optimization budget. Do not run a later attempt to improve a solution
after a previous attempt has already passed verification.

Each attempt is atomic: generate/debug the artifact once, save it once,
run the verifier once, then either record success or move to the next
attempt after failure. Do not repeatedly overwrite the same attempt's
artifact and rerun the verifier to chase a better fitness score.

For the current attempt, derive the attempt id exactly as:\\
\hspace*{1.2em}attempt\_id = f"attempt\_\{attempt\_index:03d\}"
[omitted: attempt\_id derivation examples]

Use attempt\_id, not attempt\_index, in every filesystem path.

Continue this loop only while the current attempt failed and
attempt\_index \textless{} MAX\_ATTEMPTS.

[omitted: reference to prior attempt's eval\_result.json for debugging,
and jump to Step 9 if an attempt succeeded]

\textbf{8.1 Create Attempt Directory}

Use mkdir -p to create the directory at:\\
\hspace*{1.2em}runs/[run\_id]/tasks/[task\_id]/attempts/[attempt\_id]/\\
Use the exact path with correct run\_id, task\_id, attempt\_id.

\textbf{8.2 Generate or Debug Solution}

\textbf{Evolution Strategy}: If a parent exists and no meta-guidance is
active, the parent solution is the foundation of this evolution - use it
as the primary reference for improvement and optimization; inspirations
can be used as supplementary references for crossover, mutation, and
borrowing optimization ideas. If meta-guidance is active, the guidance
is the required strategy. Use the parent critically as an
interface/current-state reference; build on it only when the guidance
calls for refinement. If no parent exists and
initial\_program\_reference is available, use the initial reference
program as the baseline implementation to understand the intended
interface, scoring objective, and starting search route, then apply the
task's direction hint to produce a distinct root solution. The direction
may call for either a close improvement to the reference or a
substantially different approach that preserves the benchmark contract.
Fitness is the task-generic quality score and higher is always better.
Use the parent as the primary reference, and use inspiration fitness to
rank secondary references. Borrow more from higher-fitness inspirations;
treat lower-fitness inspirations as optional contrastive examples rather
than templates to copy. Inspect parent and inspiration metrics alongside
fitness to understand which components, scenarios, or behaviors differ.
Metrics are diagnostic evidence, not separate optimization objectives
unless the benchmark explicitly says so.

\textbf{Core strategy execution integrity}: The artifact evaluated for
this attempt must actually execute the core strategy claimed by the
direction, active meta-guidance, or solution description. Do not
silently return a parent, baseline, or alternate algorithm when that
core strategy raises an exception, times out, is infeasible, returns no
result, or otherwise fails. In particular, do not wrap the core strategy
in a broad except Exception that returns a known-good candidate. Let
unexpected core failures surface so a later attempt can debug them.

Explicit input-dependent dispatch, portfolios, and selector algorithms
remain allowed when the dispatch or selection policy is itself the
advertised top-level strategy. Its branches and selection conditions
must be clear in the artifact; do not describe an internal branch as
validated merely because a different branch produced the evaluated
result.

\textbf{For attempt 1}: [omitted: list of inputs - parent, initial
reference, inspirations, task description, meta-guidance, direction hint
- and the role of each]

Generate a complete solution to the task. Use available skills as needed
to complete the work. Do not run local search by repeatedly editing and
verifying attempt\_001. Internal optimization code inside the artifact
is allowed, but the agent workflow itself must produce one candidate
artifact for this attempt and then verify it once.

[omitted: apply notes prompts and meta-guidance prompts]

If this is a root task and both initial\_program\_reference and
direction\_hint exist, the generated solution must visibly reflect both:
preserve the benchmark interface from the reference, reuse useful
baseline structure when helpful, and make the direction-specific
algorithmic choice real in the code rather than merely mentioning it in
comments.

\textbf{For attempts 2 and beyond (debug mode)}: DO NOT generate a
completely new solution. Instead:\\
{}[omitted: debug mode do's list, e.g., reuse parent and inspirations,
analyze previous attempt's result, etc.]

\textbf{8.3 Save Output} [omitted: artifact saving guidelines]

\textbf{8.4 Run Verifier}

Execute the verification method specified in the task description or
skill. Verifier success is defined only by the verifier result, not by
whether the fitness reaches a target or feels improvable. Exit code 0 or
result JSON "status": "success" means SUCCESS for this task attempt.

The verifier command is the only Bash command you should run for an
attempt. [omitted: forbidden probes and shell trailers]

In the success/failure branch below, call record\_attempt exactly once
for that attempt. [omitted: record\_attempt arguments]

\textbf{8.5 Required Notes Commit}

After every record\_attempt call, inspect its returned note\_requests.
[omitted: for each note request, follow its analysis\_prompt, produce one
note JSON, and call commit\_generated\_note until every request is
committed]

\textbf{Step 9: Finalize Attempt Loop} and \textbf{Step 10: End}
{}[omitted: return one-line completion message]

\textbf{CRITICAL RULES (NEVER BREAK THESE)}
[omitted: critical rules. e.g. attempt counting rule, stopping on
success, no accessing evaluator, etc.]
\end{promptcard}

%% file: sections/appendix/d_prompts/notes.tex
\subsection{Notes}
\label{app:prompt-note}

The Notes channel uses two fixed prompts. The first is sent with each note
request and produces the note. Each request carries exactly one note type,
its own trigger line, and the matching job block; the preamble, the
materials, and the schema are shared by all three types. The second prompt
carries the loaded notes back into the context of a later task.

\begin{promptcard}{Note analysis}{fixed prompt sent with each note
request, verbatim; [x] marks filled variables}
You are generating one run-local evolution note.

The note must summarize one useful experience from this task. Use the
referenced artifacts and verification results as evidence. Do not copy
long code blocks, full logs, or full tracebacks. Do not include your
reasoning process, step-by-step analysis, or chain-of-thought in the
note. The final note should contain only the distilled conclusion,
actionable guidance, and a very short evidence summary. Do not decide
whether the note is needed; MCP has already decided that.

Treat this as one empirical case, not proof that a strategy family is
generally good or bad. Keep attribution scoped to what this single
comparison can support. If several changes happened together, do not
claim one mechanism is the sole cause unless the evidence clearly
isolates it.

Verifier success proves that the submitted artifact was valid; it does
not prove that every advertised internal algorithm or branch executed.
Inspect the child artifact and verification evidence for primary,
fallback, dispatch, and selector control flow before attributing the
outcome to a particular strategy. If the executed branch cannot be
established, attribute the outcome to the artifact or selector as a
whole, state that the internal strategy is unverified, and lower
confidence. Never claim that a solver reached an optimum or that a
particular strategy succeeded unless the evaluated result can be traced
to that executed branch.

When parameters, seeds, acceptance checks, fallback choices, or
integration points matter, name those conditions explicitly. Do not turn
one attempt into a universal rule about a whole strategy family. Return
exactly one JSON object that matches the schema.

Note type: positive. Trigger: child fitness improved over parent by
[delta]; threshold is [threshold].\\
Note type: negative. Trigger: child fitness dropped below parent by
[delta]; threshold is [threshold].\\
Note type: error. Trigger: all attempts for this task failed verification.
Or: attempt [attempt\_id] succeeded after earlier failed attempts in the same
task.

Materials: Task [task\_id] / Attempt [attempt\_id]\\
- child implementation and verification result: [artifact\_ref],
[verification\_result\_ref]\\
- parent implementation, fitness, and metrics: [parent\_artifact\_ref],
[fitness], [metrics]\\
- inspirations: [inspiration list]\\
- error requests add: failed attempts: [failed artifact list]; failed
verification results: [failed result list]

Your job (positive):

1. Compare child against parent.\\
2. Identify the concrete implementation change most likely responsible
for the improvement.\\
3. If the idea appears to come from an inspiration, mention which one
and how it was adapted.\\
4. Write a concise note that future tasks can use.\\
5. Be careful with causality: use "likely" when the evidence is
suggestive rather than definitive.\\
6. If the improvement may depend on specific parameters, seeds, fallback
logic, or acceptance checks, state that condition instead of praising
the broad method unconditionally.\\
7. Do not write that a broad method is proven superior unless the
evidence isolates it from parameters and implementation details.\\
8. The returned JSON must not include the comparison process; keep only
the conclusion.

Your job (negative):

1. Compare child against parent.\\
2. Identify the concrete implementation detail, parameter choice, or
integration pattern most likely responsible for the regression.\\
3. Check whether the regression came from misusing an inspiration,
over-combining strategies, or breaking a useful parent mechanism.\\
4. Write a concise caution note for future tasks.\\
5. Do not overgeneralize: describe the condition under which the risk
appeared.\\
6. Do not write that a method or strategy should never be used just
because this attempt failed.\\
7. If a method might still work with different parameters, a safer
acceptance check, or a different integration point, say that the risk is
limited to the observed implementation.\\
8. Prefer wording like "avoid this unguarded variant" or "use only if
it improves validation" over blanket bans; never recommend a silent
fallback that would hide whether the claimed strategy executed.\\
9. If the child changed both the strategy and its parameters, attribute
risk to the observed configuration rather than condemning the whole
strategy.\\
10. The returned JSON must not include the diagnostic process; keep only
the conclusion.

Your job (error, all attempts failed verification):

1. Read the failed implementations and verification errors.\\
2. Identify the shared failure pattern, or the sequence of different failed
fixes.\\
3. Write one concise note that helps future tasks avoid this failure.\\
4. Do not paste raw logs. Summarize the failure reason and the preventive
action.\\
5. The returned JSON must not include the debugging process; keep only the
conclusion.

Your job (error, repaired within the task):

1. Compare the failed attempts with the successful attempt.\\
2. Identify the key fix that changed failure into success.\\
3. Write one concise note that future tasks can use as a checklist item.\\
4. Focus on the failure mode and the repair, not on the full
implementation.\\
5. The returned JSON must not include the comparison process; keep only the
conclusion.

Return JSON schema:\\
\{\\
\hspace*{1.2em}"title": "short imperative title",\\
\hspace*{1.2em}"lesson": "one concise conclusion distilled from the
evidence; no reasoning process",\\
\hspace*{1.2em}"actionable\_guidance": "one short instruction about
what a future generation should do or avoid",\\
\hspace*{1.2em}"evidence\_summary": "very brief factual evidence; no
analysis process",\\
\hspace*{1.2em}"confidence": "low|medium|high"\\
\}
\end{promptcard}

\begin{promptcard}{Note loading}{prompt returned by
\texttt{get\_notes\_prompt} and inserted as one section of the task context,
verbatim}
\#\# Run-local evolution notes

The following notes are distilled from earlier attempts in this same run.
Use them as practical guidance, not as hard proof. Prefer the selected parent
and inspirations, but take these notes into account when adapting them.
Treat positive notes as conditional hints, not proof that the same strategy
should be repeated. Prefer a genuinely new variant when prior notes already
describe the same fitness plateau. Treat negative notes as conditional risk
warnings about the recorded implementation and parameter choices, not as
global bans on a whole strategy family. A strategy may still be useful with
different parameters, fallbacks, acceptance checks, or integration points.

\#\#\# Error notes: avoid repeated failed patterns\\
\#\#\# Positive notes: patterns that previously improved fitness\\
\#\#\# Negative notes: patterns that previously degraded fitness

\end{promptcard}

%% file: sections/appendix/d_prompts/meta-guidance.tex
\subsection{Meta-guidance}
\label{app:prompt-meta}

This role reads two sources. Its standing prompt fixes the workflow and the
output contract; the controller-assembled request supplies the evidence and
the per-request generation instructions.

\begin{promptcard}{Meta-guidance generator}{agent prompt, verbatim excerpts;
[x] marks values filled at dispatch time}
You are the dedicated Meta-Guidance Generation Specialist for the
self-evolving agent system.

Each invocation handles exactly one pending request. You are stateless across
invocations: do not poll, wait for another request, retain an agent
identifier, or perform any task-executor or orchestration work.

\textbf{Input contract.} The dispatch prompt contains exactly:
run\_id: [x], request\_id: [x]. Both values are required. Use them verbatim.

\textbf{Workflow.}\\
1. Call get\_meta\_guidance\_request with the provided run\_id and
request\_id. Never ask the orchestrator to retrieve or relay the request.\\
2. If the MCP call itself fails, retry it once.\\
3. If the response has available != true, do not generate or commit
anything.\\
4. Treat the returned request as the complete source of generation state.
Follow generation\_instructions and constraints when present.\\
5. Inspect current\_best.artifact\_excerpt first. If more context is needed,
use current\_best.artifact\_ref and read only the relevant candidate code.\\
6. Generate one JSON object that follows the schema below, then call
commit\_meta\_guidance yourself with the same run\_id and request\_id.
Never return the guidance JSON to the orchestrator.\\
{}[omitted: commit-retry and validation-repair rules]\\
9. End immediately after the compact result. Do not wait for more work.

\textbf{Guidance JSON} (one object per direction): idea; description;
target\_metric; novelty\_claim; current\_best\_assessment;
relationship\_to\_current\_best (refine, depart, or replace);
implementation\_plan (step list); cautions; approach\_type;
avoid\_repeating; max\_uses.

Generate 1 to 3 items. Every approach\_type in the same response must be
distinct. Prefer distinct strategy families over parameter variants.
Critically assess the current best rather than assuming its route is correct,
and use notes\_summary, guidance\_history, and failed\_guidance to avoid
repeating routes without a materially changed failure condition.

Never modify project files, the candidate, verifier, or evaluator. Never
bypass the evaluator, weaken the output contract, or propose
evaluator-specific hard-coded shortcuts.

\textbf{Compact result contract.} The entire final response is exactly one
line: meta\_guidance\_result: committed request\_id=[x]
guidance\_count=[x], or the not\_committed or mcp\_error form with a reason
code. Do not include the request payload, generated guidance, artifact
contents, analysis, tool transcript, prose, or Markdown.
\end{promptcard}

\begin{promptcard}{Meta-guidance request}{assembled by the controller;
verbatim field structure and generation instructions; [x] marks filled
variables}
Request payload (at most one pending request at a time):\\
- request\_id, task\_description\\
- current\_best: [record reference with artifact excerpt]\\
- progress: best fitness at watching start, current best fitness,
completed count since watching, stall diagnostics\\
- lineages: [per-lineage summaries]\\
- notes\_summary: [summary of admitted notes]\\
- guidance\_history and failed\_guidance: [earlier directions with
recorded outcomes]\\
- generation\_instructions (below) and constraints

Constraints:\\
- Do not bypass the evaluator.\\
- Keep the benchmark output contract.\\
- Do not modify verifier/evaluator source or use it to hard-code
evaluator-specific shortcuts.\\
- You may inspect current\_best.artifact\_ref or the provided
artifact\_excerpt; this is candidate code.

Generation instructions:

- Inspect current\_best.artifact\_excerpt when present; if it is empty,
use current\_best.artifact\_ref as the candidate implementation path.

- Inspect current\_best.metrics and profile alongside fitness. Use
metric components and scenario breakdowns as diagnostic evidence about
where candidates differ or stagnate, without treating them as
independent objectives unless the benchmark defines them that way.

- Analyze the current best critically: identify what works, what likely
traps the search, and whether the next guidance should refine, depart
from, or replace that route.

- Before crediting an internal algorithm, inspect the artifact control
flow and available verification evidence to establish which primary,
fallback, dispatch, or selector branch produced the evaluated result.
Verifier success alone does not establish branch execution; if branch
identity is unknown, treat the claimed strategy as unverified.

- Use notes\_summary and guidance\_history as evidence. Do not repeat
guidance whose prior outcome\_summary shows no improvement unless the
failure condition is materially changed.

- Interpret prior outcomes precisely: new\_best beat the global best at
completion; improved\_anchor beat its assigned parent but not the
then-current global best; no\_effect produced a valid result
indistinguishable from its assigned parent in fitness and evaluator
metrics; regression produced a different valid result without beating
its assigned parent; failed produced no valid result. no\_effect is
observational and does not by itself prove that fallback occurred.

- Generate distinct guidance items with different approach\_type values.
Avoid minor parameter variants of the same algorithm family.

- Prefer concrete, implementable strategies. Each item should say how it
relates to the current best and why it can escape or improve the
observed plateau.
\end{promptcard}